\documentclass{article}
\usepackage[mode=buildnew]{standalone}

\usepackage[preprint]{neurips_2025}

\usepackage{amsmath}
\usepackage{amsfonts}
\usepackage{bm}
\usepackage{enumitem}
\usepackage{booktabs}
\usepackage{multirow}

\def\eqref#1{equation~\ref{#1}}

\def\1{\bm{1}}

\def\rmE{{\mathbf{E}}}

\def\rmM{{\mathbf{M}}}

\def\rmP{{\mathbf{P}}}

\def\rmS{{\mathbf{S}}}

\DeclareMathAlphabet{\mathsfit}{\encodingdefault}{\sfdefault}{m}{sl}
\SetMathAlphabet{\mathsfit}{bold}{\encodingdefault}{\sfdefault}{bx}{n}

\def\gR{{\mathcal{R}}}
\def\gS{{\mathcal{S}}}
\def\gT{{\mathcal{T}}}

\def\sM{{\mathbb{M}}}

\def\sP{{\mathbb{P}}}

\def\sR{{\mathbb{R}}}

\def\sT{{\mathbb{T}}}

\usepackage[utf8]{inputenc}
\usepackage[T1]{fontenc}
\usepackage{listings}
\usepackage{graphicx}
\usepackage{caption}
\usepackage{subcaption}
\usepackage{booktabs}
\usepackage{multirow}
\usepackage{hyperref}
\usepackage{inlinedef}
\usepackage[poorman]{cleveref}
\usepackage{url}
\usepackage{booktabs}
\usepackage{amsfonts}
\usepackage{nicefrac}
\usepackage{chemarr}
\usepackage{microtype}
\usepackage{xcolor}
\usepackage{nicematrix}
\usepackage{tikz}
\usepackage{graphicx}
\usepackage{amssymb}
\usepackage{enumitem}
\usepackage{pifont}
\usepackage{algorithm}
\usepackage[noend]{algpseudocode}
\usepackage{wrapfig}
\usepackage{multirow}
\usepackage{colortbl}
\usepackage{xspace}
\usepackage{subcaption}
\usepackage{adjustbox}
\usepackage{titletoc}
\usepackage[symbol]{footmisc}
\usepackage[mode=buildnew]{standalone}
\usepackage{longtable}
\usepackage{booktabs}
\usepackage{tcolorbox}
\usepackage{multirow}
\usepackage{fontawesome5}
\usepackage{svg}

\newcommand{\ourbench}{\textsc{SciGym}}

\newcommand{\simulationmetric}{Simulation Trajectory Error (STE)}
\newcommand{\reactionmetric}{Reaction Matching Score (RMS)}
\newcommand{\topologymetric}{Network Topology Score (NTS)}
\newcommand{\simulation}{STE}
\newcommand{\reaction}{RMS}

\definecolor{sbml_blue}{RGB}{0,122,255}
\definecolor{sbml_green}{RGB}{47, 218, 95}
\definecolor{sbml_orange}{RGB}{255,153,77} 
\definecolor{sbml_red}{RGB}{255,0,106} 
\definecolor{sbml_purple}{RGB}{153,0,255} 

\definecolor{codegray}{rgb}{0.95,0.95,0.95}
\definecolor{codegreen}{rgb}{0.4,0.6,0.4}
\definecolor{codepurple}{rgb}{0.58,0,0.82}

\usepackage{listings}
\usepackage{xcolor}
\usepackage{tcolorbox}
\tcbuselibrary{listings}
\tcbuselibrary{breakable} 

\lstdefinestyle{markdown}{
  backgroundcolor=\color{codegray},
  commentstyle=\color{codegreen},
  keywordstyle=\color{blue},
  stringstyle=\color{codepurple},
  basicstyle=\ttfamily\small,
  breakatwhitespace=false,
  breaklines=true,
  captionpos=b,
  keepspaces=true,
  showspaces=false,
  showstringspaces=false,
  showtabs=false,
  tabsize=2
}

\newtcblisting{boxedmarkdown}{
  listing only,
  listing style=markdown,
  breakable,           %
  break at=20cm,        %
  pad at break=1mm,    %
  boxrule=0.5pt,
  colback=codegray,
  colframe=black,
  arc=0pt,
  outer arc=0pt,
  top=2pt,
  bottom=2pt,
  left=2pt,
  right=2pt,
  boxsep=0pt,
  listing options={style=markdown, language={}}
}

\title{Measuring Scientific Capabilities of Language Models\\with a Systems Biology Dry Lab}

\author{%
Haonan Duan$^{*}$  \\
University of Toronto
\And 
Stephen Zhewen Lu$^{*}$ \\
SickKids
\And 
Caitlin F. Harrigan \\
University of Toronto
\And 
Nishkrit Desai \\
Axiom
\And 
Jiarui Lu \\
Mila
\And 
Michał Koziarski \\
SickKids
\And 
Leonardo Cotta \\
Vector Institute
\And 
Chris J. Maddison \\
University of Toronto
}

\begin{document}

\footnotetext[1]{Equal contribution.}

\maketitle
\begin{abstract}
Designing experiments and result interpretations are core scientific competencies, particularly in biology, where researchers perturb complex systems to uncover the underlying systems. Recent efforts to evaluate the scientific capabilities of large language models (LLMs) fail to test these competencies because wet-lab experimentation is prohibitively expensive: in expertise, time and equipment. We introduce \ourbench, a first-in-class benchmark that assesses LLMs' iterative experiment design and analysis abilities in open-ended scientific discovery tasks. \ourbench overcomes the challenge of wet-lab costs by running a dry lab of biological systems. These models, encoded in Systems Biology Markup Language, are efficient for generating simulated data, making them ideal testbeds for experimentation on realistically complex systems. We evaluated six frontier LLMs on $137$ small systems, and released a total of $350$ systems. Our evaluation shows that while more capable models demonstrated superior performance, all models' performance declined significantly as system complexity increased, suggesting substantial room for improvement in the scientific capabilities of LLM agents.

\faGlobe \ Website: \url{https://h4duan.github.io/scigym-benchmark/}

\includegraphics[height=1em]{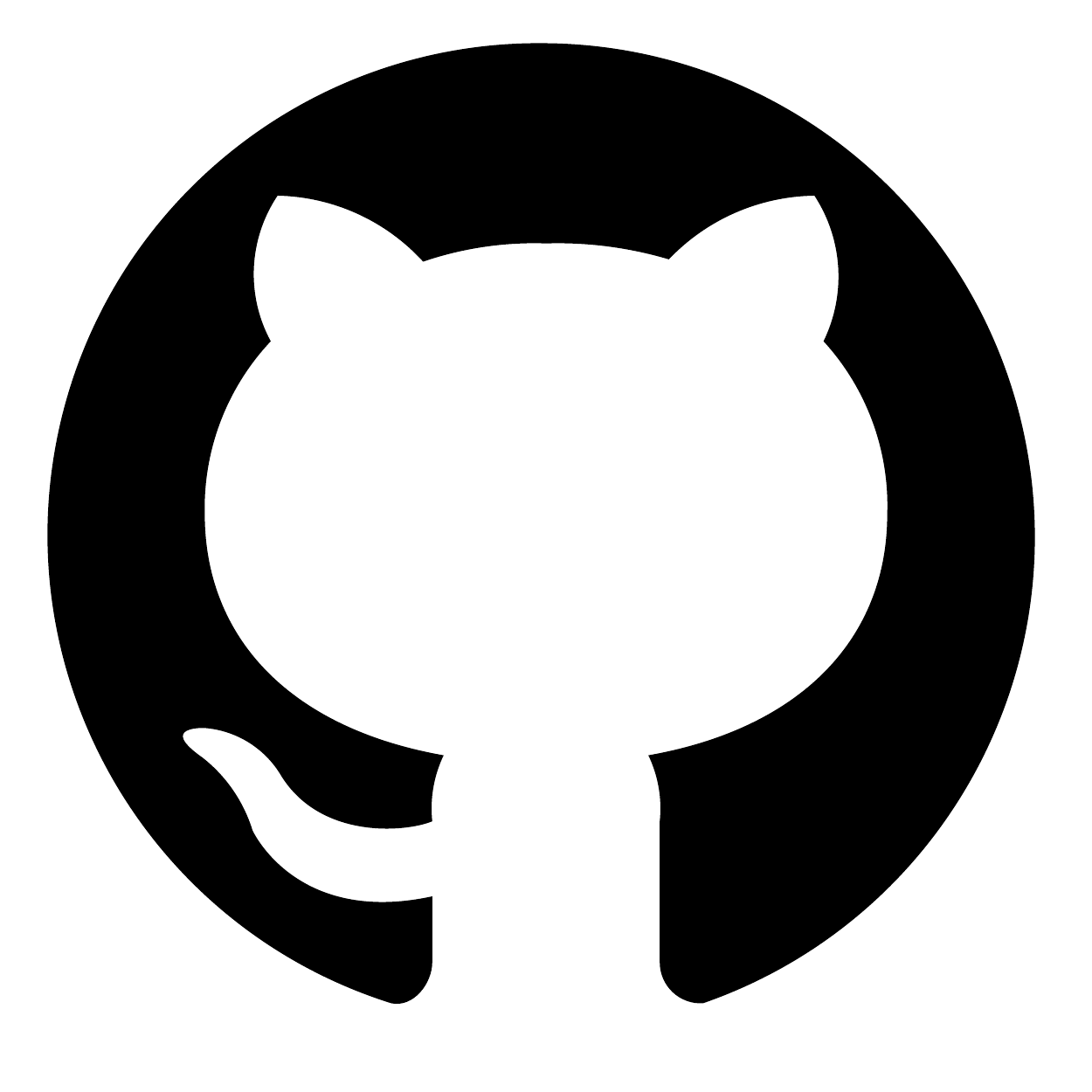} Code:
\url{https://github.com/h4duan/SciGym}

\includegraphics[height=1em]{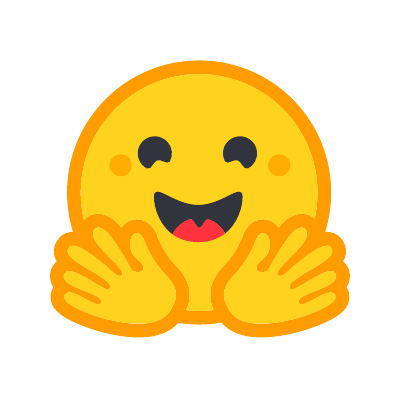} Data: \url{https://huggingface.co/datasets/h4duan/scigym-sbml}

\includegraphics[height=1em]{figures/hf-logo.pdf} Eval: \url{https://huggingface.co/datasets/h4duan/scigym-eval}

\end{abstract}

\section{Introduction}

Scientific experimentation is the primary tool that researchers in the natural sciences use to gain insight about our world's physical and biological systems. A researcher can test a hypothesis by systematically perturbing systems and observing effects. 
They then interpret their results and identify the next best experiment to perform, closing the scientific discovery loop. Thus, when assessing large language models (LLMs) capabilities as scientists, it is essential to have evaluation frameworks effectively testing these skills. However, in this context, a fundamental challenge emerges: \textit{How can we generate experimental data in the loop to evaluate an LLM's scientific iteration?}

For each experiment proposed by an LLM, we need to obtain data corresponding to how the system responds to the suggested perturbations. In other domains, the success of end-to-end benchmarks has hinged on our ability to quickly and automatically assess LLM actions. For example, SWE-bench tests an LLM's ability to resolve real-world GitHub issues, as evaluated by a suite of unit tests \cite{jimenez2024swebench}. Importantly, coding is done in formal programming languages that can be executed and analyzed cheaply. Unfortunately, this is not attainable with a traditional laboratory setup in which experiments are expensive and laborious to perform. Substantial progress has been made in robotics and control for self-driving labs, however these technologies are not yet mature enough to easily be used to assess LLM performance  at scale \cite{abolhasani2023rise}.

\begin{figure*}[t]
    \centering
    \includegraphics[width=\textwidth]{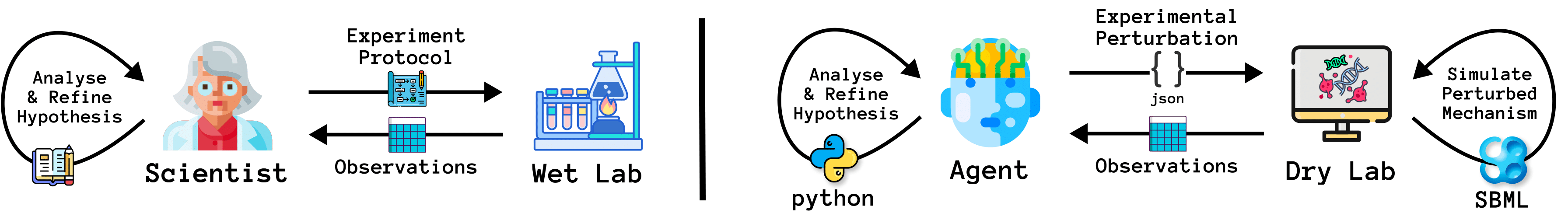}
    \caption{\textbf{\ourbench{} simulates scientific discovery.} Real-world scientists (left) iteratively refine their hypotheses by designing experiments, collecting lab data, and analyzing results. In \ourbench{} (right), agents request experimental actions, receive the observation data simulated from the perturbed reference system, and then perform analyses in a Python shell to refine their hypotheses.}
    \label{fig:scigym-overview}
\end{figure*}

A promising approach comes from systems biology, where formal mathematical models of various biological processes have been developed. 
BioModels \citep{malik2020biomodels} is a public repository that hosts manually-curated models from published literature in variety of fields ---such as cell signaling, metabolic pathways, gene regulatory networks, and epidemiological models of infectious disease. These models are stored in Systems Biology Markup Language (SBML), a standard exchange format  which provides a machine-readable representation supported by numerous software tools for simulation and analysis \citep{noauthor_sbmlorg_nodate, dubitzky_encyclopedia_2013, Medley2018TelluriumNE, hoops_copasi_2006}. Crucially, SBMLs can facilitate a ``dry lab'' in the context of a benchmark:
when an LLM requests an experiment, we can create an SBML file describing the proposed perturbation, and return the corresponding simulated data.

Based on this insight, we introduce \ourbench{}, an agentic benchmark that evaluates LLMs' end-to-end scientific discovery abilities. We leverage 350 SBML models from BioModels \citep{malik2020biomodels} ranging in complexity from simple linear pathways with a handful of species and reactions to sophisticated networks containing hundreds of molecular components and interactions. We release two benchmark splits: \textit{small}, which contains 137 models with fewer than 10 reactions each, and \textit{large}, which contains the remaining 213 models with up to 400 reactions each. Our framework operates as follows: the agent is tasked with discovering a reference system described by a biology model by analyzing data simulated from the SBML. The agent can perturb the simulated system and write Python code to analyze the resulting data. Performance is assessed by measuring correctness in topology of the graph of the true system, recovery of the reactions in the system, and percent error in data generated by a model proposed by the agent. To succeed in our tasks, the agent must excel in both experimental design and data analysis in an open-ended manner, mirroring the core competencies of human scientists.

In addition to our specific benchmarking settings, \ourbench is highly extensible, allowing researchers to add various configurations that test different tasks, abilities, and biological systems. Although these SBML models do not fully capture all physiological variables, these carefully curated dynamic systems represent a treasure trove of biochemical insights derived from expert knowledge. By harnessing these models to simulate from, we posit that faithfully recovering the underlying systems which generate the trajectories of species concentration over time is a challenge which is suitably complex to evaluate an agents’s scientific reasoning. We foresee this is a capability which will translate to real-world performance when it comes to inferring mechanisms from experimental data.

We evaluated six LLMs from three frontier model families (Gemini, Claude, GPT-4) on \ourbench-small. We found that more capable models generally outperform their smaller counterparts, with Gemini-2.5-Pro leading our benchmark followed by Claude-Sonnet. However, we also identified consistent limitations across all models: 1) performance decreases as the underlying biological system becomes more complex; 2) proposed mechanisms often overfit to experimental data without generalizing to unseen initial conditions; and 3) models struggle to identify subtle relationships, particularly those involving modifiers.

\section{Background on formal biology models and SBML}
\label{sec:sbml-def}

Mathematical modeling of biochemical reaction networks, such as metabolic, gene regulatory, and protein-signaling networks is one of the central tasks in systems biology \citep{sauro2020systems}. Motivated by the need to share models, researchers have developed standard machine-readable formats for describing chemical reaction networks. Among these, the Systems Biology Markup Language (SBML) has become a de facto standard for formal specifications of dynamic network models via an XML-based standard. Going forward, we use \texttt{texttt} when referring to specific SBML tag types. 

SBML adopts terminologies from biochemistry: \texttt{reaction} is the central object for describing processes that change the quantities of \texttt{species} (entities such as small molecules, proteins, \textit{etc}.) in a model. The \texttt{kineticLaw} of a \texttt{reaction} specifies in MathML the speed with which this process happens. Systems described by SBMLs can be simulated as systems of ordinary differential equations, as we discuss below, or other frameworks like discrete stochastic systems. The \texttt{reaction} tag can be understood in terms of the participating \texttt{species} found in the {\texttt{listOfReactants}}, {\texttt{listOfProducts}}, and possibly {\texttt{listOfModifiers}}. A \texttt{reaction} is a directed relationship: the quantities of reactants decrease as they are consumed, while the quantities of products increase as they are generated due to the reaction proceeding. Modifiers affect reaction rates, and their quantities remain unchanged through the reactions. A reaction can also define the relevant {\texttt{kineticLaw}} and {\texttt{listOfParameters}} to characterize the rate of reactions. SBML can include other optional components, such as \texttt{rule} and \texttt{event} tags. We refer readers to \Cref{sec:sbml-tags} major SBML components and the SBML Level 3 language specification for more details \citep{hucka_systems_2019}.

\begin{wrapfigure}{r}{0.4\textwidth}
    \centering
    \includegraphics[width=0.35\textwidth]{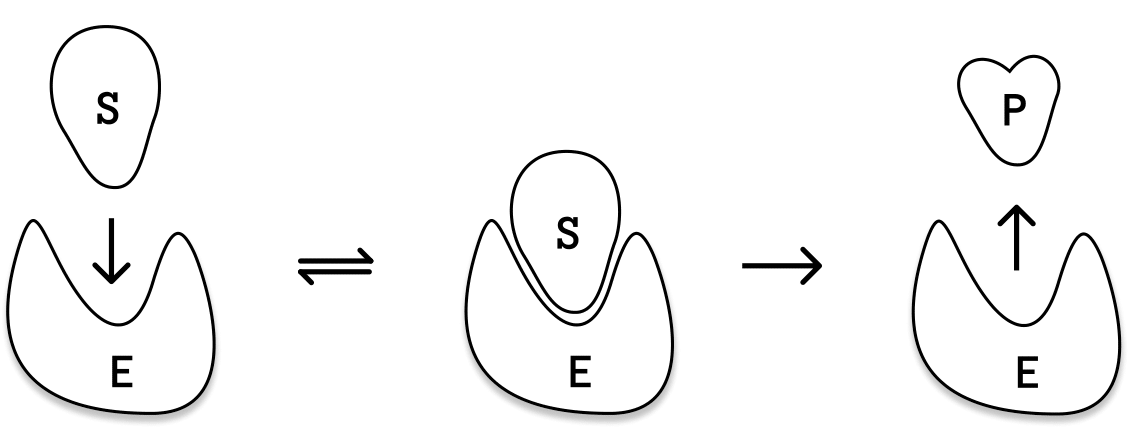}
    \caption{\textbf{Biological process in The Michaelis–Menten  enzymatic process.} The substrate $\rmS$ binds to the enzyme $\rmE$ (left) which catalyzes the formation of the product $\rmP$ (right).}
    \vspace{-1em}
    \label{fig:enzyme-diagram}
\end{wrapfigure}

\textbf{Reduced SBML Representation.} For our purposes, an SBML is a 4-tuple $(\gS,\Theta,\gR, \gT) \in \mathfrak{S}$ consisting of its \texttt{listOfSpecies} $\gS:=\{S_j\}_{j=1}^n$, \texttt{listOfParameters} $\Theta$, \texttt{listOfReactions} $\gR:=\{R_i\}_{i=1}^m$, and all other tags $\gT$. Each reaction $R_i: = (\sR_i,\sP_i,\sM_i, \theta_i, r_i, \sT_i)$ consists of its \texttt{listOfReactants} $\sR_i$, \texttt{listOfProducts} $\sP_i$, \texttt{listOfModifiers} $\sM_i$, \texttt{listOfParameters} $\theta_i$, a \texttt{kineticLaw} represented as a function $r_i: \mathfrak{S} \rightarrow\sR^+$, and all other tags $\sT_i$. 

\textbf{Example system.} The Michaelis–Menten~\citep{Michaelis2011TheOM} enzymatic process describes a system that produces product $\rmP$ from the substrate $\rmS$, catalyzed by the enzyme $\rmE$. The reaction is illustrated in \Cref{fig:enzyme-diagram} and represented via the chemical equation: $\text{E} + \text{S} \xrightleftharpoons[k_{\text{off}}]{k_{\text{on}}} \text{ES} \xrightarrow{k_{\text{cat}}} \text{E} + \text{P}$. The SBML model to describe this system consists of the set of species $\gS = \{\rmS,  \rmE,\rmE\rmS, \rmP \}$, two reactions $R_1, R_2$, and parameters $\Theta = \{k_{\text{off}}, k_{\text{on}}, k_{\text{cat}}\}$. The first reaction describes the reversible formation of the enzyme-substrate complex with rate $r_1(\gS,\Theta,\gR, \gT) = v k_{\text{on}}[\rmE][\rmS]- v k_{\text{off}}[\rmE\rmS]$, where $v > 0$ is the volume of the compartment where the reaction occurs, specified in $\gT$. The second reaction represents the conversion of the substrate to the product with rate $r_2(\gS,\Theta,\gR, \gT) = v k_{\text{cat}}[\rmE\rmS]$.  The square brackets $[\cdot]$ map a species to its time-varying concentration. We omit the time index to avoid notational clutter. 

To translate this into an ODE, we sum the rates implied by the reactions for each species. For $S_j$ we have $v \cdot d[S_j]/dt = \sum_{i=1}^m s_{ij} \cdot r_i(\gS,\Theta,\gR, \gT)$. $s_{ij}$ are stoichiometric coefficients (specified in $\sR_i$ or $\sP_i$) that describe how much of each species is consumed or produced in the reaction. They are signed according to the role of $S_j$ in $R_i$: concentration decreases and $s_{ij} < 0$ if $S_j$ is a reactant in reaction $R_i$, concentration increases and $s_{ij} > 0$ if $S_j$ is a product, and $s_{ij} = 0$ otherwise. We include a full worked example of this enzyme system's SBML and ODE in \Cref{sec:sbml-example}, and a worked example of an SBML with a modifier in \Cref{sec:sbml-modifier-example}.

\begin{figure}
    \centering    \includegraphics[width=0.9\linewidth]{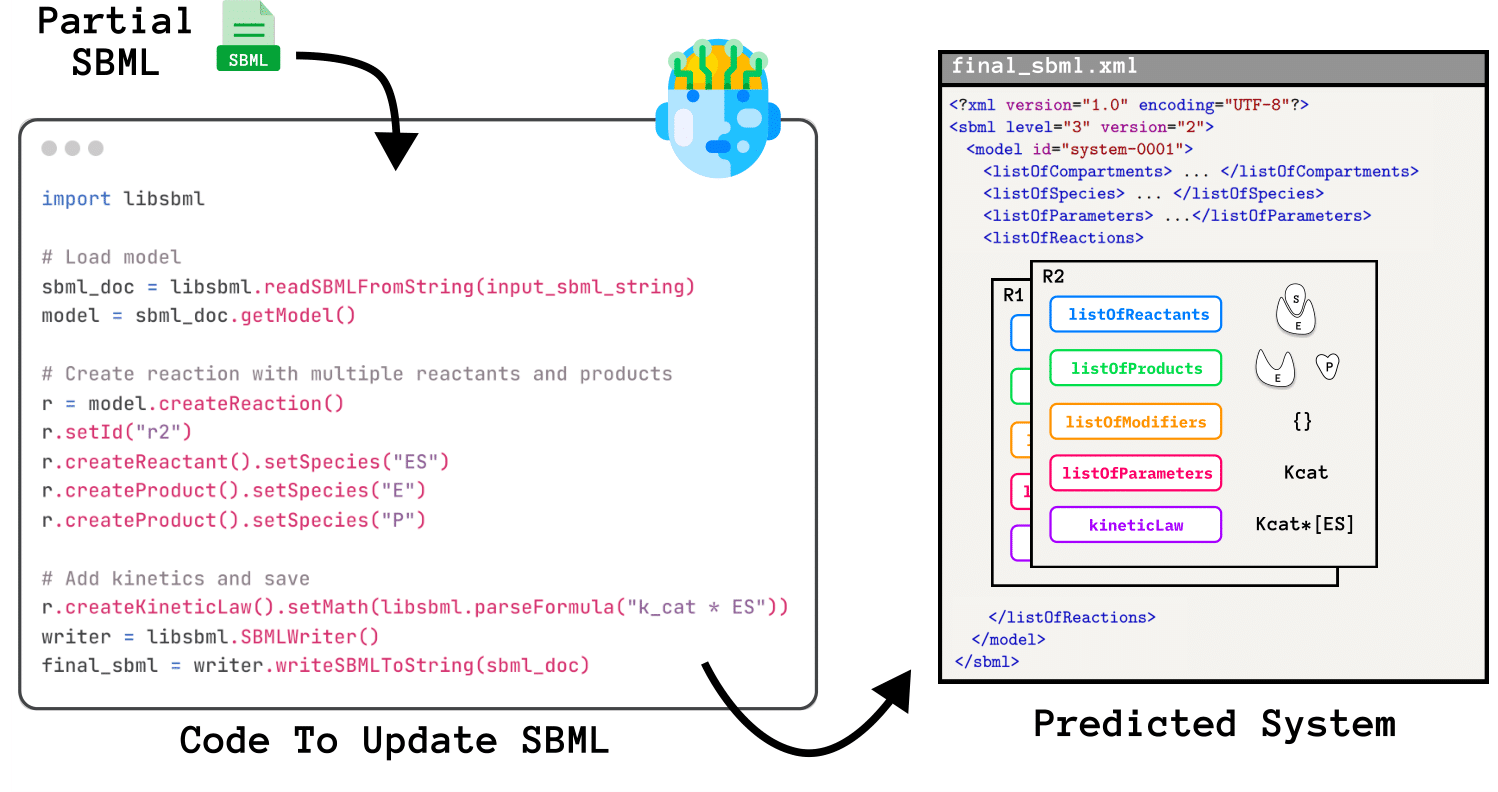}
    \vspace{-1em}
   \caption{\textbf{Open-ended scientific iteration} Agents must discover all components of missing reactions and can update the SBML models using Python code (left). SBML models are encoded as structured XML files, with reactions consisting of five components: reactants, products, modifiers, kinetic laws and parameters (right).} 
   \label{fig:sbml-anatomy}
\end{figure}

\textbf{Simulation.} %
Numerous simulation softwares have been developed to efficiently generate time-series data from SBML models, with the most popular ones being libRoadRunner \citep{somogyi2015libroadrunner}, and COPASI \citep{hoops_copasi_2006}. These simulators are typically based on ordinary differential equation (ODE) solvers. They are highly efficient and can handle various complexities in SBML specifications, such as stiff equations arising from different reaction timescales and algebraic constraints defined by rules. In this work, we use {Tellurium} \citep{Medley2018TelluriumNE}, a python-based library using libRoadRunner as the backend simulation engine.

\section{\ourbench{} Benchmark}

\subsection{Data Curation}

The full curation pipeline is shown in \Cref{fig:benchmark-curation-pipeline}, and we describe briefly the major steps below.

\textbf{Filtering}. Starting from the curated collection of 1096 peer-reviewed models in the BioModels repository \citep{malik2020biomodels}, we first removed models with no reactions or no species. Next, we removed models containing \texttt{rule} and \texttt{event} tags. We found that rules could leak information about held-out reactions, artificially inflating scores. On the other hand, agents often got stuck trying to trigger events, which can require complex combinations of interventions that our experimental interface may not support. We classified the models in our benchmark by their curated gene ontology~\citep{Ashburner2000GeneOT} (GO) terms (\Cref{fig:go-ontology}). Finally, we removed models that failed to parse with libsbml or failed to simulate using Tellurium. This resulted in 350 models in our benchmark.  We provide a detailed breakdown of removed instances in \Cref{apx:filtering-sbml}.

\textbf{Preprocessing}. In order to prevent memorization of SBML documents by the agent, we preprocessed each of the filtered models by stripping optional metadata fields, shuffling core components, and anonymizing identifiers to a unique 4-character alphanumeric string (see \Cref{apx:sim-timescale}). This preprocessing step is how we derive an initial partial SBML which the agent has access to, from the reference system which is kept hidden from the agent. 

\textbf{Determining Simulation Duration}. We found that a large proportion of the models in the BioModels database did not provide an appropriate simulation timescale in order to observe the characteristic behavior of the mechanism. Thus, we systematically analyzed each model in our benchmark and chose a simulation duration that either enabled the system to reach a steady state or a maximum duration of 10000 seconds (see \Cref{apx:sim-anon}).

\subsection{Framework}

\textbf{Task.} As illustrated in Figure \ref{fig:scigym-overview}, SBML models provide a general framework to simulate the scientific discovery process: an SBML model is used to generate time-series data; the agent is tasked with discovering the underlying reference system by designing experiments and analyzing their results. One can design tasks of varying difficulty by modulating the agent's access to the system (what species can be observed and what perturbations are permitted). We decided to focus on a fully-observed reaction discovery task, but \ourbench{} is easily extended to any task or degree of observability that SBML simulators can support. We allowed the agent to fully observe the concentration time series of all \texttt{species} and asked it to discover the missing \texttt{reaction} tags connecting \texttt{species} in the system. This task mirrors fundamental research problems in biology, such as inferring gene regulatory networks from Perturb-seq experiments~\citep{dixit2016perturb} or reconstructing cellular signaling pathways from spatial transcriptomics data~\citep{Papin2005ReconstructionOC}. 

Starting with a reference SBML $m_{\text{ref}} = (\gS, \Theta, \gR, \gT)$, the agent is given a partial SBML $m_{\text{hyp}} = (\gS, \Theta', \emptyset, \gT)$ where the reactions $\gR$ have been removed completely and the parameter set $\Theta'$ contains all parameters except those that exclusively appear in a reaction's $\theta_i$. Additionally, we remove \texttt{initialAssignment}, \texttt{functionDefinition}, and \texttt{constraint} tags that reference the removed reactions to prevent information leakage. The agent is given all units, compartments, and species from the original model. The agent's task is to recover the missing reactions $\gR$ in $m_{\text{ref}}$.

\textbf{Agent.} We implemented a ReAct-style agent \citep{yao2023react}using a Thoughts-Actions-Observations framework due to its simplicity for comparing performance across different models. The agent is required to articulate its reasoning process before taking actions, structuring its response in markdown format with \verb|## Thoughts| and \verb|## Action| sections. The environment's outputs (code execution and experiment results) are then added as \verb|## Observation| sections. In the prompt, we inform the agent that its goal is to discover the underlying biological mechanisms. To maintain an open-ended task that better mirrors actual scientific discovery processes, we deliberately avoid specifying exact evaluation metrics to the agent. The system prompt is provided in Section \ref{apx:system-prompt}.

\begin{wrapfigure}[25]{r}{0.6\textwidth}
    \centering    \includegraphics[width=\linewidth]{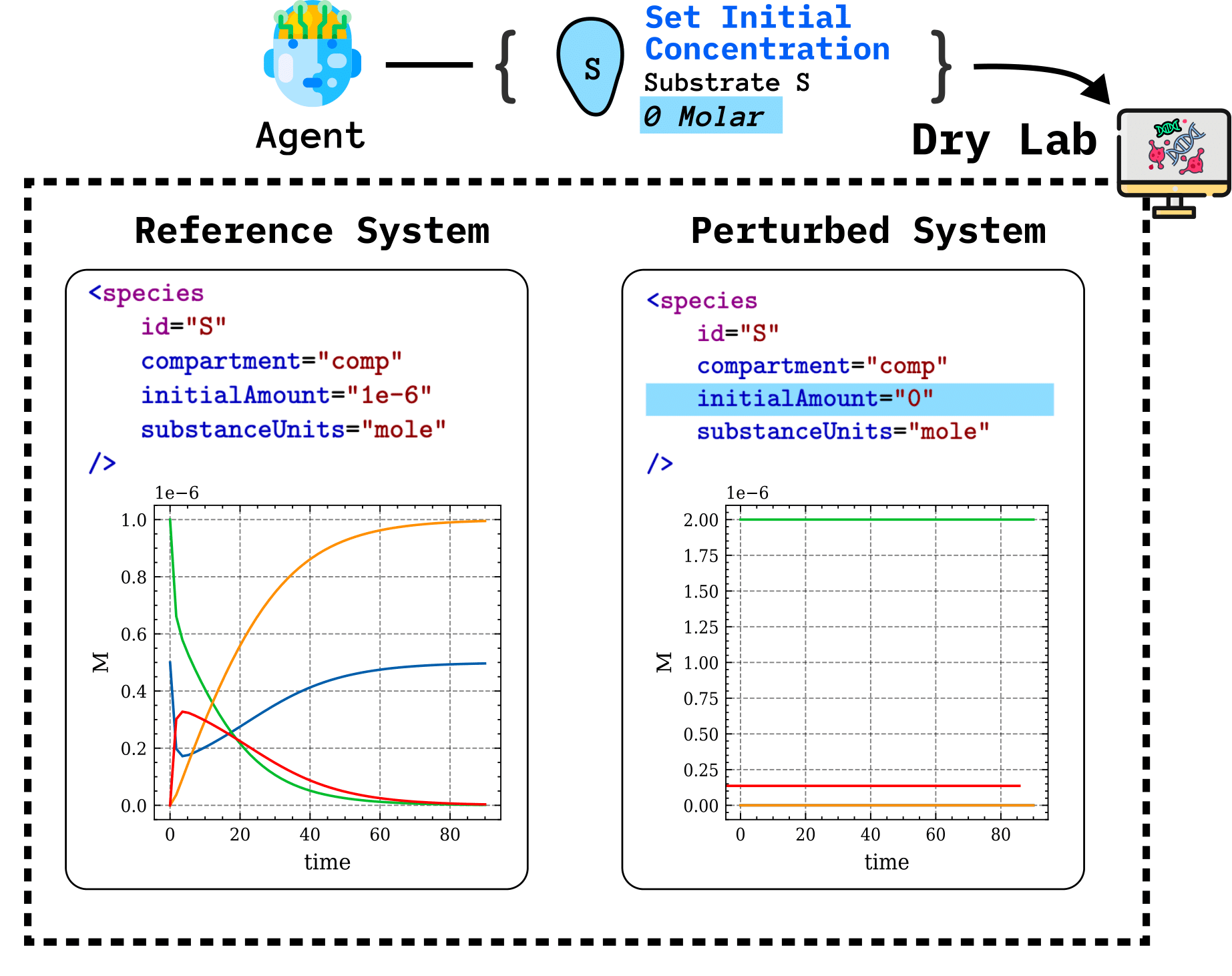}
    \label{fig:perturbation-actions}
    \vspace{-2em}
   \caption{\textbf{SBML models serve as data simulators for various experimental perturbations.} The agent proposes perturbations to the system, such as changing its initial conditions and can receive the observed data of the perturbed system.}
\end{wrapfigure}

\textbf{Action Space.} In each iteration, the agent can choose from three actions: writing code, conducting experiments, or submitting the model. The agent can choose to write code or conduct experiments or both within a single iteration. However, once the agent chooses to submit the model, no further experiments or code execution is permitted. Each action is signaled with a corresponding markdown subsection, with specific response formats detailed in the system prompt.

\textbf{Coding.} We provide a Python execution environment that enables the agent to analyze experimental results. When the agent chooses to write code, the environment executes it and returns stdout output or stderr messages for any errors. The agent is instructed in the system prompts to use print statements for important outputs that should be communicated back to its reasoning process. The environment maintains state as global variables in its Python shell between iterations, but the agent must explicitly choose which variables to store for future access. This design allows the agent to reuse some important data structures especially large objects like pandas DataFrames. The agent has access to a specified set of libraries and customized functions. A customized tool documentation is provided with docstrings (example in Section \ref{apx:tool-manual}). In our experiments, we provide a wrapper around the {Tellurium} simulator as the customized functions to help them simulate the proposed mechanisms.

\textbf{Experiments.} The agent is provided with a list of permitted experimental perturbations, detailed in an experiment manual (example in Section \ref{sec:experiment-manual}). This documentation explains the effects of available perturbations and specifies the JSON format for requesting each type. In our experiments, we allow agents to use one simple type of perturbations: changing the initial concentrations of a species, with a detailed discussion in Section \ref{sec:experiments}. When the agent requests an experiment, our environment applies the specified perturbation to the SBML model (using libSBML libraries), simulates the modified system, and returns time-series data to the agent. The resulting time-series data is stored in a dictionary where the key is the iteration number and the value is the pandas DataFrame. Each dataset remains accessible throughout the session, allowing the agent to reference previous experimental results. The environment also provides a summary of each dataset in the observation output, giving the agent an overview of the experimental results to inform its subsequent actions.

\textbf{Model Submission.} The agent may submit its final model and conclude the discovery process at any point, subject to a maximum iteration limit ($20$ in our benchmark). To submit a model, the agent must provide a code block that defines the final model in a variable named \texttt{finalsbml}. The agent has direct access to the initial incomplete SBML structure \texttt{inputsbml} and can reference it in their code. This approach encourages the agent to use libSBML libraries for model manipulation rather than manual SBML construction. If the submitted model is invalid or cannot be simulated, we allow for 3 additional debugging iterations before evaluating against the incomplete SBML structure.

\begin{table}[t]
\caption{\textbf{Pro models outperform their mini counterparts in \ourbench{}, with Gemini dominating in both size categories.} We report the average Simulation Trajectory Error (STE) and Reaction Matching Score (RMS) across all benchmark instances. Bold values indicate best performance.}
\label{tab:main-results}
\begin{tabular}{lccccccc}
\toprule
\multirow{3}{*}{\textbf{Model}} & \multirow{3}{*}{\textbf{STE $\downarrow$}} & \multicolumn{6}{c}{\textbf{RMS $\uparrow$}} \\
\cmidrule(lr){3-8}
 & & \multicolumn{3}{c}{\textbf{With Modifiers}} & \multicolumn{3}{c}{\textbf{Without Modifiers}} \\
 & & \textbf{Precision} & \textbf{Recall} & \textbf{F1} & \textbf{Precision} & \textbf{Recall} & \textbf{F1} \\
\midrule
Gemini-2.5-Flash & 0.4181 & 0.1527 & 0.1071 & 0.1217 & 0.2399 & 0.1839 & 0.2005 \\
GPT-4.1-mini & 0.6007 & 0.1516 & 0.1253 & 0.1320 & 0.2530 & 0.2313 & 0.2322 \\
Claude-3.5-Haiku & 0.6281 & 0.0858 & 0.0421 & 0.0530 & 0.1454 & 0.0805 & 0.0987 \\
\midrule
Gemini-2.5-Pro & \textbf{0.3212} & \textbf{0.2138} & 0.1664 & \textbf{0.1817} & \textbf{0.3781} & \textbf{0.3219} & \textbf{0.3383} \\
GPT-4.1 & 0.4611 & 0.2067 & 0.1597 & 0.1740 & 0.3517 & 0.2888 & 0.3038 \\
Claude-3.7-Sonnet & 0.3615 & 0.1780 & \textbf{0.1698} & 0.1688 & 0.3160 & 0.3170 & 0.3047 \\
\bottomrule
\end{tabular}
\end{table}

\subsection{Evaluation Metrics}

We developed three complementary metrics to evaluate how well an agent recovers the underlying SBML model. Our evaluation is designed to balance structural and dynamical similarity. However, it should be noted that these metrics do not capture all aspects of SBML equivalence. For example, we do not explicitly evaluate whether the proposed kinetic laws match the ground truth formulations. For formal definitions of these metrics, see \Cref{apx:reaction-recovery}.

\textbf{\topologymetric.} To assess success in recovering the correct system topology, we measure pairwise interactions between all species in the system and calculate F1 score. If identical relationships between the same species appear in multiple reactions, we count them only once.

\textbf{\reactionmetric.} We evaluate the agent's ability to recover the structures of ground truth reactions. We consider two reactions "matched" if they have identical sets of reactants and products. In the second, more stringent version, we additionally require matching modifiers.

\textbf{\simulationmetric.} This metric evaluates how accurately the predicted system matches the dynamic behavior of the true system by comparing time-series trajectories of each species. We employ the Symmetric Mean Absolute Percentage Error (SMAPE)~\citep{makridakis1993accuracy}, which normalizes errors relative to magnitude.

\begin{table}[t]
\caption{\textbf{All models struggle significantly more with identifying modifier relationships compared to reactant-product.} We report network topolgy performance (NTS precision and recall) across different types of edges in reaction networks.}
\label{tab:edge-comparison}
\label{tab:components-comparison}
\begin{tabular}{clcccccc}
\toprule
\textbf{Metric} & \textbf{Relationship} & \textbf{Gemi-M} & \textbf{Gemi-P} & \textbf{GPT-M} & \textbf{GPT-P} & \textbf{Clau-M} & \textbf{Clau-P} \\
\midrule
\multirow{3}{*}{\textbf{Pre}} & Reactant-Product & \textbf{0.453} & \textbf{0.448} & \textbf{0.415} & \textbf{0.371} & \textbf{0.385} & \textbf{0.442} \\
& Reactant-Modifier & 0.005 & 0.075 & 0.010 & 0.000 & 0.037 & 0.068 \\
& Modifier-Product & 0.000 & 0.106 & 0.014 & 0.000 & 0.054 & 0.119 \\
\midrule
\multirow{3}{*}{\textbf{Rec}} & Reactant-Product & \textbf{0.362} & \textbf{0.346} & \textbf{0.340} & \textbf{0.305} & \textbf{0.214} & \textbf{0.435} \\
& Reactant-Modifier & 0.001 & 0.032 & 0.002 & 0.000 & 0.016 & 0.042 \\
& Modifier-Product & 0.000 & 0.055 & 0.007 & 0.000 & 0.021 & 0.088 \\
\bottomrule
\vspace{-0.5cm}
\end{tabular}
\end{table}

\section{Related Work}

\textbf{AI Scientist Benchmarks.}
Most existing benchmarks have either focused on evaluating a single skill in the scientific discovery pipeline~\citep{labbench2024, aviary_2024} or the one-shot ability of LLMs to solve scientific tasks~\citep{Newsham_Kovačević_Moulange_Ke_Mukherjee_2025}. To the best of our knowledge, no previous work evaluating the scientific method end-to-end has effectively challenged LLMs to perform open-ended biological discovery research using experimental perturbations. DiscoveryWorld~\citep{discoveryworld} evaluates protein outlier detection and chemical optimization in a fixed, observed dataset. That is, unlike \ourbench{}, DiscoveryWorld does not evaluate the LLM's ability to discover the underlying mechanisms of action in a system ---it only asks the agent to reproduce the observed data. Moreover, the evaluation pipeline used in BioDiscoveryAgent~\citep{biodiscoveryagent2024} only deals with genetic perturbations from a fixed, predefined dataset, rather than a general model that can evaluate open-ended hypothesis from the agent. Finally, there exist other benchmarks exploring the discovery of equations~\citep{Shojaee_Nguyen_Meidani_Farimani_Doan_Reddy_2025, Ma_Wang_Guo_Sun_Tenenbaum_Rus_Gan_Matusik_2024, Romera_Paredes_funsearch} and chemicals~\citep{Sprueill_Edwards_Agarwal_Olarte_Sanyal_Johnston_Liu_Ji_Choudhury_2024} with LLMs and simulated data. This line of work does not allow the agent to perturb the system and feedback is given through fixed reward functions, a fundamentally different pipeline from the open-ended setting we explore in \ourbench{}.

\textbf{Causal Discovery Benchmarks.}
Causal discovery seeks to identify cause-effect relationships between variables from observational or, when possible, interventional data~\citep{tian2013causaldiscoverychanges}. Many recent efforts have applied LLMs to causal discovery tasks with varying degrees of success~\citep{long2023causaldiscoverylanguagemodels, jiralerspong2024efficientcausalgraphdiscovery}. This has motivated the development of causal reasoning benchmarks across a wide-range of scientific domains including gene regulatory networks~\citep{chevalley2023causalbenchlargescalebenchmarknetwork}, clinical disease modeling~\citep{abdulaal2024causal}, and causal question answering for coding and math~\citep{wang-2024-causalbench}. \ourbench{} differs from existing causality benchmarks in two main ways: 1) \ourbench{} requires active planning and design of intervention experiments rather than passive analysis of a static dataset or context; 2) \ourbench{} demands a more comprehensive skill-set to succeed, requiring agents to combine domain knowledge, data analysis skills, and coding ability to effectively discover biological mechanisms ---not just their structural relationships.

\paragraph{Simulations in AI for Science.} When the cost of obtaining experimental data in the real world is prohibitively expensive, simulations offer a cheaper alternative, albeit at the risk of a simulation-to-reality gap. In the biochemical sciences, simulation tools such as molecular dynamics~\citep{Hollingsworth2018MolecularDS} and quantum computational chemistry~\citep{McArdle_2020} have provided valuable datasets leading to significant advances in drug discovery~\citep{Kitchen2004DockingAS} and molecular modeling~\citep{Ramakrishnan2014QuantumCS}. Furthermore, many scientific discovery benchmarks incorporate simulations into their workflow to evaluate AI systems without the constraints of physical experimentation. ChemReasoner~\citep{sprueill2024chemreasonerheuristicsearchlarge} provides quantum-chemical feedback to discover active catalysts while ~\cite{Ma_Wang_Guo_Sun_Tenenbaum_Rus_Gan_Matusik_2024} use differentiable simulation to fit the coordinates of LLM-proposed molecules. Despite these advances, most simulations in AI benchmarks fail to capture the iterative, open-ended nature of real-world scientific discovery. Our benchmark addresses this gap by utilizing structured SBML documents that are manually curated with experimental support to create the drylab-in-the-loop.

\section{Experiments}
\label{sec:experiments}

\begin{figure}[t]
    \centering
    \includegraphics[width=\linewidth]{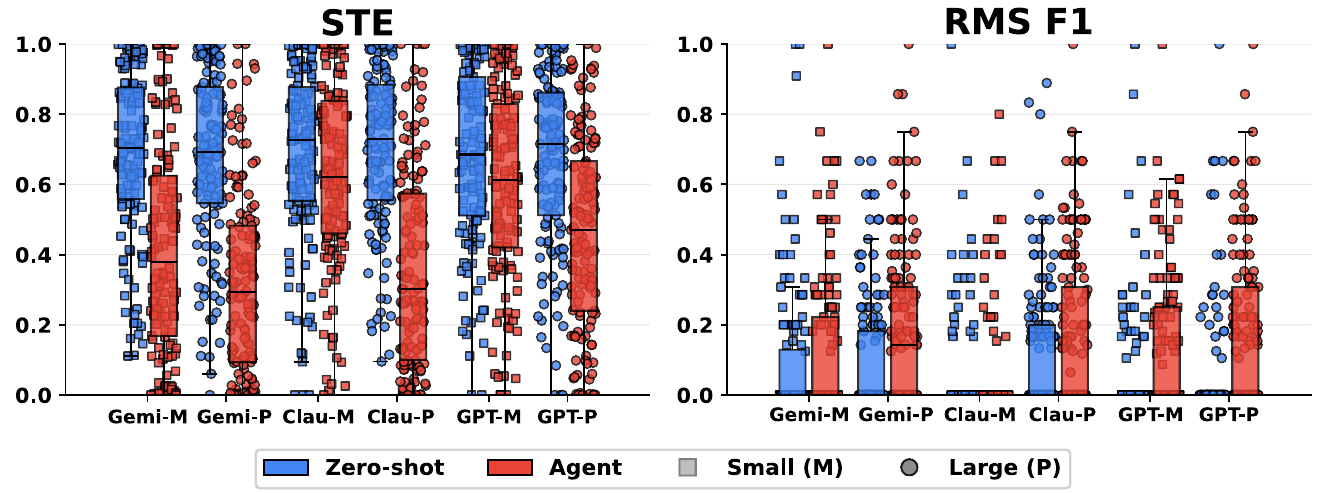}
    \caption{\textbf{All agents outperform their respective zero-shot baselines.} We compare each agent's performance with zero-shot prompting. Each marker represents performance on one system, with circles representing pro models and squares representing mini models. Blue shows zero-shot performance while red shows agent performance.}
    \label{fig:zero-shot}
    \vspace{-0.5cm}
\end{figure}

We evaluated six models from three frontier model families, each represented by their professional and mini variants: Anthropic's Claude-3.7-Sonnet-20250219 (P) and Claude-3.5-Haiku-20241022 (M); Google's Gemini-2.5-Pro-Preview-03-25 (P) and Gemini-2.5-Flash-Preview-04-17 (M); and OpenAI's GPT-4.1-2025-04-14 (P) and GPT-4.1-Mini-2025-04-14 (M). These variants are categorized based on their API pricing tiers. Due to computational resource constraints, we restricted our evaluation to \ourbench-small, with a total of 137 tasks.

\textbf{Perturbations.} For our main experiments, we permitted one type of experimental perturbation: changing the initial concentration of a specified species to a designated amount. We selected this approach as it represents a common and cost-effective perturbation in biological research. For example, researchers might alter initial protein 
concentrations to observe downstream effects on a signaling pathway. This perturbation type, when strategically designed, provides substantial information about species relationships, reaction types, kinetic laws, and reaction rates.
In an ablation study, we evaluated a second perturbation type: species knockout, which completely removes a species from the system. 
We used this condition to assess whether models could improve performance when given more powerful experimental tools. The results are shown in Appendix \ref{sec:knockout}. This style of perturbation has two main limitations: 1) they may not always be biologically plausible. For example, scientists may not have the equipments to change initial concentrations of the requested species to the specified amount. 2) our perturbations cannot guarantee complete system recovery of every SBML model \citep{villaverde2016structural}. However, this mirrors real-world biological research where scientists often work with limited experimental tools.

\subsection{Results}

\begin{wrapfigure}[18]{r}
{0.55\textwidth}
\vspace{-0.5em}
\centering\includegraphics[width=\linewidth]{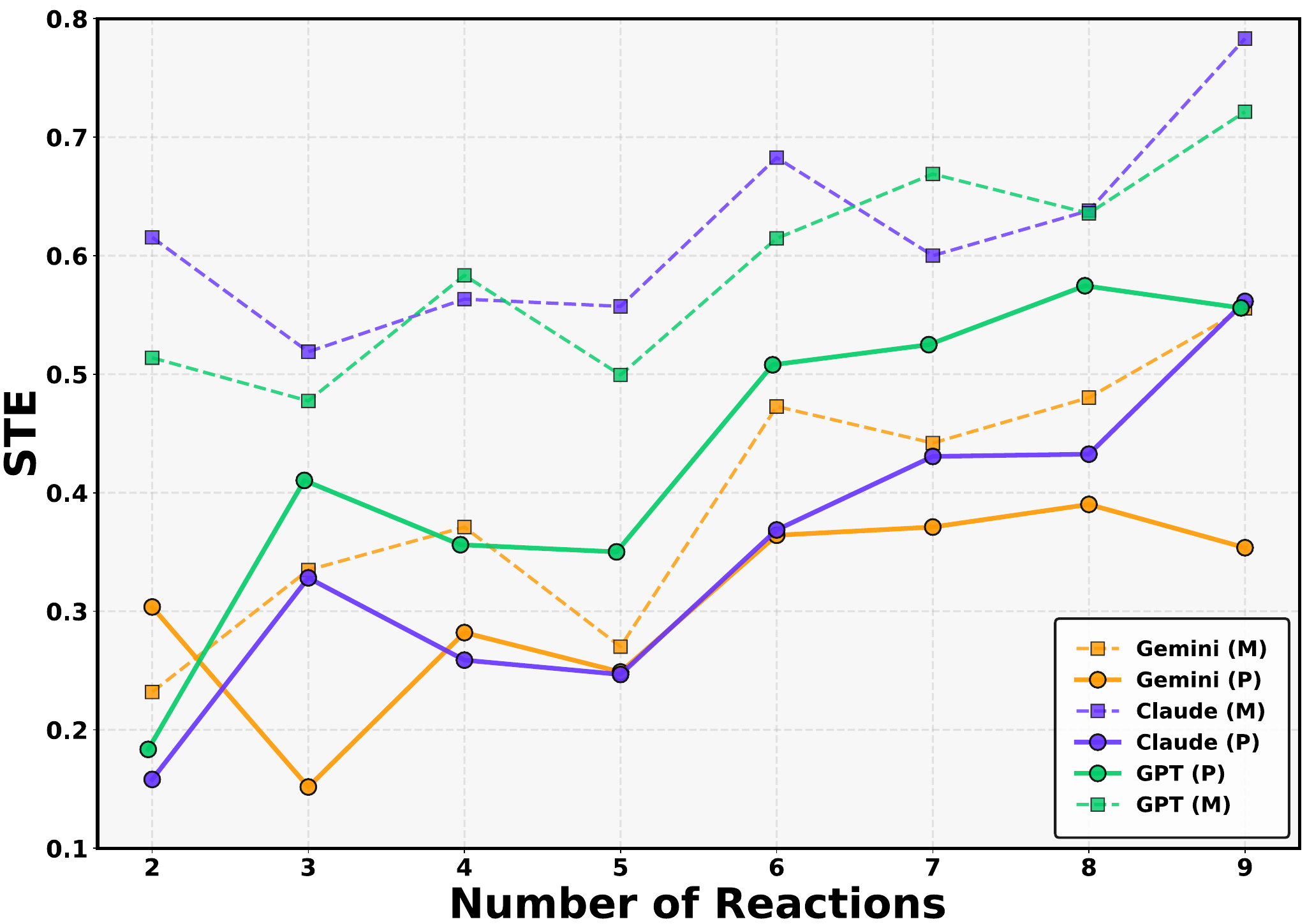}
    \caption{\textbf{Agents' performance decline as the underlying system's complexity grows.}}
    \label{fig:model-complexity}
\end{wrapfigure}

\textbf{Do models learn from experiments?} As an initial validation, we implemented a baseline direct prompting of the models. For fair comparison, we employed similar ReAct prompts across conditions, excluding only the experiment and tools components. We also permitted three additional debugging rounds in the baseline condition.
Results shown in Figure \ref{fig:zero-shot} demonstrate that all models achieved lower \simulation{} and higher \reaction{} when using the \ourbench{} framework compared to zero-shot prompting. This confirms that models are effectively extracting and incorporating information from experimental results during the scientific discovery process.

\textbf{Do pro models demonstrate superior scientific discovery capabilities?} Table \ref{tab:main-results} presents the average performance of all six models across all systems. Detailed results per system can be found in Section \ref{sec:detail-results}. Our results show that across all model families, pro variants consistently outperform their mini counterparts, with
both lower STE errors and higher RMS scores. This performance difference suggests that success in the scientific discovery process benefits from the enhanced capabilities of models. Among all models evaluated, Gemini-Pro achieved the best overall performance. Gemini-Mini also demonstrated superior performance among the mini models.

\textbf{How does model performance scale with the complexity of the underlying mechanism?} We plot the performance of each model on instances aggregated by the number of reactions and species in Figure \ref{fig:model-complexity}. We find that as system complexity grows, the simulation error consistently increases across all models tested. For example, both Claude-Sonnet and GPT-4.1 show error increases from close to 0.1 to 0.55 when the number of reactions increases from 2 to 10. This demonstrates that agents struggle increasingly with larger and more complex biological systems.

\textbf{How does performance vary across different types of species relationships?} We analyzed the models' ability to discover different types of relationships in biological systems. We categorized relationships based on the roles of connected species: reactant/product, reactant/modifier, and product/modifier connections. For each category, we computed average precision, recall, and F1 scores of NTS. Table \ref{tab:edge-comparison} shows that all models performed substantially better at discovering reactant-product relationships compared to relationships involving modifiers. Claude-Sonnet achieved the highest F1 score across all relationship types, yet its score for reactant-product relationships was more than 5 times higher than for modifier-related connections. This performance gap is expected, as identifying modifiers requires more targeted experiments to test how specific species affect reaction rates rather than simply observing which species are consumed or produced.

\begin{wrapfigure}[19]{l}{0.55\textwidth}
\centering\includegraphics[width=1.0\linewidth]{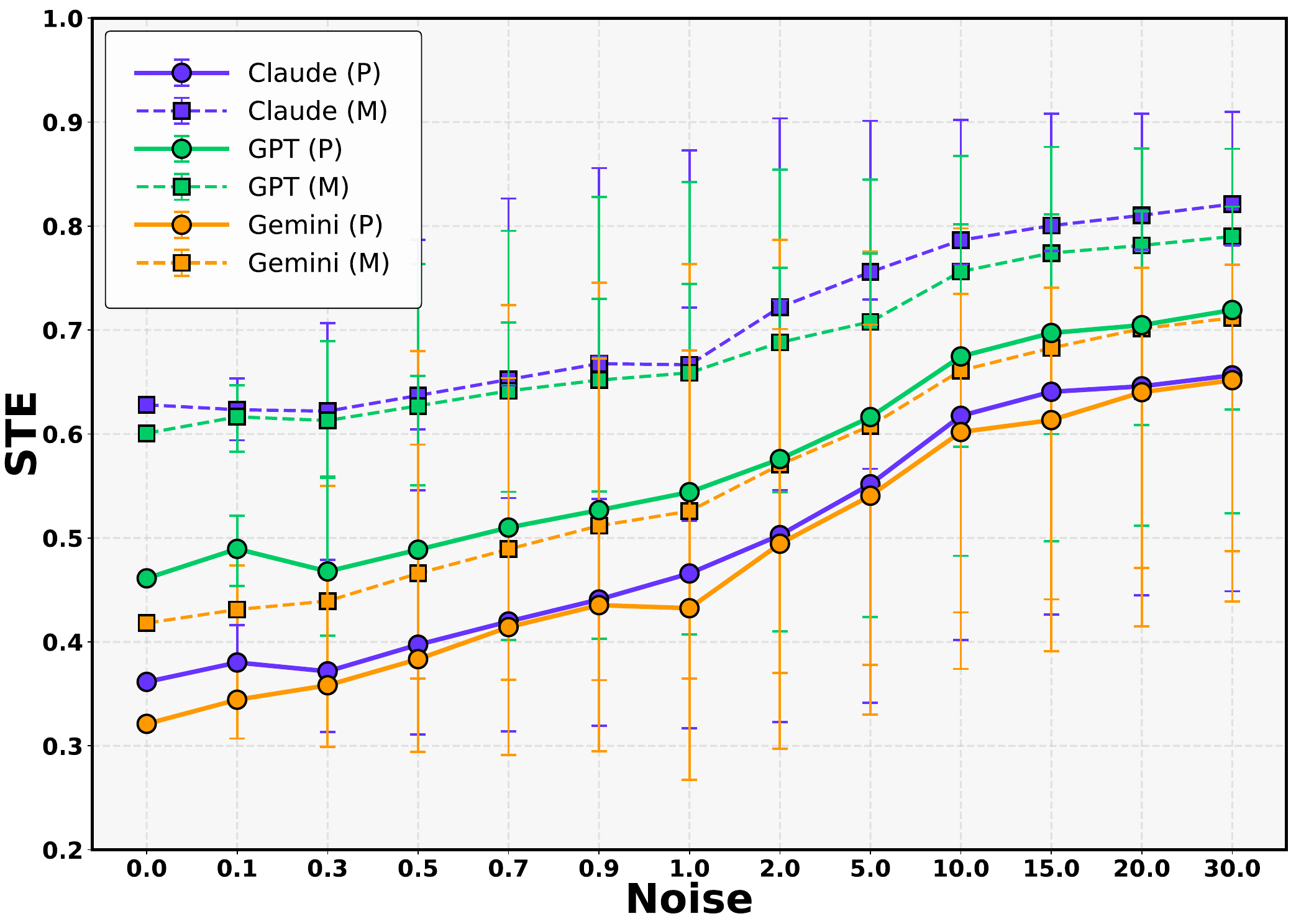}
    \vspace{-1em}
    \caption{\textbf{Agents' proposed systems are sensitive to perturbing initial conditions.}}
    \label{fig:perturbation-noise}
\vspace{-0.75em}
\end{wrapfigure} 

\textbf{How robust are the proposed systems?} We conducted analyses to determine whether agents' proposed SBML models generalize to different initial conditions or overfit to the specific experimental data. We added Gaussian noise (mean zero, variance proportional to the original concentration) to the initial concentrations. We then compared the STE between proposed models and the true system under these perturbed conditions. Figure \ref{fig:perturbation-noise} shows that STE increases consistently with noise levels across all models. This degradation in performance under perturbations indicates that agents' proposed mechanisms overfit to the specific experimental data to some degree rather than capturing the underlying biological system's fundamental properties.

\section{Conclusion and Limitations}
We introduced \ourbench{}, a benchmark that evaluates an LLM's scientific discovery ability through a complete discovery cycle, with emphasis on experiment design and analysis. Our work addressed the challenge of collecting experimental data by using formal biology models constructed by biologists to create a dry lab in the loop. These models represent established knowledge of biological systems with precise mathematical formulations of reactions, species interactions, and kinetics. In creating \ourbench{} we identified the following limitations: 

\textbf{Simulation to Reality Gap.} Although the SBMLs in \ourbench{} are peer-reviewed models of real-world biological mechanisms, many simplifications are made in order to transform these into systems that are possible to represent with an ODE. Therefore, conclusions derived from \ourbench{} may be systematically biased in subtle ways and not fully generalize to real-world settings. There is also \textbf{Lack of realistic noise.} The ODE-based SBMLs we selected for \ourbench{} gave us a realistic level of complexity, but they do not contain any noise in the experimental simulations. As such, we are not able to measure the effect of noise level on LLM's ability to recover the underlying system. In a similar vein, our simulations always provide comprehensive time-series data with many sampled timepoints, and we did not study how measurement sparsity impacts performance.

\textbf{Simplification of SBML Representations.} In the curation pipeline of \ourbench{}, we filter out SBML models with more complex components like \texttt{rules} and \texttt{events}. This restricts the set of biological systems we can evaluate and possible experiment perturbations we can design. \ourbench{} does not make full use of the richness of the SBML standard. For example, we are not specifically interrogating the \texttt{kineticLaw} equations when comparing \texttt{reaction}s. Thus, there is significant scope to increase the richness and impact of \ourbench{}.

Despite these limitations, we believe that \ourbench{} realizes an interesting new direction for AI scientist benchmarks. To our knowledge, \ourbench{} is the first benchmark to evaluate LLMs on the full cycle of scientific experimentation, and makes possible the study of agentic scientific decision-making.  
\ourbench{} is also a framework which can improve over time. As systems biologists continue to make discoveries in their domain and add to the BioModels database, \ourbench{} will also become more robust and potentially expand to cover more, different, experimental modalities. For experimental protocols that can be precisely specified and simulated, \ourbench{} can easily be extended.

Our benchmarking results on six frontier LLMs are immediately relevant to LLM-based autonomous experiment planning in self-driving labs, where selecting the right AI scientist to put in-the-loop has the potential to enhance search efficiency, lower operational costs, and accelerate scientific discovery. Looking forward, our framework produces artifacts: logs from best performer agents could be used to improve future agent's performance. Examples of successful strategies may aid reasoning by chain-of-thought prompting \cite{wei_chain--thought_2023} or be useful data to improve agents via finetuning \cite{grattafiori_llama_2024}. We hope that our work introduces a valuable testbed that can continue to evolve to better capture the complexities of real scientific discovery.

\bibliography{neurips_2025}

\begin{thebibliography}{39}
\providecommand{\natexlab}[1]{#1}
\providecommand{\url}[1]{\texttt{#1}}
\expandafter\ifx\csname urlstyle\endcsname\relax
  \providecommand{\doi}[1]{doi: #1}\else
  \providecommand{\doi}{doi: \begingroup \urlstyle{rm}\Url}\fi

\bibitem[Abdulaal et~al.(2024)Abdulaal, adamos hadjivasiliou, Montana-Brown, He, Ijishakin, Drobnjak, Castro, and Alexander]{abdulaal2024causal}
Ahmed Abdulaal, adamos hadjivasiliou, Nina Montana-Brown, Tiantian He, Ayodeji Ijishakin, Ivana Drobnjak, Daniel~C. Castro, and Daniel~C. Alexander.
\newblock Causal modelling agents: Causal graph discovery through synergising metadata- and data-driven reasoning.
\newblock In \emph{The Twelfth International Conference on Learning Representations}, 2024.
\newblock URL \url{https://openreview.net/forum?id=pAoqRlTBtY}.

\bibitem[Abolhasani \& Kumacheva(2023)Abolhasani and Kumacheva]{abolhasani2023rise}
Milad Abolhasani and Eugenia Kumacheva.
\newblock The rise of self-driving labs in chemical and materials sciences.
\newblock \emph{Nature Synthesis}, 2\penalty0 (6):\penalty0 483--492, 2023.

\bibitem[Ashburner et~al.(2000)Ashburner, Ball, Blake, Botstein, Butler, Cherry, Davis, Dolinski, Dwight, Eppig, Harris, Hill, Issel-Tarver, Kasarskis, Lewis, Matese, Richardson, Martin, Ringwald, Rubin, and Sherlock]{Ashburner2000GeneOT}
Michael Ashburner, Catherine~A. Ball, Judith~A. Blake, David Botstein, Heather~L. Butler, J.~Michael Cherry, Allan~Peter Davis, Kara Dolinski, Selina~S. Dwight, Janan~T. Eppig, Midori~A. Harris, David~P. Hill, Laurie Issel-Tarver, Andrew Kasarskis, Suzanna~E. Lewis, John~C. Matese, Joel~E. Richardson, Martin, Ringwald, Gerald~M. Rubin, and Gavin Sherlock.
\newblock Gene ontology: tool for the unification of biology.
\newblock \emph{Nature Genetics}, 25:\penalty0 25--29, 2000.
\newblock URL \url{https://api.semanticscholar.org/CorpusID:10718909}.

\bibitem[Chevalley et~al.(2023)Chevalley, Roohani, Mehrjou, Leskovec, and Schwab]{chevalley2023causalbenchlargescalebenchmarknetwork}
Mathieu Chevalley, Yusuf Roohani, Arash Mehrjou, Jure Leskovec, and Patrick Schwab.
\newblock Causalbench: A large-scale benchmark for network inference from single-cell perturbation data, 2023.
\newblock URL \url{https://arxiv.org/abs/2210.17283}.

\bibitem[Dixit et~al.(2016)Dixit, Parnas, Li, Chen, Fulco, Jerby-Arnon, Marjanovic, Dionne, Burks, Raychowdhury, et~al.]{dixit2016perturb}
Atray Dixit, Oren Parnas, Biyu Li, Jenny Chen, Charles~P Fulco, Livnat Jerby-Arnon, Nemanja~D Marjanovic, Danielle Dionne, Tyler Burks, Raktima Raychowdhury, et~al.
\newblock Perturb-seq: dissecting molecular circuits with scalable single-cell rna profiling of pooled genetic screens.
\newblock \emph{cell}, 167\penalty0 (7):\penalty0 1853--1866, 2016.

\bibitem[Dubitzky et~al.(2013)Dubitzky, Wolkenhauer, Yokota, and Cho]{dubitzky_encyclopedia_2013}
Werner Dubitzky, Olaf Wolkenhauer, Hiroki Yokota, and Kwang-Hyun Cho.
\newblock \emph{Encyclopedia of {Systems} {Biology}}.
\newblock Springer New York, New York, NY, 1st ed. 2013. edition, 2013.
\newblock ISBN 978-1-4419-9863-7.
\newblock \doi{10.1007/978-1-4419-9863-7}.

\bibitem[Grattafiori et~al.(2024)Grattafiori, Dubey, Jauhri, Pandey, Kadian, Al-Dahle, Letman, Mathur, Schelten, Vaughan, Yang, Fan, Goyal, Hartshorn, Yang, Mitra, Sravankumar, Korenev, Hinsvark, Rao, Zhang, Rodriguez, Gregerson, Spataru, Roziere, Biron, Tang, Chern, Caucheteux, Nayak, Bi, Marra, McConnell, Keller, Touret, Wu, Wong, Ferrer, Nikolaidis, Allonsius, Song, Pintz, Livshits, Wyatt, Esiobu, Choudhary, Mahajan, Garcia-Olano, Perino, Hupkes, Lakomkin, AlBadawy, Lobanova, Dinan, Smith, Radenovic, Guzmán, Zhang, Synnaeve, Lee, Anderson, Thattai, Nail, Mialon, Pang, Cucurell, Nguyen, Korevaar, Xu, Touvron, Zarov, Ibarra, Kloumann, Misra, Evtimov, Zhang, Copet, Lee, Geffert, Vranes, Park, Mahadeokar, Shah, Linde, Billock, Hong, Lee, Fu, Chi, Huang, Liu, Wang, Yu, Bitton, Spisak, Park, Rocca, Johnstun, Saxe, Jia, Alwala, Prasad, Upasani, Plawiak, Li, Heafield, Stone, El-Arini, Iyer, Malik, Chiu, Bhalla, Lakhotia, Rantala-Yeary, Maaten, Chen, Tan, Jenkins, Martin, Madaan, Malo, Blecher, Landzaat, Oliveira,
  Muzzi, Pasupuleti, Singh, Paluri, Kardas, Tsimpoukelli, Oldham, Rita, Pavlova, Kambadur, Lewis, Si, Singh, Hassan, Goyal, Torabi, Bashlykov, Bogoychev, Chatterji, Zhang, Duchenne, Çelebi, Alrassy, Zhang, Li, Vasic, Weng, Bhargava, Dubal, Krishnan, Koura, Xu, He, Dong, Srinivasan, Ganapathy, Calderer, Cabral, Stojnic, Raileanu, Maheswari, Girdhar, Patel, Sauvestre, Polidoro, Sumbaly, Taylor, Silva, Hou, Wang, Hosseini, Chennabasappa, Singh, Bell, Kim, Edunov, Nie, Narang, Raparthy, Shen, Wan, Bhosale, Zhang, Vandenhende, Batra, Whitman, Sootla, Collot, Gururangan, Borodinsky, Herman, Fowler, Sheasha, Georgiou, Scialom, Speckbacher, Mihaylov, Xiao, Karn, Goswami, Gupta, Ramanathan, Kerkez, Gonguet, Do, Vogeti, Albiero, Petrovic, Chu, Xiong, Fu, Meers, Martinet, Wang, Wang, Tan, Xia, Xie, Jia, Wang, Goldschlag, Gaur, Babaei, Wen, Song, Zhang, Li, Mao, Coudert, Yan, Chen, Papakipos, Singh, Srivastava, Jain, Kelsey, Shajnfeld, Gangidi, Victoria, Goldstand, Menon, Sharma, Boesenberg, Baevski, Feinstein, Kallet,
  Sangani, Teo, Yunus, Lupu, Alvarado, Caples, Gu, Ho, Poulton, Ryan, Ramchandani, Dong, Franco, Goyal, Saraf, Chowdhury, Gabriel, Bharambe, Eisenman, Yazdan, James, Maurer, Leonhardi, Huang, Loyd, Paola, Paranjape, Liu, Wu, Ni, Hancock, Wasti, Spence, Stojkovic, Gamido, Montalvo, Parker, Burton, Mejia, Liu, Wang, Kim, Zhou, Hu, Chu, Cai, Tindal, Feichtenhofer, Gao, Civin, Beaty, Kreymer, Li, Adkins, Xu, Testuggine, David, Parikh, Liskovich, Foss, Wang, Le, Holland, Dowling, Jamil, Montgomery, Presani, Hahn, Wood, Le, Brinkman, Arcaute, Dunbar, Smothers, Sun, Kreuk, Tian, Kokkinos, Ozgenel, Caggioni, Kanayet, Seide, Florez, Schwarz, Badeer, Swee, Halpern, Herman, Sizov, {Guangyi}, {Zhang}, Lakshminarayanan, Inan, Shojanazeri, Zou, Wang, Zha, Habeeb, Rudolph, Suk, Aspegren, Goldman, Zhan, Damlaj, Molybog, Tufanov, Leontiadis, Veliche, Gat, Weissman, Geboski, Kohli, Lam, Asher, Gaya, Marcus, Tang, Chan, Zhen, Reizenstein, Teboul, Zhong, Jin, Yang, Cummings, Carvill, Shepard, McPhie, Torres, Ginsburg, Wang, Wu,
  U, Saxena, Khandelwal, Zand, Matosich, Veeraraghavan, Michelena, Li, Jagadeesh, Huang, Chawla, Huang, Chen, Garg, A, Silva, Bell, Zhang, Guo, Yu, Moshkovich, Wehrstedt, Khabsa, Avalani, Bhatt, Mankus, Hasson, Lennie, Reso, Groshev, Naumov, Lathi, Keneally, Liu, Seltzer, Valko, Restrepo, Patel, Vyatskov, Samvelyan, Clark, Macey, Wang, Hermoso, Metanat, Rastegari, Bansal, Santhanam, Parks, White, Bawa, Singhal, Egebo, Usunier, Mehta, Laptev, Dong, Cheng, Chernoguz, Hart, Salpekar, Kalinli, Kent, Parekh, Saab, Balaji, Rittner, Bontrager, Roux, Dollar, Zvyagina, Ratanchandani, Yuvraj, Liang, Alao, Rodriguez, Ayub, Murthy, Nayani, Mitra, Parthasarathy, Li, Hogan, Battey, Wang, Howes, Rinott, Mehta, Siby, Bondu, Datta, Chugh, Hunt, Dhillon, Sidorov, Pan, Mahajan, Verma, Yamamoto, Ramaswamy, Lindsay, Lindsay, Feng, Lin, Zha, Patil, Shankar, Zhang, Zhang, Wang, Agarwal, Sajuyigbe, Chintala, Max, Chen, Kehoe, Satterfield, Govindaprasad, Gupta, Deng, Cho, Virk, Subramanian, Choudhury, Goldman, Remez, Glaser, Best,
  Koehler, Robinson, Li, Zhang, Matthews, Chou, Shaked, Vontimitta, Ajayi, Montanez, Mohan, Kumar, Mangla, Ionescu, Poenaru, Mihailescu, Ivanov, Li, Wang, Jiang, Bouaziz, Constable, Tang, Wu, Wang, Wu, Gao, Kleinman, Chen, Hu, Jia, Qi, Li, Zhang, Zhang, Adi, Nam, {Yu}, {Wang}, Zhao, Hao, Qian, Li, He, Rait, DeVito, Rosnbrick, Wen, Yang, Zhao, and Ma]{grattafiori_llama_2024}
Aaron Grattafiori, Abhimanyu Dubey, Abhinav Jauhri, Abhinav Pandey, Abhishek Kadian, Ahmad Al-Dahle, Aiesha Letman, Akhil Mathur, Alan Schelten, Alex Vaughan, Amy Yang, Angela Fan, Anirudh Goyal, Anthony Hartshorn, Aobo Yang, Archi Mitra, Archie Sravankumar, Artem Korenev, Arthur Hinsvark, Arun Rao, Aston Zhang, Aurelien Rodriguez, Austen Gregerson, Ava Spataru, Baptiste Roziere, Bethany Biron, Binh Tang, Bobbie Chern, Charlotte Caucheteux, Chaya Nayak, Chloe Bi, Chris Marra, Chris McConnell, Christian Keller, Christophe Touret, Chunyang Wu, Corinne Wong, Cristian~Canton Ferrer, Cyrus Nikolaidis, Damien Allonsius, Daniel Song, Danielle Pintz, Danny Livshits, Danny Wyatt, David Esiobu, Dhruv Choudhary, Dhruv Mahajan, Diego Garcia-Olano, Diego Perino, Dieuwke Hupkes, Egor Lakomkin, Ehab AlBadawy, Elina Lobanova, Emily Dinan, Eric~Michael Smith, Filip Radenovic, Francisco Guzmán, Frank Zhang, Gabriel Synnaeve, Gabrielle Lee, Georgia~Lewis Anderson, Govind Thattai, Graeme Nail, Gregoire Mialon, Guan Pang,
  Guillem Cucurell, Hailey Nguyen, Hannah Korevaar, Hu~Xu, Hugo Touvron, Iliyan Zarov, Imanol~Arrieta Ibarra, Isabel Kloumann, Ishan Misra, Ivan Evtimov, Jack Zhang, Jade Copet, Jaewon Lee, Jan Geffert, Jana Vranes, Jason Park, Jay Mahadeokar, Jeet Shah, Jelmer van~der Linde, Jennifer Billock, Jenny Hong, Jenya Lee, Jeremy Fu, Jianfeng Chi, Jianyu Huang, Jiawen Liu, Jie Wang, Jiecao Yu, Joanna Bitton, Joe Spisak, Jongsoo Park, Joseph Rocca, Joshua Johnstun, Joshua Saxe, Junteng Jia, Kalyan~Vasuden Alwala, Karthik Prasad, Kartikeya Upasani, Kate Plawiak, Ke~Li, Kenneth Heafield, Kevin Stone, Khalid El-Arini, Krithika Iyer, Kshitiz Malik, Kuenley Chiu, Kunal Bhalla, Kushal Lakhotia, Lauren Rantala-Yeary, Laurens van~der Maaten, Lawrence Chen, Liang Tan, Liz Jenkins, Louis Martin, Lovish Madaan, Lubo Malo, Lukas Blecher, Lukas Landzaat, Luke~de Oliveira, Madeline Muzzi, Mahesh Pasupuleti, Mannat Singh, Manohar Paluri, Marcin Kardas, Maria Tsimpoukelli, Mathew Oldham, Mathieu Rita, Maya Pavlova, Melanie Kambadur,
  Mike Lewis, Min Si, Mitesh~Kumar Singh, Mona Hassan, Naman Goyal, Narjes Torabi, Nikolay Bashlykov, Nikolay Bogoychev, Niladri Chatterji, Ning Zhang, Olivier Duchenne, Onur Çelebi, Patrick Alrassy, Pengchuan Zhang, Pengwei Li, Petar Vasic, Peter Weng, Prajjwal Bhargava, Pratik Dubal, Praveen Krishnan, Punit~Singh Koura, Puxin Xu, Qing He, Qingxiao Dong, Ragavan Srinivasan, Raj Ganapathy, Ramon Calderer, Ricardo~Silveira Cabral, Robert Stojnic, Roberta Raileanu, Rohan Maheswari, Rohit Girdhar, Rohit Patel, Romain Sauvestre, Ronnie Polidoro, Roshan Sumbaly, Ross Taylor, Ruan Silva, Rui Hou, Rui Wang, Saghar Hosseini, Sahana Chennabasappa, Sanjay Singh, Sean Bell, Seohyun~Sonia Kim, Sergey Edunov, Shaoliang Nie, Sharan Narang, Sharath Raparthy, Sheng Shen, Shengye Wan, Shruti Bhosale, Shun Zhang, Simon Vandenhende, Soumya Batra, Spencer Whitman, Sten Sootla, Stephane Collot, Suchin Gururangan, Sydney Borodinsky, Tamar Herman, Tara Fowler, Tarek Sheasha, Thomas Georgiou, Thomas Scialom, Tobias Speckbacher,
  Todor Mihaylov, Tong Xiao, Ujjwal Karn, Vedanuj Goswami, Vibhor Gupta, Vignesh Ramanathan, Viktor Kerkez, Vincent Gonguet, Virginie Do, Vish Vogeti, Vítor Albiero, Vladan Petrovic, Weiwei Chu, Wenhan Xiong, Wenyin Fu, Whitney Meers, Xavier Martinet, Xiaodong Wang, Xiaofang Wang, Xiaoqing~Ellen Tan, Xide Xia, Xinfeng Xie, Xuchao Jia, Xuewei Wang, Yaelle Goldschlag, Yashesh Gaur, Yasmine Babaei, Yi~Wen, Yiwen Song, Yuchen Zhang, Yue Li, Yuning Mao, Zacharie~Delpierre Coudert, Zheng Yan, Zhengxing Chen, Zoe Papakipos, Aaditya Singh, Aayushi Srivastava, Abha Jain, Adam Kelsey, Adam Shajnfeld, Adithya Gangidi, Adolfo Victoria, Ahuva Goldstand, Ajay Menon, Ajay Sharma, Alex Boesenberg, Alexei Baevski, Allie Feinstein, Amanda Kallet, Amit Sangani, Amos Teo, Anam Yunus, Andrei Lupu, Andres Alvarado, Andrew Caples, Andrew Gu, Andrew Ho, Andrew Poulton, Andrew Ryan, Ankit Ramchandani, Annie Dong, Annie Franco, Anuj Goyal, Aparajita Saraf, Arkabandhu Chowdhury, Ashley Gabriel, Ashwin Bharambe, Assaf Eisenman, Azadeh
  Yazdan, Beau James, Ben Maurer, Benjamin Leonhardi, Bernie Huang, Beth Loyd, Beto~De Paola, Bhargavi Paranjape, Bing Liu, Bo~Wu, Boyu Ni, Braden Hancock, Bram Wasti, Brandon Spence, Brani Stojkovic, Brian Gamido, Britt Montalvo, Carl Parker, Carly Burton, Catalina Mejia, Ce~Liu, Changhan Wang, Changkyu Kim, Chao Zhou, Chester Hu, Ching-Hsiang Chu, Chris Cai, Chris Tindal, Christoph Feichtenhofer, Cynthia Gao, Damon Civin, Dana Beaty, Daniel Kreymer, Daniel Li, David Adkins, David Xu, Davide Testuggine, Delia David, Devi Parikh, Diana Liskovich, Didem Foss, Dingkang Wang, Duc Le, Dustin Holland, Edward Dowling, Eissa Jamil, Elaine Montgomery, Eleonora Presani, Emily Hahn, Emily Wood, Eric-Tuan Le, Erik Brinkman, Esteban Arcaute, Evan Dunbar, Evan Smothers, Fei Sun, Felix Kreuk, Feng Tian, Filippos Kokkinos, Firat Ozgenel, Francesco Caggioni, Frank Kanayet, Frank Seide, Gabriela~Medina Florez, Gabriella Schwarz, Gada Badeer, Georgia Swee, Gil Halpern, Grant Herman, Grigory Sizov, {Guangyi}, {Zhang}, Guna
  Lakshminarayanan, Hakan Inan, Hamid Shojanazeri, Han Zou, Hannah Wang, Hanwen Zha, Haroun Habeeb, Harrison Rudolph, Helen Suk, Henry Aspegren, Hunter Goldman, Hongyuan Zhan, Ibrahim Damlaj, Igor Molybog, Igor Tufanov, Ilias Leontiadis, Irina-Elena Veliche, Itai Gat, Jake Weissman, James Geboski, James Kohli, Janice Lam, Japhet Asher, Jean-Baptiste Gaya, Jeff Marcus, Jeff Tang, Jennifer Chan, Jenny Zhen, Jeremy Reizenstein, Jeremy Teboul, Jessica Zhong, Jian Jin, Jingyi Yang, Joe Cummings, Jon Carvill, Jon Shepard, Jonathan McPhie, Jonathan Torres, Josh Ginsburg, Junjie Wang, Kai Wu, Kam~Hou U, Karan Saxena, Kartikay Khandelwal, Katayoun Zand, Kathy Matosich, Kaushik Veeraraghavan, Kelly Michelena, Keqian Li, Kiran Jagadeesh, Kun Huang, Kunal Chawla, Kyle Huang, Lailin Chen, Lakshya Garg, Lavender A, Leandro Silva, Lee Bell, Lei Zhang, Liangpeng Guo, Licheng Yu, Liron Moshkovich, Luca Wehrstedt, Madian Khabsa, Manav Avalani, Manish Bhatt, Martynas Mankus, Matan Hasson, Matthew Lennie, Matthias Reso, Maxim
  Groshev, Maxim Naumov, Maya Lathi, Meghan Keneally, Miao Liu, Michael~L. Seltzer, Michal Valko, Michelle Restrepo, Mihir Patel, Mik Vyatskov, Mikayel Samvelyan, Mike Clark, Mike Macey, Mike Wang, Miquel~Jubert Hermoso, Mo~Metanat, Mohammad Rastegari, Munish Bansal, Nandhini Santhanam, Natascha Parks, Natasha White, Navyata Bawa, Nayan Singhal, Nick Egebo, Nicolas Usunier, Nikhil Mehta, Nikolay~Pavlovich Laptev, Ning Dong, Norman Cheng, Oleg Chernoguz, Olivia Hart, Omkar Salpekar, Ozlem Kalinli, Parkin Kent, Parth Parekh, Paul Saab, Pavan Balaji, Pedro Rittner, Philip Bontrager, Pierre Roux, Piotr Dollar, Polina Zvyagina, Prashant Ratanchandani, Pritish Yuvraj, Qian Liang, Rachad Alao, Rachel Rodriguez, Rafi Ayub, Raghotham Murthy, Raghu Nayani, Rahul Mitra, Rangaprabhu Parthasarathy, Raymond Li, Rebekkah Hogan, Robin Battey, Rocky Wang, Russ Howes, Ruty Rinott, Sachin Mehta, Sachin Siby, Sai~Jayesh Bondu, Samyak Datta, Sara Chugh, Sara Hunt, Sargun Dhillon, Sasha Sidorov, Satadru Pan, Saurabh Mahajan,
  Saurabh Verma, Seiji Yamamoto, Sharadh Ramaswamy, Shaun Lindsay, Shaun Lindsay, Sheng Feng, Shenghao Lin, Shengxin~Cindy Zha, Shishir Patil, Shiva Shankar, Shuqiang Zhang, Shuqiang Zhang, Sinong Wang, Sneha Agarwal, Soji Sajuyigbe, Soumith Chintala, Stephanie Max, Stephen Chen, Steve Kehoe, Steve Satterfield, Sudarshan Govindaprasad, Sumit Gupta, Summer Deng, Sungmin Cho, Sunny Virk, Suraj Subramanian, Sy~Choudhury, Sydney Goldman, Tal Remez, Tamar Glaser, Tamara Best, Thilo Koehler, Thomas Robinson, Tianhe Li, Tianjun Zhang, Tim Matthews, Timothy Chou, Tzook Shaked, Varun Vontimitta, Victoria Ajayi, Victoria Montanez, Vijai Mohan, Vinay~Satish Kumar, Vishal Mangla, Vlad Ionescu, Vlad Poenaru, Vlad~Tiberiu Mihailescu, Vladimir Ivanov, Wei Li, Wenchen Wang, Wenwen Jiang, Wes Bouaziz, Will Constable, Xiaocheng Tang, Xiaojian Wu, Xiaolan Wang, Xilun Wu, Xinbo Gao, Yaniv Kleinman, Yanjun Chen, Ye~Hu, Ye~Jia, Ye~Qi, Yenda Li, Yilin Zhang, Ying Zhang, Yossi Adi, Youngjin Nam, {Yu}, {Wang}, Yu~Zhao, Yuchen Hao,
  Yundi Qian, Yunlu Li, Yuzi He, Zach Rait, Zachary DeVito, Zef Rosnbrick, Zhaoduo Wen, Zhenyu Yang, Zhiwei Zhao, and Zhiyu Ma.
\newblock The {Llama} 3 {Herd} of {Models}, November 2024.
\newblock URL \url{http://arxiv.org/abs/2407.21783}.
\newblock arXiv:2407.21783 [cs].

\bibitem[Hollingsworth \& Dror(2018)Hollingsworth and Dror]{Hollingsworth2018MolecularDS}
Scott~A. Hollingsworth and Ron~O. Dror.
\newblock Molecular dynamics simulation for all.
\newblock \emph{Neuron}, 99:\penalty0 1129--1143, 2018.
\newblock URL \url{https://api.semanticscholar.org/CorpusID:52311344}.

\bibitem[Hoops et~al.(2006)Hoops, Sahle, Gauges, Lee, Pahle, Simus, Singhal, Xu, Mendes, and Kummer]{hoops_copasi_2006}
Stefan Hoops, Sven Sahle, Ralph Gauges, Christine Lee, Jürgen Pahle, Natalia Simus, Mudita Singhal, Liang Xu, Pedro Mendes, and Ursula Kummer.
\newblock Copasi—a complex pathway simulator.
\newblock \emph{Bioinformatics}, 22\penalty0 (24):\penalty0 3067--3074, 10 2006.
\newblock ISSN 1367-4803.
\newblock \doi{10.1093/bioinformatics/btl485}.
\newblock URL \url{https://doi.org/10.1093/bioinformatics/btl485}.

\bibitem[Hucka et~al.(2019)Hucka, Bergmann, Chaouiya, Dräger, Hoops, Keating, König, Novère, Myers, Olivier, Sahle, Schaff, Sheriff, Smith, Waltemath, Wilkinson, and Zhang]{hucka_systems_2019}
Michael Hucka, Frank~T. Bergmann, Claudine Chaouiya, Andreas Dräger, Stefan Hoops, Sarah~M. Keating, Matthias König, Nicolas~Le Novère, Chris~J. Myers, Brett~G. Olivier, Sven Sahle, James~C. Schaff, Rahuman Sheriff, Lucian~P. Smith, Dagmar Waltemath, Darren~J. Wilkinson, and Fengkai Zhang.
\newblock The systems biology markup language (sbml): Language specification for level 3 version 2 core release 2.
\newblock \emph{Journal of Integrative Bioinformatics}, 16\penalty0 (2):\penalty0 20190021, 2019.
\newblock \doi{doi:10.1515/jib-2019-0021}.
\newblock URL \url{https://doi.org/10.1515/jib-2019-0021}.

\bibitem[Jansen et~al.(2024)Jansen, Côté, Khot, Bransom, Mishra, Majumder, Tafjord, and Clark]{discoveryworld}
Peter Jansen, Marc-Alexandre Côté, Tushar Khot, Erin Bransom, Bhavana~Dalvi Mishra, Bodhisattwa~Prasad Majumder, Oyvind Tafjord, and Peter Clark.
\newblock Discoveryworld: A virtual environment for developing and evaluating automated scientific discovery agents.
\newblock \penalty0 (arXiv:2406.06769arXiv:2406.06769), October 2024.
\newblock \doi{10.48550/arXiv.2406.06769}.
\newblock URL \url{http://arxiv.org/abs/2406.06769}.
\newblock arXiv:2406.06769 [cs].

\bibitem[Jimenez et~al.(2024)Jimenez, Yang, Wettig, Yao, Pei, Press, and Narasimhan]{jimenez2024swebench}
Carlos~E Jimenez, John Yang, Alexander Wettig, Shunyu Yao, Kexin Pei, Ofir Press, and Karthik~R Narasimhan.
\newblock {SWE}-bench: Can language models resolve real-world github issues?
\newblock In \emph{The Twelfth International Conference on Learning Representations}, 2024.
\newblock URL \url{https://openreview.net/forum?id=VTF8yNQM66}.

\bibitem[Jiralerspong et~al.(2024)Jiralerspong, Chen, More, Shah, and Bengio]{jiralerspong2024efficientcausalgraphdiscovery}
Thomas Jiralerspong, Xiaoyin Chen, Yash More, Vedant Shah, and Yoshua Bengio.
\newblock Efficient causal graph discovery using large language models, 2024.
\newblock URL \url{https://arxiv.org/abs/2402.01207}.

\bibitem[Kitchen et~al.(2004)Kitchen, Decornez, Furr, and Bajorath]{Kitchen2004DockingAS}
Douglas~B. Kitchen, Helene~Yvonne Decornez, Johnathan~R. Furr, and J{\"u}rgen Bajorath.
\newblock Docking and scoring in virtual screening for drug discovery: methods and applications.
\newblock \emph{Nature Reviews Drug Discovery}, 3:\penalty0 935--949, 2004.
\newblock URL \url{https://api.semanticscholar.org/CorpusID:1069493}.

\bibitem[Laurent et~al.(2024)Laurent, Janizek, Ruzo, Hinks, Hammerling, Narayanan, Ponnapati, White, and Rodriques]{labbench2024}
Jon~M. Laurent, Joseph~D. Janizek, Michael Ruzo, Michaela~M. Hinks, Michael~J. Hammerling, Siddharth Narayanan, Manvitha Ponnapati, Andrew~D. White, and Samuel~G. Rodriques.
\newblock Lab-bench: Measuring capabilities of language models for biology research.
\newblock \penalty0 (arXiv:2407.10362arXiv:2407.10362), July 2024.
\newblock \doi{10.48550/arXiv.2407.10362}.
\newblock URL \url{http://arxiv.org/abs/2407.10362}.
\newblock arXiv:2407.10362 [cs].

\bibitem[Long et~al.(2023)Long, Piché, Zantedeschi, Schuster, and Drouin]{long2023causaldiscoverylanguagemodels}
Stephanie Long, Alexandre Piché, Valentina Zantedeschi, Tibor Schuster, and Alexandre Drouin.
\newblock Causal discovery with language models as imperfect experts, 2023.
\newblock URL \url{https://arxiv.org/abs/2307.02390}.

\bibitem[Ma et~al.(2024)Ma, Wang, Guo, Sun, Tenenbaum, Rus, Gan, and Matusik]{Ma_Wang_Guo_Sun_Tenenbaum_Rus_Gan_Matusik_2024}
Pingchuan Ma, Tsun-Hsuan Wang, Minghao Guo, Zhiqing Sun, Joshua~B. Tenenbaum, Daniela Rus, Chuang Gan, and Wojciech Matusik.
\newblock Llm and simulation as bilevel optimizers: A new paradigm to advance physical scientific discovery, May 2024.
\newblock URL \url{https://arxiv.org/abs/2405.09783v1}.

\bibitem[Makridakis(1993)]{makridakis1993accuracy}
Spyros Makridakis.
\newblock Accuracy measures: theoretical and practical concerns.
\newblock \emph{International journal of forecasting}, 9\penalty0 (4):\penalty0 527--529, 1993.

\bibitem[Malik-Sheriff et~al.(2020)Malik-Sheriff, Glont, Nguyen, Tiwari, Roberts, Xavier, Vu, Men, Maire, Kananathan, et~al.]{malik2020biomodels}
Rahuman~S Malik-Sheriff, Mihai Glont, Tung~VN Nguyen, Krishna Tiwari, Matthew~G Roberts, Ashley Xavier, Manh~T Vu, Jinghao Men, Matthieu Maire, Sarubini Kananathan, et~al.
\newblock Biomodels—15 years of sharing computational models in life science.
\newblock \emph{Nucleic acids research}, 48\penalty0 (D1):\penalty0 D407--D415, 2020.

\bibitem[McArdle et~al.(2020)McArdle, Endo, Aspuru-Guzik, Benjamin, and Yuan]{McArdle_2020}
Sam McArdle, Suguru Endo, Alán Aspuru-Guzik, Simon~C. Benjamin, and Xiao Yuan.
\newblock Quantum computational chemistry.
\newblock \emph{Reviews of Modern Physics}, 92\penalty0 (1), March 2020.
\newblock ISSN 1539-0756.
\newblock \doi{10.1103/revmodphys.92.015003}.
\newblock URL \url{http://dx.doi.org/10.1103/RevModPhys.92.015003}.

\bibitem[Medley et~al.(2018)Medley, Choi, K{\"o}nig, Smith, Gu, Hellerstein, Sealfon, and Sauro]{Medley2018TelluriumNE}
J.~Kyle Medley, Kiri Choi, Matthias K{\"o}nig, Lucian~P. Smith, Stanley Gu, Joseph~L. Hellerstein, Stuart~C. Sealfon, and Herbert~M. Sauro.
\newblock Tellurium notebooks—an environment for reproducible dynamical modeling in systems biology.
\newblock \emph{PLoS Computational Biology}, 14, 2018.
\newblock URL \url{https://api.semanticscholar.org/CorpusID:49223654}.

\bibitem[Michaelis et~al.(2011)Michaelis, Menten, Johnson, and Goody]{Michaelis2011TheOM}
Leonor Michaelis, Maud~L. Menten, Kenneth~A. Johnson, and Roger~S. Goody.
\newblock The original michaelis constant: translation of the 1913 michaelis-menten paper.
\newblock \emph{Biochemistry}, 50 39:\penalty0 8264--9, 2011.
\newblock URL \url{https://doi.org/10.1021/bi201284u}.

\bibitem[Narayanan et~al.(2024)Narayanan, Braza, Griffiths, Ponnapati, Bou, Laurent, Kabeli, Wellawatte, Cox, Rodriques, and White]{aviary_2024}
Siddharth Narayanan, James~D. Braza, Ryan-Rhys Griffiths, Manu Ponnapati, Albert Bou, Jon Laurent, Ori Kabeli, Geemi Wellawatte, Sam Cox, Samuel~G. Rodriques, and Andrew~D. White.
\newblock Aviary: training language agents on challenging scientific tasks.
\newblock \penalty0 (arXiv:2412.21154arXiv:2412.21154), December 2024.
\newblock \doi{10.48550/arXiv.2412.21154}.
\newblock URL \url{http://arxiv.org/abs/2412.21154}.
\newblock arXiv:2412.21154 [cs].

\bibitem[Newsham et~al.(2025)Newsham, Kovačević, Moulange, Ke, and Mukherjee]{Newsham_Kovačević_Moulange_Ke_Mukherjee_2025}
Izzy Newsham, Luka Kovačević, Richard Moulange, Nan~Rosemary Ke, and Sach Mukherjee.
\newblock Large language models for zero-shot inference of causal structures in biology.
\newblock \penalty0 (arXiv:2503.04347arXiv:2503.04347), March 2025.
\newblock \doi{10.48550/arXiv.2503.04347}.
\newblock URL \url{http://arxiv.org/abs/2503.04347}.
\newblock arXiv:2503.04347 [cs].

\bibitem[Papin et~al.(2005)Papin, Hunter, Palsson, and Subramaniam]{Papin2005ReconstructionOC}
Jason~A. Papin, Tony Hunter, Bernhard~O. Palsson, and Shankar Subramaniam.
\newblock Reconstruction of cellular signalling networks and analysis of their properties.
\newblock \emph{Nature Reviews Molecular Cell Biology}, 6:\penalty0 99--111, 2005.
\newblock URL \url{https://doi.org/10.1038/nrm1570}.

\bibitem[Ramakrishnan et~al.(2014)Ramakrishnan, Dral, Dral, Rupp, and von Lilienfeld]{Ramakrishnan2014QuantumCS}
Raghunathan Ramakrishnan, Pavlo~O. Dral, Pavlo~O. Dral, Matthias Rupp, and O.~Anatole von Lilienfeld.
\newblock Quantum chemistry structures and properties of 134 kilo molecules.
\newblock \emph{Scientific Data}, 1, 2014.
\newblock URL \url{https://api.semanticscholar.org/CorpusID:15367821}.

\bibitem[Romera-Paredes et~al.(2024)Romera-Paredes, Barekatain, Novikov, Balog, Kumar, Dupont, Ruiz, Ellenberg, Wang, Fawzi, Kohli, and Fawzi]{Romera_Paredes_funsearch}
Bernardino Romera-Paredes, Mohammadamin Barekatain, Alexander Novikov, Matej Balog, M.~Pawan Kumar, Emilien Dupont, Francisco J.~R. Ruiz, Jordan~S. Ellenberg, Pengming Wang, Omar Fawzi, Pushmeet Kohli, and Alhussein Fawzi.
\newblock Mathematical discoveries from program search with large language models.
\newblock \emph{Nature}, 625\penalty0 (79957995):\penalty0 468–475, January 2024.
\newblock ISSN 1476-4687.
\newblock \doi{10.1038/s41586-023-06924-6}.

\bibitem[Roohani et~al.(2024)Roohani, Lee, Huang, Vora, Steinhart, Huang, Marson, Liang, and Leskovec]{biodiscoveryagent2024}
Yusuf Roohani, Andrew Lee, Qian Huang, Jian Vora, Zachary Steinhart, Kexin Huang, Alexander Marson, Percy Liang, and Jure Leskovec.
\newblock Biodiscoveryagent: An ai agent for designing genetic perturbation experiments.
\newblock \penalty0 (arXiv:2405.17631), October 2024.
\newblock \doi{10.48550/arXiv.2405.17631}.
\newblock URL \url{http://arxiv.org/abs/2405.17631}.
\newblock 4 citations (Semantic Scholar/DOI) [2025-02-18] arXiv:2405.17631 [cs].

\bibitem[Sauro(2020)]{sauro2020systems}
Herbert~M Sauro.
\newblock \emph{Systems biology: introduction to pathway modeling}.
\newblock Ambrosius Publishing, 2020.

\bibitem[SBML.org()]{noauthor_sbmlorg_nodate}
SBML.org.
\newblock {SBML}.org: {What} is {SBML}?
\newblock URL \url{https://sbml.org/documents/what-is-sbml/}.

\bibitem[Shojaee et~al.(2025)Shojaee, Nguyen, Meidani, Farimani, Doan, and Reddy]{Shojaee_Nguyen_Meidani_Farimani_Doan_Reddy_2025}
Parshin Shojaee, Ngoc-Hieu Nguyen, Kazem Meidani, Amir~Barati Farimani, Khoa~D. Doan, and Chandan~K. Reddy.
\newblock Llm-srbench: A new benchmark for scientific equation discovery with large language models.
\newblock \penalty0 (arXiv:2504.10415arXiv:2504.10415), April 2025.
\newblock \doi{10.48550/arXiv.2504.10415}.
\newblock URL \url{http://arxiv.org/abs/2504.10415}.
\newblock arXiv:2504.10415 [cs].

\bibitem[Somogyi et~al.(2015)Somogyi, Bouteiller, Glazier, K{\"o}nig, Medley, Swat, and Sauro]{somogyi2015libroadrunner}
Endre~T Somogyi, Jean-Marie Bouteiller, James~A Glazier, Matthias K{\"o}nig, J~Kyle Medley, Maciej~H Swat, and Herbert~M Sauro.
\newblock libroadrunner: a high performance sbml simulation and analysis library.
\newblock \emph{Bioinformatics}, 31\penalty0 (20):\penalty0 3315--3321, 2015.

\bibitem[Sprueill et~al.(2024{\natexlab{a}})Sprueill, Edwards, Agarwal, Olarte, Sanyal, Johnston, Liu, Ji, and Choudhury]{Sprueill_Edwards_Agarwal_Olarte_Sanyal_Johnston_Liu_Ji_Choudhury_2024}
Henry~W. Sprueill, Carl Edwards, Khushbu Agarwal, Mariefel~V. Olarte, Udishnu Sanyal, Conrad Johnston, Hongbin Liu, Heng Ji, and Sutanay Choudhury.
\newblock Chemreasoner: Heuristic search over a large language model’s knowledge space using quantum-chemical feedback.
\newblock \penalty0 (arXiv:2402.10980arXiv:2402.10980), December 2024{\natexlab{a}}.
\newblock \doi{10.48550/arXiv.2402.10980}.
\newblock URL \url{http://arxiv.org/abs/2402.10980}.
\newblock arXiv:2402.10980 [physics].

\bibitem[Sprueill et~al.(2024{\natexlab{b}})Sprueill, Edwards, Agarwal, Olarte, Sanyal, Johnston, Liu, Ji, and Choudhury]{sprueill2024chemreasonerheuristicsearchlarge}
Henry~W. Sprueill, Carl Edwards, Khushbu Agarwal, Mariefel~V. Olarte, Udishnu Sanyal, Conrad Johnston, Hongbin Liu, Heng Ji, and Sutanay Choudhury.
\newblock Chemreasoner: Heuristic search over a large language model's knowledge space using quantum-chemical feedback, 2024{\natexlab{b}}.
\newblock URL \url{https://arxiv.org/abs/2402.10980}.

\bibitem[Tian \& Pearl(2013)Tian and Pearl]{tian2013causaldiscoverychanges}
Jin Tian and Judea Pearl.
\newblock Causal discovery from changes, 2013.
\newblock URL \url{https://arxiv.org/abs/1301.2312}.

\bibitem[Villaverde et~al.(2016)Villaverde, Barreiro, and Papachristodoulou]{villaverde2016structural}
Alejandro~F Villaverde, Antonio Barreiro, and Antonis Papachristodoulou.
\newblock Structural identifiability of dynamic systems biology models.
\newblock \emph{PLoS computational biology}, 12\penalty0 (10):\penalty0 e1005153, 2016.

\bibitem[Wang(2024)]{wang-2024-causalbench}
Zeyu Wang.
\newblock {C}ausal{B}ench: A comprehensive benchmark for evaluating causal reasoning capabilities of large language models.
\newblock In Kam-Fai Wong, Min Zhang, Ruifeng Xu, Jing Li, Zhongyu Wei, Lin Gui, Bin Liang, and Runcong Zhao (eds.), \emph{Proceedings of the 10th SIGHAN Workshop on Chinese Language Processing (SIGHAN-10)}, pp.\  143--151, Bangkok, Thailand, August 2024. Association for Computational Linguistics.
\newblock URL \url{https://aclanthology.org/2024.sighan-1.17/}.

\bibitem[Wei et~al.(2023)Wei, Wang, Schuurmans, Bosma, Ichter, Xia, Chi, Le, and Zhou]{wei_chain--thought_2023}
Jason Wei, Xuezhi Wang, Dale Schuurmans, Maarten Bosma, Brian Ichter, Fei Xia, Ed~Chi, Quoc Le, and Denny Zhou.
\newblock Chain-of-{Thought} {Prompting} {Elicits} {Reasoning} in {Large} {Language} {Models}, January 2023.
\newblock URL \url{http://arxiv.org/abs/2201.11903}.
\newblock arXiv:2201.11903 [cs].

\bibitem[Yao et~al.(2023)Yao, Zhao, Yu, Du, Shafran, Narasimhan, and Cao]{yao2023react}
Shunyu Yao, Jeffrey Zhao, Dian Yu, Nan Du, Izhak Shafran, Karthik Narasimhan, and Yuan Cao.
\newblock React: Synergizing reasoning and acting in language models.
\newblock In \emph{International Conference on Learning Representations (ICLR)}, 2023.

\end{thebibliography}
\bibliographystyle{neurips_2025}

\newpage
\appendix

\renewcommand{\thefigure}{S\arabic{figure}}
\renewcommand{\thetable}{S\arabic{table}}
\setcounter{figure}{0}
\setcounter{table}{0}

\begin{center}
\large\bfseries Contents of Appendices
\end{center}

\startcontents

\printcontents{}{1}{\setcounter{tocdepth}{2}}

\renewcommand{\thesection}{\Alph{section}}
\setcounter{section}{0}

\lstdefinelanguage{XML}{
    morestring=[b]",
    morestring=[s]{>}{<},
    morecomment=[s]{<?}{?>},
    stringstyle=\color{black},
    identifierstyle=\color{blue},
    keywordstyle=\color{cyan},
    morekeywords={xmlns,version,type},
    showspaces=false,
    showstringspaces=false,
    showtabs=false,
    basicstyle=\small\ttfamily,
    columns=flexible,
    tabsize=2
}

\clearpage

\section{Agent Framework}
\subsection{System Prompts}\label{apx:system-prompt}
\begin{boxedmarkdown}
You are a biologist investigating a biological system. Your goal is to discover the biological mechanisms missing from your model by designing experiments and analyzing results. You must ultimately express your findings as a complete SBML model that accurately represents the biological system. Your final model will be evaluated based on how accurately it represents the true biological system.

Your final model will be evaluated by its similarity with the actual system under different perturbations, so discovering the true underlying mechanisms rather than overfitting to observed data is crucial.

# Action 

Each time, you can choose one of the three actions:

1. Request Experiments: You can request experiments to gather data from the true biological system you are studying. You can also choose to perturb the system and see how the system responds. This will help you better understand the mechanism of the system. 

2. Write Code: You have access to a Python environment to run analysis. You can use several scientific computing libraries and customized functions. Your code is executed as provided, so ensure your syntax is correct.

3. Submit the model: You can choose to submit the model and end the process if you think your hypothesis completely explains the mechanism. 


<!-- BEGIN EXPERIMENTAL_ACTIONS -->
<!-- END EXPERIMENTAL_ACTIONS -->

## Code Execution

For your code, print the results you want to see, and we will provide them for you. However, ensure your print content isn't too large, as large outputs will be truncated. For large variables like long arrays or dataframes, you can store them using the `shared_variables` and access them in future sessions:

* `shared_variables.add(variable_name, val)`: Store a variable for future access
* `shared_variables.access(variable_name)`: Retrieve a previously stored variable

### Libraries 

You are allowed to import the following libaries in your code: `numpy`, `pandas`, `math`, `scipy`, `sklearn`, `libsbml`

### Global variable access 

- input_sbml_string (str): Initial incomplete SBML model
- experiment_history (Dict[str, pd.DataFrame]): Time-series data for all experiments 
- shared_variables: Storage for all variables you've added in previous code executions

**Important** You can access these variables directly in your code. You can assume they are global variables provided to you. 

### Customized Functions

You can also call the follow functions in your code. 

<!-- BEGIN CUSTOMIZED_FUNCTIONS -->
<!-- END CUSTOMIZED_FUNCTIONS -->

## Add reactions using libsbml

```python
# Example of adding a reaction to an SBML model using libSBML
import libsbml

# Assuming we already have an SBML string loaded
sbml_doc = libsbml.readSBMLFromString(input_sbml_string)
model = sbml_doc.getModel()

# Create a new reaction
reaction = model.createReaction()
reaction.setId("reaction1")
reaction.setReversible(False)
reaction.setFast(False)  # Required in SBML Level 3

# Add a reactant
reactant = reaction.createReactant()
reactant.setSpecies("A")  # Species ID
reactant.setStoichiometry(1.0)
reactant.setConstant(False)  # Required in SBML Level 3

# Add a product
product = reaction.createProduct()
product.setSpecies("B")  # Species ID
product.setStoichiometry(1.0)
product.setConstant(True)  # Required in SBML Level 3

# Write the updated SBML
writer = libsbml.SBMLWriter()
updated_sbml = writer.writeSBMLToString(sbml_doc)
```

# Submit the model

If you want to submit the model and end the process, put your final model as a string variable called `final_sbml` in your python code. It is recommended using libsbml to modify `input_sbml_string` rather than write the entire xml on your own.

# Response Format

Your response should follow thought-action framework in markdown formats. You should have a thoughts section followed by an action section.


"""
## Thoughts
write down your thoughs here.

## Action

### Code
Include this if you want to write codes. Put your code in a python block. You can only include one code block in each response.
```python
import numpy as np
import pandas as pd
```

### Experiment
Include this if you want to request experiments. Put your experiment configuration in a json block. You can only include one json block in each response.
```json
{
    "action": "",
    "meta_data": {}
}
```

### Submit
Include this if you want to submit the model and end the process. Put your final model as a string variable called `final_sbml` in your python code.
```python
import libsbml
final_sbml=...
```
"""
\end{boxedmarkdown}

\subsection{Tool Manual}
\label{apx:tool-manual}
\begin{boxedmarkdown}
```python
def simulate(sbml_string: str) -> pd.DataFrame:
   """
   Simulates an SBML model and returns time series data.

   You can use this function to run simulations on your hypothesis model and compare it with the data gathered from the experiments.
   
   Args:
       sbml_string: an SBML model in xml format
       
   Returns:
       - A pandas dataframe of time series data for the given sbml models (with columns 'Time' and the species ID.)
   """
```   
\end{boxedmarkdown}

\subsection{Experiment Manual}
\label{sec:experiment-manual}
\begin{boxedmarkdown}
## Available Experiment Actions

### Observe
This experiment runs the system with default settings.

```json
{
   "action": "observe",
   "meta_data": {}
}
```

### change initial concentrations

This perturbation changes the initial concentrations of the given species. You cannot change the concentration of boundary and constant species. 

```json
{
    "action": "change_initial_concentration", 
    "meta_data": {
        "id_species1": 0.2, // Set the initial concentration of species id_species1 to 0.2.
        "id_species2": 0.5 
        // Only include the id of the species you want to modify. Any species not listed will keep their default values
    }
}
```
\end{boxedmarkdown}

\clearpage
\section{Evaluation Metrics}
In this section, we provide additional details on the evaluation metrics used in the paper.

\subsection{Network Topology Scores}\label{apx:network-topology}

For each SBML model, we construct a directed graph $G = (\mathcal{S}, \mathcal{E})$ where nodes $\mathcal{S}$ represent species and edges $\mathcal{E}$ represent direct relationships between species in reactions. An edge $(s_i, s_j) \in \mathcal{E}$ exists if species $s_i$ appears as a reactant and $s_j$ appears as a product in any reaction. %
This score focuses on the recovery of existence of pairwise interactions between species, without requiring all species in a reaction to be correctly identified.

\subsection{Reaction Recovery Scores}\label{apx:reaction-recovery}

The language model is shown all the species in the SBML file and is explicitly tasked with recovering deleted reactions. Thus, we can evaluate its performance using standard classification metrics if we define a binary function $f:R\times R \rightarrow\{0,1\}$ that indicates whether two reactions are equivalent. Formally, let $\gR_{\text{true}}=\{R_1,R_2,\dots,R_n\}$ denote the set of ground truth reactions and let $\gR_{\text{pred}}=\{R'_1,R'_2,\dots,R'_m\}$ denote the set of predicted reactions. Given an equivalence function $f$, we compute the true positive (TP) count, the false positive (FP) count, and the false negative (FN) count.

\begin{align*}
    TP&=\gR_{\text{pred}}\cap\gR_{\text{true}} = \{R'\in\gR_{\text{pred}} |\;\exists\, R\in\gR_{\text{true}} : f(R',R)=1\} \\
    FP&=\gR_{\text{pred}}-\gR_{\text{true}} = \{R'\in\gR_{\text{pred}} |\;\forall\, R \in \gR_{\text{true}}: f(R',R)=0\} \\
    FN&=\gR_{\text{true}}-\gR_{\text{pred}} = \{R'\in\gR_{\text{pred}}|\;\forall\, R'\in\gR_{\text{pred}}:f(R',R)=0\}
\end{align*}

Using these counts, we report the precision, recall, and F1 score as follows:

\begin{align*}
    \text{Precision} = \frac{|TP|}{|TP|+|FP|} \qquad \text{Recall} = \frac{|TP|}{|TP|+|FN|} \qquad \text{F1} = 2\cdot\frac{\text{Precision}\cdot\text{Recall}}{\text{Precision}+\text{Recall}}
\end{align*}

Note how RMS differs from NTS: NTS focuses on species relationships at a coarser level, awards partial credit for partial relationships, and does not account for multiple relationships between species.

\subsection{Symmetric Mean Absolute Percentage Error (SMAPE)}\label{apx:smape}

 For ground truth time-series $y$ and predicted series $\hat{y}$ of length $N$, SMAPE is calculated as $\text{SMAPE}(y, \hat{y}) = \frac{1}{N}\sum_{i=0}^{N - 1} \frac{|y_i - \hat{y}_i|}{|y| + |\hat{y}|}$. We then average SMAPE across all species. We evaluate models under both original and perturbed initial conditions to test whether they capture underlying mechanistic principles or merely overfit specific experiment data.

\clearpage
\section{Benchmark Curation Details}\label{apx:benchmark-curation}

\begin{figure}[h]
    \centering
    \includegraphics[width=\linewidth]{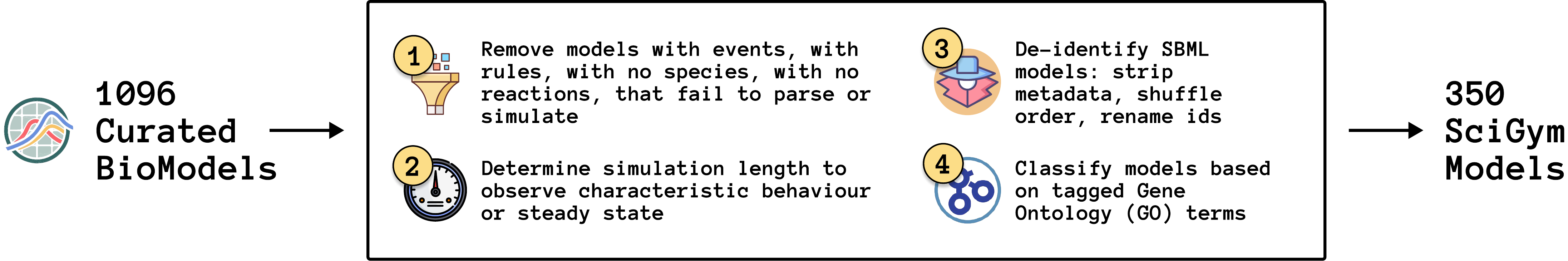}
    \caption{\ourbench{} benchmark curation pipeline}
    \label{fig:benchmark-curation-pipeline}
\end{figure}

\subsection{Filtering BioModels}\label{apx:filtering-sbml}

Among the 1,096 manually curated systems on BioModels, we filter out models that:

\begin{enumerate}
    \item Can not be parsed by \texttt{libsbml} (49)
    \item Can not be simulated with \texttt{Tellurium} (19)
    \item Have no reactions (106)
    \item Have no species (5)
    \item Have events (145)
    \item Have rules (422)
\end{enumerate}

This resulted in 350 SBML models out of the original 1,096 manually curated instances.

\begin{figure}[h!]
    \centering
    \includegraphics[width=\linewidth]{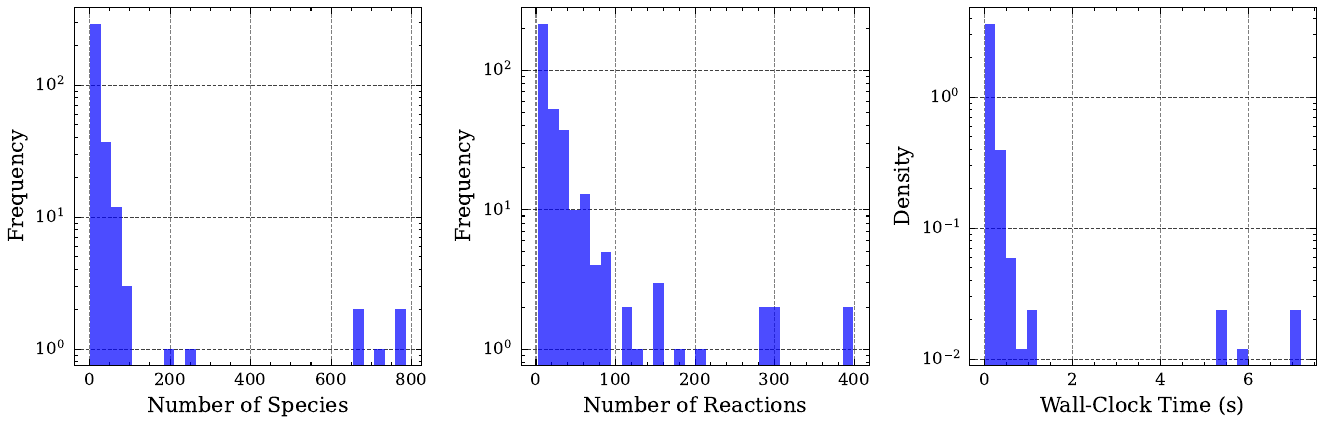}
    \caption{Number of species (left), reactions (middle), and simulation wall-clock time (right) for filtered BioModels.}
    \label{fig:filtered-sbml-stats}
\end{figure}

\subsection{De-identifying SBML files}\label{apx:sim-anon}
To prevent language models from memorizing the SBML files, we design a pre-processing pipeline that removes potentially compromising information from the models without affecting their functionality or simulation dynamics. This de-identification pipeline consists of three steps.

\textbf{1. Strip metadata.} We remove optional metadata fields from the SBML entities. These include \textit{metaid}, \textit{notes}, \textit{annotations}, \textit{model history}, \textit{control vocabulary}, \textit{dates}, \textit{author information}, and \textit{names}\footnote{Species names are kept such that biological entities in the system remain identifiable.}.

\textbf{2. Shuffle components.} We shuffle the order of parameters, reactions, species, and compartments.

\textbf{3. Renaming ids.} We rename the ids of all components in the model to a unique 4 character alphanumeric identifier.

\subsection{Determining Simulation Timescale}\label{apx:sim-timescale}

Each SBML instance in the BioModels database is accompanied by a SED-ML file that specifies the simulation setup in order to reproduce the curated model. However, 204 out of the 350 filtered models provided an auto-generated template SED-ML file with a default simulation duration of 10 seconds. We found this to be troublesome since some models require a longer simulation budget to observe interesting behavior or reach a steady state. Thus, we developed a pipeline to systematically compute a suitable simulation duration for every filtered BioModel. Our procedure consists of three stages.

\textbf{1. Steady-state analysis:} For each model, we first attempt to solve for the steady state of the ODE system (when every species has a rate of change less than $10^-6$) using the NLEQ2 algorithm. This step may fail for some models due to numerical issues or the absence of a steady state.

\textbf{2. Time-course simulation:} We then simulate the ODE system by integrating over time with the following specifications: We use the integrator defined in the original SED-ML file, defaulting to CVODE if missing. We use a fixed step size of $t=0.05$ seconds. We simulate until at least one of three termination criteria is met: 
\begin{itemize}
    \item A steady state is reached
    \item The integrator fails (typically due to stiffness issues or numerical instabilities)
    \item The maximum simulation budget of 10,000 seconds is exhausted
\end{itemize}

\textbf{3. Final duration determination:} We select the final simulation duration by taking the maximum value among the time required to reach steady state (if successfully solved in step 1), the end time of the time-course simulation, and the original end time specified in the SED-ML file.

\begin{figure}[h]
    \centering
    \includegraphics[width=\linewidth]{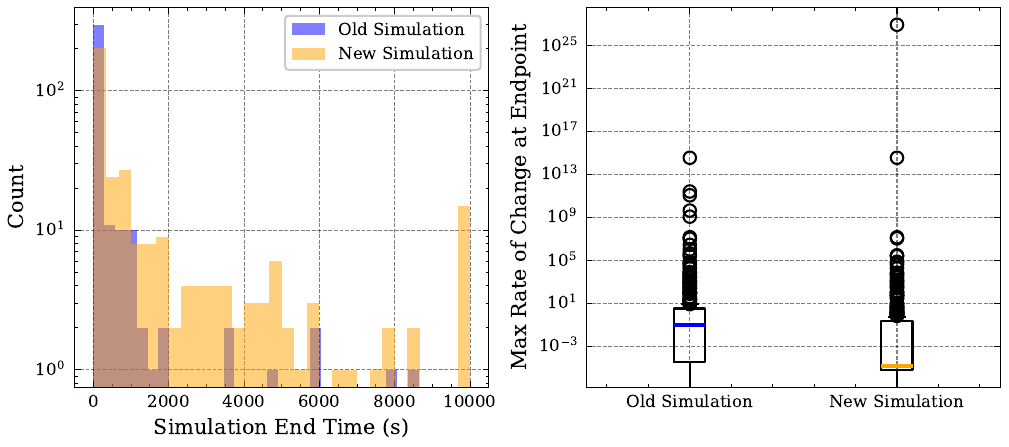}
    \caption{Simulation duration (left) and maximum species rate of change at endpoint (right) before and after processing.}
    \label{fig:simulation-timescale}
\end{figure}

This approach ensures that sufficient simulation time is allocated for each model to either reach steady state or exhibit its characteristic dynamic behavior, while still respecting any intentionally specified simulation parameters in the original SED-ML file. We have plotted the simulation duration and maximum species rate of change at simulation endpoint in \Cref{fig:simulation-timescale}.

\subsection{Creating Incomplete SBML Models}\label{apx:removing-reactions}

To create benchmark questions, the final step consists of removing reactions in the SBML files that the language model is tasked with recovering. Below, we describe this deletion operation in greater detail.

\textbf{Removing reactions}. To remove a reaction without leaving any traces, we edit the SBML file such that all references to the deleted reaction are also removed. This includes \textit{initial assignments}, \textit{function definitions}, \textit{constraints} that use the deleted reaction's rate to define its own declaration.

\textbf{Removing orphaned parameters.} After removing the reactions for each question, we make a final pass through the SBML file and remove orphaned parameters that are no longer referenced. This step ensures that we aren't leaking any unnecessary information to the language model.

To ensure that masked models remain valid SBML files, we only remove core SBML objects\footnote{We identify first level objects as the list of core SBML objects in the \texttt{libsbml} package.} that contain nested references to the deleted reaction. For example, if we find a constraint with a clause referencing a deleted reaction, we remove the entire constraint instead of removing only the problematic clause. This guarantees that components are never functionally modified in a way that might mutate their biological significance.

\subsection{Classifying Curated Models}

\begin{figure}[h!]
  \centering
  \includegraphics[width=\textwidth]{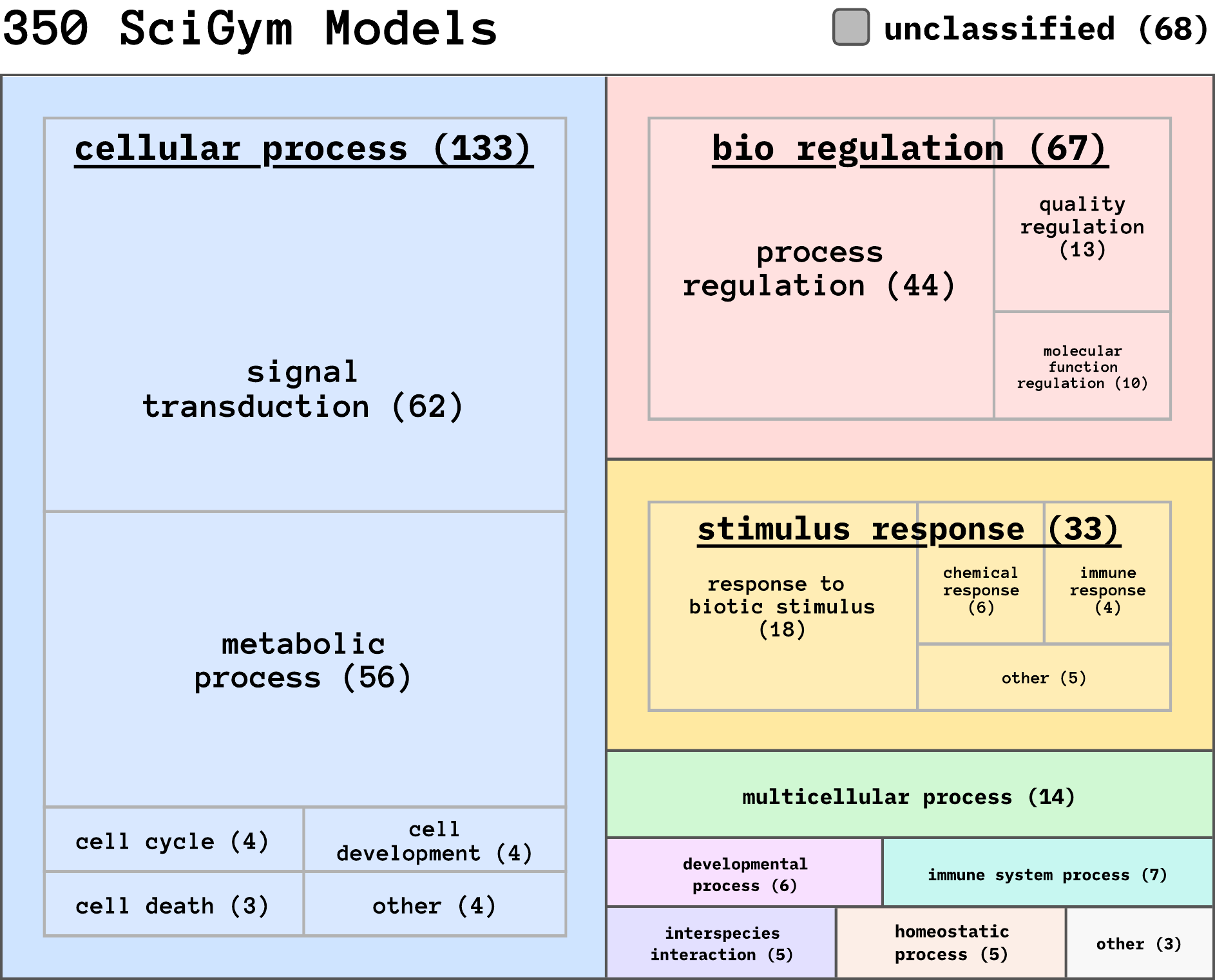}
  \caption{\textbf{\ourbench{} covers a wide range of biology processes}. Classification of \ourbench models by Gene Ontology (GO) terms.}
  \label{fig:go-ontology}
\end{figure}

To gain further insight into the types of biochemical networks that our benchmark contains, we perform an analysis of the gene ontology~\citep{Ashburner2000GeneOT} (GO) terms that are tagged by each SBML model that pass the filtration step. More precisely, for each GO term, we identify its ancestors and attempt to map it onto one of the 318 grandchildren of the \texttt{biological\_process} root term. Altogether, we were able to map 601 of the 765 filtered models to target GO terms. We provide a breakdown of these terms in \Cref{fig:go-ontology}.

\clearpage
\section{Additional Details on SBML}\label{sec:sbml-apx}

\subsection{Major SBML Components}
\label{sec:sbml-tags}

\begin{table}[hbt]\centering
\caption{SBML Element Definitions, reproduced from \cite{hucka_systems_2019} Level 3 Version 2 Core specification}
\vspace{5pt}
\label{tab:go-term-classification}
\small
\begin{NiceTabular}{l|p{10cm}}
\toprule[1.5pt]
    \textbf{SBML Element} & \textbf{Description} \\
    \hline
    Function & A named mathematical function that may be used throughout the rest of a model. \\
    \hline
    Unit  & A named definition of a new unit of measurement. Named units can be used in the expression of quantities in a model. \\
    \hline
    Compartment & A well-stirred container of finite size where species may be located. Compartments may or may not represent actual physical structures. \\
    \hline
    Species & A pool of entities of the same kind located in a compartment and participating in reactions (processes). In biochemical network models, common examples of species include ions, proteins and other molecules; however, in practice, an SBML species can be any kind of entity that makes sense in the context of a given model. \\
    \hline
    Parameter & A quantity with a symbolic name. In SBML, the term parameter is used in a generic sense to refer to named quantities regardless of whether they are constants or variables in a model. SBML Level 3 provides the ability to define parameters that are global to a model as well as parameters that are local to a single reaction. \\
    \hline
    Initial Assignment & A mathematical expression used to determine the initial conditions of a model. This type of object can only be used to define how the value of a symbol can be calculated from other values and symbols at the start of simulated time. \\
    \hline
    Rule & A mathematical expression added to the set of equations constructed based on the reactions defined in a model. Rules can be used to define how a symbol's value can be calculated from other symbols, or used to define the rate of change of a symbol. The set of rules in a model can be used with the reaction rate equations to determine the behavior of the model with respect to time. Rules constrain the model for the entire duration of simulated time. \\
    \hline
    Constraint & A means of detecting out-of-bounds conditions during a dynamical simulation and optionally issuing diagnostic messages. Constraints are defined by an arbitrary mathematical expression computing a true/false value from model symbols. An SBML constraint applies at all instants of simulated time; however, the set of constraints in model should not be used to determine the behavior of the model with respect to time. \\
    \hline
    Reaction & A statement describing some transformation, transport or binding process that can change the amount of one or more species. For example, a reaction may describe how certain entities (reactants) are transformed into certain other entities (products). Reactions have associated kinetic rate expressions describing how quickly they take place. \\
    \hline
    Event & A statement describing an instantaneous, discontinuous change in one or more symbols of any type (species, compartment, parameter, etc.) when a triggering condition is satisfied. \\
  \bottomrule[1.5pt]
    \end{NiceTabular}
    \end{table}
  \label{apx:sbml-defs}

\subsection{Example SBML Document of Enzyme Process}
\label{sec:sbml-example}

We provide in \Cref{lst:sbml-enzyme} the full SBML document of the enzymatic process introduced as an example in \Cref{sec:sbml-def} and used throughout the text.

\label{list:sbml_eg}
\begin{lstlisting}[language=XML,
                   caption={SBML file for Enzymatic Process},
                   label={lst:sbml-enzyme},
                   breaklines=true,
                   numbers=left,
                   frame=single]
<?xml version="1.0" encoding="UTF-8"?>
<sbml xmlns="http://www.sbml.org/sbml/level3/version2/core" level="3" version="2">
  <model timeUnits="second" extentUnits="mole">
    <listOfUnitDefinitions>
      <unitDefinition id="per_second">
        <listOfUnits>
          <unit kind="second" exponent="-1" scale="0" multiplier="1" />
        </listOfUnits>
      </unitDefinition>
      <unitDefinition id="litre_per_mole_sec">
        <listOfUnits>
          <unit kind="mole" exponent="-1" scale="0" multiplier="1" />
          <unit kind="litre" exponent="1" scale="0" multiplier="1" />
          <unit kind="second" exponent="-1" scale="0" multiplier="1" />
        </listOfUnits>
      </unitDefinition>
    </listOfUnitDefinitions>
    <listOfCompartments>
      <compartment id="comp" spatialDimensions="3" size="1e-14" units="litre" constant="true" />
    </listOfCompartments>
    <listOfSpecies>
      <species id="E" compartment="comp" initialAmount="5e-21" substanceUnits="mole"
        hasOnlySubstanceUnits="false" boundaryCondition="false" constant="false" />
      <species id="S" compartment="comp" initialAmount="1e-20" substanceUnits="mole"
        hasOnlySubstanceUnits="false" boundaryCondition="false" constant="false" />
      <species id="P" compartment="comp" initialAmount="0" substanceUnits="mole"
        hasOnlySubstanceUnits="false" boundaryCondition="false" constant="false" />
      <species id="ES" compartment="comp" initialAmount="0" substanceUnits="mole"
        hasOnlySubstanceUnits="false" boundaryCondition="false" constant="false" />
    </listOfSpecies>
    <listOfParameters>
      <parameter id="veq_koff" value="0.2" units="per_second" constant="true" />
      <parameter id="veq_kon" value="1e6" units="litre_per_mole_sec" constant="true" />
      <parameter id="vcat_kcat" value="0.1" units="per_second" constant="true" />
    </listOfParameters>
    <listOfReactions>
      <reaction id="veq" reversible="true">
        <listOfReactants>
          <speciesReference species="E" stoichiometry="1" constant="true" />
          <speciesReference species="S" stoichiometry="1" constant="true" />
        </listOfReactants>
        <listOfProducts>
          <speciesReference species="ES" stoichiometry="1" constant="true" />
        </listOfProducts>
        <kineticLaw>
          <math xmlns="http://www.w3.org/1998/Math/MathML">
            <apply>
              <times />
              <ci> comp </ci>
              <apply>
                <minus />
                <apply>
                  <times />
                  <ci> veq_kon </ci>
                  <ci> E </ci>
                  <ci> S </ci>
                </apply>
                <apply>
                  <times />
                  <ci> veq_koff </ci>
                  <ci> ES </ci>
                </apply>
              </apply>
            </apply>
          </math>
        </kineticLaw>
      </reaction>
      <reaction id="vcat" reversible="false">
        <listOfReactants>
          <speciesReference species="ES" stoichiometry="1" constant="true" />
        </listOfReactants>
        <listOfProducts>
          <speciesReference species="E" stoichiometry="1" constant="true" />
          <speciesReference species="P" stoichiometry="1" constant="true" />
        </listOfProducts>
        <kineticLaw>
          <math xmlns="http://www.w3.org/1998/Math/MathML">
            <apply>
              <times />
              <ci> comp </ci>
              <ci> vcat_kcat </ci>
              <ci> ES </ci>
            </apply>
          </math>
        </kineticLaw>
      </reaction>
    </listOfReactions>
  </model>
</sbml>
\end{lstlisting}

To translate this SBML model into an ODE, we sum the rates of change implied by the reactions for each species. The \texttt{kineticLaw} specifies the rate for each reaction as a function of the participating species. To calculate the differential equation for each species, we multiply its signed stoichiometric coefficient by the rate of change specified in the \texttt{kineticLaw}. The sign of the stoichiometric coefficient is -1 if the species is a reactant in the reaction, +1 if it is a product in the reaction, and 0 otherwise. 

For this example, we have
\begin{align}
    v\frac{d [E]}{dt} &= -(v k_{\text{on}}[\rmE][\rmS]- v k_{\text{off}}[\rmE\rmS]) + (v k_{\text{cat}}[\rmE\rmS])\\
    v\frac{d [S]}{dt} &= -(v k_{\text{on}}[\rmE][\rmS]- v k_{\text{off}}[\rmE\rmS])\\
    v\frac{d [ES]}{dt} &= (v k_{\text{on}}[\rmE][\rmS]- v k_{\text{off}}[\rmE\rmS]) - (v k_{\text{cat}}[\rmE\rmS])\\
    v\frac{d [P]}{dt} &= (v k_{\text{cat}}[\rmE\rmS])
\end{align}
where $v > 0$ is the volume of compartment \texttt{comp}.

\subsection{Example SBML Document with Modifier}
\label{sec:sbml-modifier-example}

We provide in \Cref{lst:sbml-modifier} a full SBML document that includes a modifier. Modifiers change kinetic laws but are not consumed or produced in the reaction.

\label{list:sbml_eg}
\begin{lstlisting}[language=XML,
                   caption={SBML example with a modifier},
                   label={lst:sbml-modifier},
                   breaklines=true,
                   numbers=left,
                   frame=single]
<?xml version="1.0" encoding="UTF-8"?>
<sbml xmlns="http://www.sbml.org/sbml/level3/version2/core" level="3" version="2">
  <model id="catalyzed_reaction_model">
    <listOfCompartments>
      <compartment id="v" spatialDimensions="3" size="1" constant="true"/>
    </listOfCompartments>
    <listOfSpecies>
      <species id="S1" compartment="v" initialAmount="10" hasOnlySubstanceUnits="true" boundaryCondition="false" constant="false"/>
      <species id="S2" compartment="v" initialAmount="0" hasOnlySubstanceUnits="true" boundaryCondition="false" constant="false"/>
      <species id="M" compartment="v" initialAmount="5" hasOnlySubstanceUnits="true" boundaryCondition="false" constant="false"/>
    </listOfSpecies>
    <listOfParameters>
      <parameter id="k1" value="0.1" constant="true"/>
    </listOfParameters>
    <listOfReactions>
      <reaction id="R1" reversible="false">
        <listOfReactants>
          <speciesReference species="S1" stoichiometry="1" constant="true"/>
        </listOfReactants>
        <listOfProducts>
          <speciesReference species="S2" stoichiometry="1" constant="true"/>
        </listOfProducts>
        <listOfModifiers>
          <modifierSpeciesReference species="M"/>
        </listOfModifiers>
        <kineticLaw>
          <math xmlns="http://www.w3.org/1998/Math/MathML">
            <apply>
              <times/>
              <ci> v </ci>
              <ci> k1 </ci>
              <ci> S1 </ci>
              <ci> M </ci>
            </apply>
          </math>
        </kineticLaw>
      </reaction>
    </listOfReactions>
  </model>
</sbml>
\end{lstlisting}

To translate this SBML model into an ODE, we sum the rates of change implied by the reactions for each species. The \texttt{kineticLaw} specifies the rate for each reaction as a function of the participating species. To calculate the differential equation for each species, we multiply its signed stoichiometric coefficient by the rate of change specified in the \texttt{kineticLaw}. The sign of the stoichiometric coefficient is -1 if the species is a reactant in the reaction, +1 if it is a product in the reaction, and 0 otherwise. 

For this example, we have
\begin{align}
    v\frac{d [\rmS1]}{dt} &= - v k_1[\rmS1][\rmM]\\
    v\frac{d [\rmS2]}{dt} &= vk_1[\rmS1][\rmM]\\
    v\frac{d [\rmM]}{dt} &= 0
\end{align}
where $v > 0$ is the volume of compartment.

\clearpage

\section{Additional Experiment Results}

\subsection{Other types of experiment perturbations}
\label{sec:knockout}
\textbf{How do models perform with other types of experimental perturbations?
} We explored an additional experimental capability by enabling agents to perform \textit{species knockouts}, which completely removes a specified species from the system by eliminating all reactions where it participates and setting its initial concentration to zero. Such knockouts are typically extremely expensive or impossible in general biological systems. We also acknowledge this abstraction is our custom design rather than a standard SBML operation. Due to the computational constraints, we only did this ablation study on two Gemini models. We observed modest performance improvements as shown in Table \ref{tab:knockout}. Future work should focus on designing biologically meaningful perturbations and exploring whether more realistic experimental tools can substantially improve discovery performance.

\begin{table}[h]
\caption{\textbf{Adding knockout perturbations improve model performance modestly.} We compared Gemini models' performance with and without access to our designed knockout operations. C indicates changing initial concentrations, and K indicates knockout experiments. Models generally demonstrated modest performance improvements when given access to knockout capabilities.}
\begin{tabular}{lllll}
\hline
\multicolumn{1}{c}{\multirow{2}{*}{\textbf{Model}}} & \multicolumn{1}{c}{\multirow{2}{*}{\textbf{Experiment}}} & \multirow{2}{*}{\textbf{STE $\downarrow$}} & \multicolumn{2}{c}{\textbf{RMS $\uparrow$}} \\
\multicolumn{1}{c}{}                                & \multicolumn{1}{c}{}                                     &                                            & \textbf{with mod}     & \textbf{w/o mod}    \\ \hline
\textbf{Gemi-M}     & C          & \textbf{0.4181}                            & 0.1217     & 0.2005              \\ & C + K            & 0.4276     & \textbf{0.1417}       & \textbf{0.2454}     \\ \hline
\textbf{Gemi-P}    & C    & 0.3212  & 0.1817    & 0.3383    \\
 & C + K    & \textbf{0.3003}   & \textbf{0.1923} & \textbf{0.3527}  \\ \hline
\end{tabular}
\label{tab:knockout}
\end{table}

\subsection{Detailed Results by BioMD}
\label{sec:detail-results}
\begin{table}[H]
\centering
\caption{Detailed results for benchmark BIOMD0000000039 across all models.}
\label{tab:BIOMD0000000039}
\begin{tabular}{l|c|ccc|ccc}
\toprule
\multirow{2}{*}{\textbf{Model}} & \multirow{2}{*}{\textbf{STE $\downarrow$}} & \multicolumn{3}{c|}{\textbf{RMS (with modifiers)} $\uparrow$} & \multicolumn{3}{c}{\textbf{RMS (without modifiers)} $\uparrow$} \\ 
 & & \textbf{Precision} & \textbf{Recall} & \textbf{F1} & \textbf{Precision} & \textbf{Recall} & \textbf{F1} \\
\midrule
Gemini-2.5-Flash & 0.2213 & 0.1667 & 0.1429 & 0.1538 & 0.3333 & 0.3333 & 0.3333 \\
Gemini-2.5-Pro & 0.1022 & 0.2500 & 0.1429 & 0.1818 & 0.5000 & 0.3333 & 0.4000 \\
\midrule
Gpt-4.1-mini & 0.5452 & 0.0000 & 0.0000 & 0.0000 & 0.2500 & 0.1667 & 0.2000 \\
Gpt-4.1 & 0.5729 & 0.2000 & 0.1429 & 0.1667 & 0.2000 & 0.1667 & 0.1818 \\
\midrule
Claude-3.5-Haiku & 0.1629 & 0.0000 & 0.0000 & 0.0000 & 0.0000 & 0.0000 & 0.0000 \\
Claude-3.7-Sonnet & 0.1298 & 0.0000 & 0.0000 & 0.0000 & 0.2500 & 0.1667 & 0.2000 \\
\bottomrule
\end{tabular}
\end{table}

\subsection{Detailed Results by BioMD}
\label{sec:detail-results}
\begin{table}[H]
\centering
\caption{Detailed results for benchmark BIOMD0000000039 across all models.}
\label{tab:BIOMD0000000039}
\begin{tabular}{l|c|ccc|ccc}
\toprule
\multirow{2}{*}{\textbf{Model}} & \multirow{2}{*}{\textbf{STE $\downarrow$}} & \multicolumn{3}{c|}{\textbf{RMS (with modifiers)} $\uparrow$} & \multicolumn{3}{c}{\textbf{RMS (without modifiers)} $\uparrow$} \\ 
 & & \textbf{Precision} & \textbf{Recall} & \textbf{F1} & \textbf{Precision} & \textbf{Recall} & \textbf{F1} \\
\midrule
Gemini-2.5-Flash & 0.2213 & 0.1667 & 0.1429 & 0.1538 & 0.3333 & 0.3333 & 0.3333 \\
Gemini-2.5-Pro & 0.1022 & 0.2500 & 0.1429 & 0.1818 & 0.5000 & 0.3333 & 0.4000 \\
\midrule
Gpt-4.1-mini & 0.5452 & 0.0000 & 0.0000 & 0.0000 & 0.2500 & 0.1667 & 0.2000 \\
Gpt-4.1 & 0.5729 & 0.2000 & 0.1429 & 0.1667 & 0.2000 & 0.1667 & 0.1818 \\
\midrule
Claude-3.5-Haiku & 0.1629 & 0.0000 & 0.0000 & 0.0000 & 0.0000 & 0.0000 & 0.0000 \\
Claude-3.7-Sonnet & 0.1298 & 0.0000 & 0.0000 & 0.0000 & 0.2500 & 0.1667 & 0.2000 \\
\bottomrule
\end{tabular}
\end{table}

\begin{table}[H]
\centering
\caption{Detailed results for benchmark BIOMD0000000041 across all models.}
\label{tab:BIOMD0000000041}
\begin{tabular}{l|c|ccc|ccc}
\toprule
\multirow{2}{*}{\textbf{Model}} & \multirow{2}{*}{\textbf{STE $\downarrow$}} & \multicolumn{3}{c|}{\textbf{RMS (with modifiers)} $\uparrow$} & \multicolumn{3}{c}{\textbf{RMS (without modifiers)} $\uparrow$} \\ 
 & & \textbf{Precision} & \textbf{Recall} & \textbf{F1} & \textbf{Precision} & \textbf{Recall} & \textbf{F1} \\
\midrule
Gemini-2.5-Flash & 0.6552 & 0.0000 & 0.0000 & 0.0000 & 0.0000 & 0.0000 & 0.0000 \\
Gemini-2.5-Pro & 0.3252 & 0.0000 & 0.0000 & 0.0000 & 0.0000 & 0.0000 & 0.0000 \\
\midrule
Gpt-4.1-mini & 0.8718 & 0.1429 & 0.1111 & 0.1250 & 0.1429 & 0.1111 & 0.1250 \\
Gpt-4.1 & 0.7187 & 0.0000 & 0.0000 & 0.0000 & 0.0000 & 0.0000 & 0.0000 \\
\midrule
Claude-3.5-Haiku & 0.8415 & 0.0000 & 0.0000 & 0.0000 & 0.0000 & 0.0000 & 0.0000 \\
Claude-3.7-Sonnet & 0.3036 & 0.0455 & 0.1111 & 0.0645 & 0.0455 & 0.1111 & 0.0645 \\
\bottomrule
\end{tabular}
\end{table}

\begin{table}[H]
\centering
\caption{Detailed results for benchmark BIOMD0000000043 across all models.}
\label{tab:BIOMD0000000043}
\begin{tabular}{l|c|ccc|ccc}
\toprule
\multirow{2}{*}{\textbf{Model}} & \multirow{2}{*}{\textbf{STE $\downarrow$}} & \multicolumn{3}{c|}{\textbf{RMS (with modifiers)} $\uparrow$} & \multicolumn{3}{c}{\textbf{RMS (without modifiers)} $\uparrow$} \\ 
 & & \textbf{Precision} & \textbf{Recall} & \textbf{F1} & \textbf{Precision} & \textbf{Recall} & \textbf{F1} \\
\midrule
Gemini-2.5-Flash & 0.9965 & 0.4000 & 0.2857 & 0.3333 & 0.6000 & 0.5000 & 0.5455 \\
Gemini-2.5-Pro & 0.5637 & 0.1667 & 0.1429 & 0.1538 & 0.1667 & 0.1667 & 0.1667 \\
\midrule
Gpt-4.1-mini & 0.7097 & 0.0000 & 0.0000 & 0.0000 & 0.0000 & 0.0000 & 0.0000 \\
Gpt-4.1 & 0.7679 & 0.0000 & 0.0000 & 0.0000 & 0.1667 & 0.1667 & 0.1667 \\
\midrule
Claude-3.5-Haiku & 0.8383 & 0.0000 & 0.0000 & 0.0000 & 0.3333 & 0.1667 & 0.2222 \\
Claude-3.7-Sonnet & 0.5688 & 0.1000 & 0.1429 & 0.1176 & 0.2000 & 0.3333 & 0.2500 \\
\bottomrule
\end{tabular}
\end{table}

\begin{table}[H]
\centering
\caption{Detailed results for benchmark BIOMD0000000044 across all models.}
\label{tab:BIOMD0000000044}
\begin{tabular}{l|c|ccc|ccc}
\toprule
\multirow{2}{*}{\textbf{Model}} & \multirow{2}{*}{\textbf{STE $\downarrow$}} & \multicolumn{3}{c|}{\textbf{RMS (with modifiers)} $\uparrow$} & \multicolumn{3}{c}{\textbf{RMS (without modifiers)} $\uparrow$} \\ 
 & & \textbf{Precision} & \textbf{Recall} & \textbf{F1} & \textbf{Precision} & \textbf{Recall} & \textbf{F1} \\
\midrule
Gemini-2.5-Flash & 0.4545 & 0.3333 & 0.2500 & 0.2857 & 0.3333 & 0.3333 & 0.3333 \\
Gemini-2.5-Pro & 0.3169 & 0.0000 & 0.0000 & 0.0000 & 0.1429 & 0.1667 & 0.1538 \\
\midrule
Gpt-4.1-mini & 0.4833 & 0.0000 & 0.0000 & 0.0000 & 0.0000 & 0.0000 & 0.0000 \\
Gpt-4.1 & 0.4163 & 0.3333 & 0.3750 & 0.3529 & 0.3333 & 0.5000 & 0.4000 \\
\midrule
Claude-3.5-Haiku & 0.4833 & 0.0000 & 0.0000 & 0.0000 & 0.0000 & 0.0000 & 0.0000 \\
Claude-3.7-Sonnet & 0.2929 & 0.1667 & 0.1250 & 0.1429 & 0.1667 & 0.1667 & 0.1667 \\
\bottomrule
\end{tabular}
\end{table}

\begin{table}[H]
\centering
\caption{Detailed results for benchmark BIOMD0000000045 across all models.}
\label{tab:BIOMD0000000045}
\begin{tabular}{l|c|ccc|ccc}
\toprule
\multirow{2}{*}{\textbf{Model}} & \multirow{2}{*}{\textbf{STE $\downarrow$}} & \multicolumn{3}{c|}{\textbf{RMS (with modifiers)} $\uparrow$} & \multicolumn{3}{c}{\textbf{RMS (without modifiers)} $\uparrow$} \\ 
 & & \textbf{Precision} & \textbf{Recall} & \textbf{F1} & \textbf{Precision} & \textbf{Recall} & \textbf{F1} \\
\midrule
Gemini-2.5-Flash & 0.9667 & 0.2500 & 0.3333 & 0.2857 & 0.2500 & 0.3333 & 0.2857 \\
Gemini-2.5-Pro & 0.6670 & 0.2000 & 0.1667 & 0.1818 & 0.2000 & 0.1667 & 0.1818 \\
\midrule
Gpt-4.1-mini & 0.8494 & 0.5000 & 0.3333 & 0.4000 & 0.5000 & 0.3333 & 0.4000 \\
Gpt-4.1 & 0.7270 & 0.2222 & 0.3333 & 0.2667 & 0.2222 & 0.3333 & 0.2667 \\
\midrule
Claude-3.5-Haiku & 0.8563 & 0.3333 & 0.1667 & 0.2222 & 0.3333 & 0.1667 & 0.2222 \\
Claude-3.7-Sonnet & 0.9199 & 0.1667 & 0.1667 & 0.1667 & 0.1667 & 0.1667 & 0.1667 \\
\bottomrule
\end{tabular}
\end{table}

\begin{table}[H]
\centering
\caption{Detailed results for benchmark BIOMD0000000066 across all models.}
\label{tab:BIOMD0000000066}
\begin{tabular}{l|c|ccc|ccc}
\toprule
\multirow{2}{*}{\textbf{Model}} & \multirow{2}{*}{\textbf{STE $\downarrow$}} & \multicolumn{3}{c|}{\textbf{RMS (with modifiers)} $\uparrow$} & \multicolumn{3}{c}{\textbf{RMS (without modifiers)} $\uparrow$} \\ 
 & & \textbf{Precision} & \textbf{Recall} & \textbf{F1} & \textbf{Precision} & \textbf{Recall} & \textbf{F1} \\
\midrule
Gemini-2.5-Flash & 0.5611 & 0.0000 & 0.0000 & 0.0000 & 0.0000 & 0.0000 & 0.0000 \\
Gemini-2.5-Pro & 0.3937 & 0.2857 & 0.2857 & 0.2857 & 0.2857 & 0.2857 & 0.2857 \\
\midrule
Gpt-4.1-mini & 0.8539 & 0.0000 & 0.0000 & 0.0000 & 0.0000 & 0.0000 & 0.0000 \\
Gpt-4.1 & 0.6358 & 0.3333 & 0.1429 & 0.2000 & 0.3333 & 0.1429 & 0.2000 \\
\midrule
Claude-3.5-Haiku & 0.8511 & 0.0000 & 0.0000 & 0.0000 & 0.0000 & 0.0000 & 0.0000 \\
Claude-3.7-Sonnet & 0.6764 & 0.1250 & 0.1429 & 0.1333 & 0.1250 & 0.1429 & 0.1333 \\
\bottomrule
\end{tabular}
\end{table}

\begin{table}[H]
\centering
\caption{Detailed results for benchmark BIOMD0000000068 across all models.}
\label{tab:BIOMD0000000068}
\begin{tabular}{l|c|ccc|ccc}
\toprule
\multirow{2}{*}{\textbf{Model}} & \multirow{2}{*}{\textbf{STE $\downarrow$}} & \multicolumn{3}{c|}{\textbf{RMS (with modifiers)} $\uparrow$} & \multicolumn{3}{c}{\textbf{RMS (without modifiers)} $\uparrow$} \\ 
 & & \textbf{Precision} & \textbf{Recall} & \textbf{F1} & \textbf{Precision} & \textbf{Recall} & \textbf{F1} \\
\midrule
Gemini-2.5-Flash & 0.0040 & 0.0000 & 0.0000 & 0.0000 & 0.0000 & 0.0000 & 0.0000 \\
Gemini-2.5-Pro & 0.0001 & 0.0000 & 0.0000 & 0.0000 & 0.0000 & 0.0000 & 0.0000 \\
\midrule
Gpt-4.1-mini & 0.1896 & 0.0000 & 0.0000 & 0.0000 & 0.0000 & 0.0000 & 0.0000 \\
Gpt-4.1 & 0.0005 & 0.0000 & 0.0000 & 0.0000 & 0.0000 & 0.0000 & 0.0000 \\
\midrule
Claude-3.5-Haiku & 0.0257 & 0.0000 & 0.0000 & 0.0000 & 0.0000 & 0.0000 & 0.0000 \\
Claude-3.7-Sonnet & 0.0002 & 0.0000 & 0.0000 & 0.0000 & 0.0000 & 0.0000 & 0.0000 \\
\bottomrule
\end{tabular}
\end{table}

\begin{table}[H]
\centering
\caption{Detailed results for benchmark BIOMD0000000072 across all models.}
\label{tab:BIOMD0000000072}
\begin{tabular}{l|c|ccc|ccc}
\toprule
\multirow{2}{*}{\textbf{Model}} & \multirow{2}{*}{\textbf{STE $\downarrow$}} & \multicolumn{3}{c|}{\textbf{RMS (with modifiers)} $\uparrow$} & \multicolumn{3}{c}{\textbf{RMS (without modifiers)} $\uparrow$} \\ 
 & & \textbf{Precision} & \textbf{Recall} & \textbf{F1} & \textbf{Precision} & \textbf{Recall} & \textbf{F1} \\
\midrule
Gemini-2.5-Flash & 0.3495 & 0.0000 & 0.0000 & 0.0000 & 0.0000 & 0.0000 & 0.0000 \\
Gemini-2.5-Pro & 0.3401 & 0.2000 & 0.1429 & 0.1667 & 0.2000 & 0.1429 & 0.1667 \\
\midrule
Gpt-4.1-mini & 0.7310 & 0.0000 & 0.0000 & 0.0000 & 0.0000 & 0.0000 & 0.0000 \\
Gpt-4.1 & 0.4918 & 0.0000 & 0.0000 & 0.0000 & 0.0000 & 0.0000 & 0.0000 \\
\midrule
Claude-3.5-Haiku & 0.4244 & 0.0000 & 0.0000 & 0.0000 & 0.0000 & 0.0000 & 0.0000 \\
Claude-3.7-Sonnet & 0.6369 & 0.2500 & 0.2857 & 0.2667 & 0.2500 & 0.2857 & 0.2667 \\
\bottomrule
\end{tabular}
\end{table}

\begin{table}[H]
\centering
\caption{Detailed results for benchmark BIOMD0000000076 across all models.}
\label{tab:BIOMD0000000076}
\begin{tabular}{l|c|ccc|ccc}
\toprule
\multirow{2}{*}{\textbf{Model}} & \multirow{2}{*}{\textbf{STE $\downarrow$}} & \multicolumn{3}{c|}{\textbf{RMS (with modifiers)} $\uparrow$} & \multicolumn{3}{c}{\textbf{RMS (without modifiers)} $\uparrow$} \\ 
 & & \textbf{Precision} & \textbf{Recall} & \textbf{F1} & \textbf{Precision} & \textbf{Recall} & \textbf{F1} \\
\midrule
Gemini-2.5-Flash & 0.0035 & 0.0000 & 0.0000 & 0.0000 & 0.0000 & 0.0000 & 0.0000 \\
Gemini-2.5-Pro & 0.0001 & 0.0000 & 0.0000 & 0.0000 & 0.0000 & 0.0000 & 0.0000 \\
\midrule
Gpt-4.1-mini & 0.1815 & 0.0000 & 0.0000 & 0.0000 & 0.0000 & 0.0000 & 0.0000 \\
Gpt-4.1 & 0.0005 & 0.0000 & 0.0000 & 0.0000 & 0.0000 & 0.0000 & 0.0000 \\
\midrule
Claude-3.5-Haiku & 0.0136 & 0.0000 & 0.0000 & 0.0000 & 0.0000 & 0.0000 & 0.0000 \\
Claude-3.7-Sonnet & 0.0003 & 0.5000 & 0.5000 & 0.5000 & 0.5000 & 0.5000 & 0.5000 \\
\bottomrule
\end{tabular}
\end{table}

\begin{table}[H]
\centering
\caption{Detailed results for benchmark BIOMD0000000079 across all models.}
\label{tab:BIOMD0000000079}
\begin{tabular}{l|c|ccc|ccc}
\toprule
\multirow{2}{*}{\textbf{Model}} & \multirow{2}{*}{\textbf{STE $\downarrow$}} & \multicolumn{3}{c|}{\textbf{RMS (with modifiers)} $\uparrow$} & \multicolumn{3}{c}{\textbf{RMS (without modifiers)} $\uparrow$} \\ 
 & & \textbf{Precision} & \textbf{Recall} & \textbf{F1} & \textbf{Precision} & \textbf{Recall} & \textbf{F1} \\
\midrule
Gemini-2.5-Flash & 0.2022 & 0.0000 & 0.0000 & 0.0000 & 0.2500 & 0.1667 & 0.2000 \\
Gemini-2.5-Pro & 0.2869 & 0.4286 & 0.5000 & 0.4615 & 0.5714 & 0.6667 & 0.6154 \\
\midrule
Gpt-4.1-mini & 0.3794 & 0.2500 & 0.1667 & 0.2000 & 0.2500 & 0.1667 & 0.2000 \\
Gpt-4.1 & 0.5643 & 0.2857 & 0.3333 & 0.3077 & 0.4286 & 0.5000 & 0.4615 \\
\midrule
Claude-3.5-Haiku & 0.2413 & 0.0000 & 0.0000 & 0.0000 & 0.0000 & 0.0000 & 0.0000 \\
Claude-3.7-Sonnet & 0.3171 & 0.3333 & 0.5000 & 0.4000 & 0.5556 & 0.8333 & 0.6667 \\
\bottomrule
\end{tabular}
\end{table}

\begin{table}[H]
\centering
\caption{Detailed results for benchmark BIOMD0000000080 across all models.}
\label{tab:BIOMD0000000080}
\begin{tabular}{l|c|ccc|ccc}
\toprule
\multirow{2}{*}{\textbf{Model}} & \multirow{2}{*}{\textbf{STE $\downarrow$}} & \multicolumn{3}{c|}{\textbf{RMS (with modifiers)} $\uparrow$} & \multicolumn{3}{c}{\textbf{RMS (without modifiers)} $\uparrow$} \\ 
 & & \textbf{Precision} & \textbf{Recall} & \textbf{F1} & \textbf{Precision} & \textbf{Recall} & \textbf{F1} \\
\midrule
Gemini-2.5-Flash & 0.2300 & 0.0000 & 0.0000 & 0.0000 & 0.0000 & 0.0000 & 0.0000 \\
Gemini-2.5-Pro & 0.3577 & 0.0000 & 0.0000 & 0.0000 & 0.0000 & 0.0000 & 0.0000 \\
\midrule
Gpt-4.1-mini & 0.6403 & 0.0000 & 0.0000 & 0.0000 & 0.0000 & 0.0000 & 0.0000 \\
Gpt-4.1 & 0.5796 & 0.0000 & 0.0000 & 0.0000 & 0.0000 & 0.0000 & 0.0000 \\
\midrule
Claude-3.5-Haiku & 0.5873 & 0.0000 & 0.0000 & 0.0000 & 0.0000 & 0.0000 & 0.0000 \\
Claude-3.7-Sonnet & 0.1367 & 0.0000 & 0.0000 & 0.0000 & 0.0000 & 0.0000 & 0.0000 \\
\bottomrule
\end{tabular}
\end{table}

\begin{table}[H]
\centering
\caption{Detailed results for benchmark BIOMD0000000082 across all models.}
\label{tab:BIOMD0000000082}
\begin{tabular}{l|c|ccc|ccc}
\toprule
\multirow{2}{*}{\textbf{Model}} & \multirow{2}{*}{\textbf{STE $\downarrow$}} & \multicolumn{3}{c|}{\textbf{RMS (with modifiers)} $\uparrow$} & \multicolumn{3}{c}{\textbf{RMS (without modifiers)} $\uparrow$} \\ 
 & & \textbf{Precision} & \textbf{Recall} & \textbf{F1} & \textbf{Precision} & \textbf{Recall} & \textbf{F1} \\
\midrule
Gemini-2.5-Flash & 0.2705 & 0.0000 & 0.0000 & 0.0000 & 0.0000 & 0.0000 & 0.0000 \\
Gemini-2.5-Pro & 0.2033 & 0.0000 & 0.0000 & 0.0000 & 0.0000 & 0.0000 & 0.0000 \\
\midrule
Gpt-4.1-mini & 0.5927 & 0.0000 & 0.0000 & 0.0000 & 0.0000 & 0.0000 & 0.0000 \\
Gpt-4.1 & 0.4703 & 0.0000 & 0.0000 & 0.0000 & 0.0000 & 0.0000 & 0.0000 \\
\midrule
Claude-3.5-Haiku & 0.6056 & 0.0000 & 0.0000 & 0.0000 & 0.0000 & 0.0000 & 0.0000 \\
Claude-3.7-Sonnet & 0.5290 & 0.0000 & 0.0000 & 0.0000 & 0.0000 & 0.0000 & 0.0000 \\
\bottomrule
\end{tabular}
\end{table}

\begin{table}[H]
\centering
\caption{Detailed results for benchmark BIOMD0000000084 across all models.}
\label{tab:BIOMD0000000084}
\begin{tabular}{l|c|ccc|ccc}
\toprule
\multirow{2}{*}{\textbf{Model}} & \multirow{2}{*}{\textbf{STE $\downarrow$}} & \multicolumn{3}{c|}{\textbf{RMS (with modifiers)} $\uparrow$} & \multicolumn{3}{c}{\textbf{RMS (without modifiers)} $\uparrow$} \\ 
 & & \textbf{Precision} & \textbf{Recall} & \textbf{F1} & \textbf{Precision} & \textbf{Recall} & \textbf{F1} \\
\midrule
Gemini-2.5-Flash & 0.2929 & 0.0000 & 0.0000 & 0.0000 & 0.0000 & 0.0000 & 0.0000 \\
Gemini-2.5-Pro & 0.2624 & 0.2500 & 0.1250 & 0.1667 & 1.0000 & 0.5000 & 0.6667 \\
\midrule
Gpt-4.1-mini & 0.6967 & 0.0000 & 0.0000 & 0.0000 & 0.0000 & 0.0000 & 0.0000 \\
Gpt-4.1 & 0.5450 & 0.0909 & 0.1250 & 0.1053 & 0.0909 & 0.1250 & 0.1053 \\
\midrule
Claude-3.5-Haiku & 0.8307 & 0.3333 & 0.1250 & 0.1818 & 1.0000 & 0.3750 & 0.5455 \\
Claude-3.7-Sonnet & 0.2615 & 0.2500 & 0.1250 & 0.1667 & 1.0000 & 0.5000 & 0.6667 \\
\bottomrule
\end{tabular}
\end{table}

\begin{table}[H]
\centering
\caption{Detailed results for benchmark BIOMD0000000092 across all models.}
\label{tab:BIOMD0000000092}
\begin{tabular}{l|c|ccc|ccc}
\toprule
\multirow{2}{*}{\textbf{Model}} & \multirow{2}{*}{\textbf{STE $\downarrow$}} & \multicolumn{3}{c|}{\textbf{RMS (with modifiers)} $\uparrow$} & \multicolumn{3}{c}{\textbf{RMS (without modifiers)} $\uparrow$} \\ 
 & & \textbf{Precision} & \textbf{Recall} & \textbf{F1} & \textbf{Precision} & \textbf{Recall} & \textbf{F1} \\
\midrule
Gemini-2.5-Flash & 0.3167 & 0.0000 & 0.0000 & 0.0000 & 0.0000 & 0.0000 & 0.0000 \\
Gemini-2.5-Pro & 0.0021 & 0.5000 & 0.3333 & 0.4000 & 0.5000 & 0.3333 & 0.4000 \\
\midrule
Gpt-4.1-mini & 0.9322 & 0.0000 & 0.0000 & 0.0000 & 0.0000 & 0.0000 & 0.0000 \\
Gpt-4.1 & 0.1430 & 0.0000 & 0.0000 & 0.0000 & 0.0000 & 0.0000 & 0.0000 \\
\midrule
Claude-3.5-Haiku & 0.5402 & 0.5000 & 0.3333 & 0.4000 & 0.5000 & 0.3333 & 0.4000 \\
Claude-3.7-Sonnet & 0.0846 & 0.0000 & 0.0000 & 0.0000 & 0.0000 & 0.0000 & 0.0000 \\
\bottomrule
\end{tabular}
\end{table}

\begin{table}[H]
\centering
\caption{Detailed results for benchmark BIOMD0000000098 across all models.}
\label{tab:BIOMD0000000098}
\begin{tabular}{l|c|ccc|ccc}
\toprule
\multirow{2}{*}{\textbf{Model}} & \multirow{2}{*}{\textbf{STE $\downarrow$}} & \multicolumn{3}{c|}{\textbf{RMS (with modifiers)} $\uparrow$} & \multicolumn{3}{c}{\textbf{RMS (without modifiers)} $\uparrow$} \\ 
 & & \textbf{Precision} & \textbf{Recall} & \textbf{F1} & \textbf{Precision} & \textbf{Recall} & \textbf{F1} \\
\midrule
Gemini-2.5-Flash & 0.7426 & 1.0000 & 0.5000 & 0.6667 & 1.0000 & 0.5000 & 0.6667 \\
Gemini-2.5-Pro & 0.4478 & 0.3333 & 0.2500 & 0.2857 & 0.3333 & 0.2500 & 0.2857 \\
\midrule
Gpt-4.1-mini & 0.1116 & 0.6667 & 0.5000 & 0.5714 & 0.6667 & 0.5000 & 0.5714 \\
Gpt-4.1 & 0.4069 & 1.0000 & 0.5000 & 0.6667 & 1.0000 & 0.5000 & 0.6667 \\
\midrule
Claude-3.5-Haiku & 0.3561 & 1.0000 & 0.2500 & 0.4000 & 1.0000 & 0.2500 & 0.4000 \\
Claude-3.7-Sonnet & 0.2657 & 0.1667 & 0.2500 & 0.2000 & 0.1667 & 0.2500 & 0.2000 \\
\bottomrule
\end{tabular}
\end{table}

\begin{table}[H]
\centering
\caption{Detailed results for benchmark BIOMD0000000156 across all models.}
\label{tab:BIOMD0000000156}
\begin{tabular}{l|c|ccc|ccc}
\toprule
\multirow{2}{*}{\textbf{Model}} & \multirow{2}{*}{\textbf{STE $\downarrow$}} & \multicolumn{3}{c|}{\textbf{RMS (with modifiers)} $\uparrow$} & \multicolumn{3}{c}{\textbf{RMS (without modifiers)} $\uparrow$} \\ 
 & & \textbf{Precision} & \textbf{Recall} & \textbf{F1} & \textbf{Precision} & \textbf{Recall} & \textbf{F1} \\
\midrule
Gemini-2.5-Flash & 0.2671 & 0.0000 & 0.0000 & 0.0000 & 0.0000 & 0.0000 & 0.0000 \\
Gemini-2.5-Pro & 0.6261 & 0.3333 & 0.2000 & 0.2500 & 0.3333 & 0.2000 & 0.2500 \\
\midrule
Gpt-4.1-mini & 0.2769 & 0.3333 & 0.2000 & 0.2500 & 0.3333 & 0.2000 & 0.2500 \\
Gpt-4.1 & 0.3085 & 0.1667 & 0.2000 & 0.1818 & 0.3333 & 0.4000 & 0.3636 \\
\midrule
Claude-3.5-Haiku & 0.4243 & 0.0000 & 0.0000 & 0.0000 & 0.0000 & 0.0000 & 0.0000 \\
Claude-3.7-Sonnet & 0.3023 & 0.3333 & 0.4000 & 0.3636 & 0.6667 & 0.8000 & 0.7273 \\
\bottomrule
\end{tabular}
\end{table}

\begin{table}[H]
\centering
\caption{Detailed results for benchmark BIOMD0000000157 across all models.}
\label{tab:BIOMD0000000157}
\begin{tabular}{l|c|ccc|ccc}
\toprule
\multirow{2}{*}{\textbf{Model}} & \multirow{2}{*}{\textbf{STE $\downarrow$}} & \multicolumn{3}{c|}{\textbf{RMS (with modifiers)} $\uparrow$} & \multicolumn{3}{c}{\textbf{RMS (without modifiers)} $\uparrow$} \\ 
 & & \textbf{Precision} & \textbf{Recall} & \textbf{F1} & \textbf{Precision} & \textbf{Recall} & \textbf{F1} \\
\midrule
Gemini-2.5-Flash & 0.5773 & 0.0000 & 0.0000 & 0.0000 & 0.0000 & 0.0000 & 0.0000 \\
Gemini-2.5-Pro & 0.5918 & 0.3333 & 0.3333 & 0.3333 & 0.3333 & 0.4000 & 0.3636 \\
\midrule
Gpt-4.1-mini & 0.5829 & 0.3333 & 0.3333 & 0.3333 & 0.3333 & 0.4000 & 0.3636 \\
Gpt-4.1 & 0.3561 & 0.2500 & 0.1667 & 0.2000 & 0.2500 & 0.2000 & 0.2222 \\
\midrule
Claude-3.5-Haiku & 0.4965 & 0.0000 & 0.0000 & 0.0000 & 0.0000 & 0.0000 & 0.0000 \\
Claude-3.7-Sonnet & 0.4263 & 0.5000 & 0.5000 & 0.5000 & 0.6667 & 0.8000 & 0.7273 \\
\bottomrule
\end{tabular}
\end{table}

\begin{table}[H]
\centering
\caption{Detailed results for benchmark BIOMD0000000159 across all models.}
\label{tab:BIOMD0000000159}
\begin{tabular}{l|c|ccc|ccc}
\toprule
\multirow{2}{*}{\textbf{Model}} & \multirow{2}{*}{\textbf{STE $\downarrow$}} & \multicolumn{3}{c|}{\textbf{RMS (with modifiers)} $\uparrow$} & \multicolumn{3}{c}{\textbf{RMS (without modifiers)} $\uparrow$} \\ 
 & & \textbf{Precision} & \textbf{Recall} & \textbf{F1} & \textbf{Precision} & \textbf{Recall} & \textbf{F1} \\
\midrule
Gemini-2.5-Flash & 1.0000 & 0.2500 & 0.1667 & 0.2000 & 0.2500 & 0.2000 & 0.2222 \\
Gemini-2.5-Pro & 0.2588 & 0.7500 & 0.5000 & 0.6000 & 0.7500 & 0.6000 & 0.6667 \\
\midrule
Gpt-4.1-mini & 0.9990 & 0.5714 & 0.6667 & 0.6154 & 0.5714 & 0.8000 & 0.6667 \\
Gpt-4.1 & 0.4786 & 0.1667 & 0.1667 & 0.1667 & 0.3333 & 0.4000 & 0.3636 \\
\midrule
Claude-3.5-Haiku & 0.9990 & 0.3333 & 0.1667 & 0.2222 & 0.3333 & 0.2000 & 0.2500 \\
Claude-3.7-Sonnet & 0.2440 & 0.6667 & 0.6667 & 0.6667 & 0.6667 & 0.8000 & 0.7273 \\
\bottomrule
\end{tabular}
\end{table}

\begin{table}[H]
\centering
\caption{Detailed results for benchmark BIOMD0000000184 across all models.}
\label{tab:BIOMD0000000184}
\begin{tabular}{l|c|ccc|ccc}
\toprule
\multirow{2}{*}{\textbf{Model}} & \multirow{2}{*}{\textbf{STE $\downarrow$}} & \multicolumn{3}{c|}{\textbf{RMS (with modifiers)} $\uparrow$} & \multicolumn{3}{c}{\textbf{RMS (without modifiers)} $\uparrow$} \\ 
 & & \textbf{Precision} & \textbf{Recall} & \textbf{F1} & \textbf{Precision} & \textbf{Recall} & \textbf{F1} \\
\midrule
Gemini-2.5-Flash & 0.1836 & 0.5000 & 0.1429 & 0.2222 & 0.5000 & 0.1667 & 0.2500 \\
Gemini-2.5-Pro & 0.1107 & 0.6667 & 0.2857 & 0.4000 & 0.6667 & 0.3333 & 0.4444 \\
\midrule
Gpt-4.1-mini & 0.3354 & 0.0000 & 0.0000 & 0.0000 & 0.0000 & 0.0000 & 0.0000 \\
Gpt-4.1 & 0.1673 & 0.6000 & 0.4286 & 0.5000 & 0.6000 & 0.5000 & 0.5455 \\
\midrule
Claude-3.5-Haiku & 0.3201 & 0.0000 & 0.0000 & 0.0000 & 0.0000 & 0.0000 & 0.0000 \\
Claude-3.7-Sonnet & 0.4057 & 0.6000 & 0.4286 & 0.5000 & 0.6000 & 0.5000 & 0.5455 \\
\bottomrule
\end{tabular}
\end{table}

\begin{table}[H]
\centering
\caption{Detailed results for benchmark BIOMD0000000191 across all models.}
\label{tab:BIOMD0000000191}
\begin{tabular}{l|c|ccc|ccc}
\toprule
\multirow{2}{*}{\textbf{Model}} & \multirow{2}{*}{\textbf{STE $\downarrow$}} & \multicolumn{3}{c|}{\textbf{RMS (with modifiers)} $\uparrow$} & \multicolumn{3}{c}{\textbf{RMS (without modifiers)} $\uparrow$} \\ 
 & & \textbf{Precision} & \textbf{Recall} & \textbf{F1} & \textbf{Precision} & \textbf{Recall} & \textbf{F1} \\
\midrule
Gemini-2.5-Flash & 0.0275 & 0.4000 & 0.4000 & 0.4000 & 0.4000 & 0.5000 & 0.4444 \\
Gemini-2.5-Pro & 0.0163 & 0.5000 & 0.4000 & 0.4444 & 0.5000 & 0.5000 & 0.5000 \\
\midrule
Gpt-4.1-mini & 0.4203 & 0.2500 & 0.2000 & 0.2222 & 0.5000 & 0.5000 & 0.5000 \\
Gpt-4.1 & 0.0055 & 0.5000 & 0.4000 & 0.4444 & 0.5000 & 0.5000 & 0.5000 \\
\midrule
Claude-3.5-Haiku & 0.0438 & 0.0000 & 0.0000 & 0.0000 & 0.0000 & 0.0000 & 0.0000 \\
Claude-3.7-Sonnet & 0.4257 & 0.5000 & 0.4000 & 0.4444 & 0.5000 & 0.5000 & 0.5000 \\
\bottomrule
\end{tabular}
\end{table}

\begin{table}[H]
\centering
\caption{Detailed results for benchmark BIOMD0000000192 across all models.}
\label{tab:BIOMD0000000192}
\begin{tabular}{l|c|ccc|ccc}
\toprule
\multirow{2}{*}{\textbf{Model}} & \multirow{2}{*}{\textbf{STE $\downarrow$}} & \multicolumn{3}{c|}{\textbf{RMS (with modifiers)} $\uparrow$} & \multicolumn{3}{c}{\textbf{RMS (without modifiers)} $\uparrow$} \\ 
 & & \textbf{Precision} & \textbf{Recall} & \textbf{F1} & \textbf{Precision} & \textbf{Recall} & \textbf{F1} \\
\midrule
Gemini-2.5-Flash & 0.4250 & 0.0000 & 0.0000 & 0.0000 & 0.0000 & 0.0000 & 0.0000 \\
Gemini-2.5-Pro & 0.1003 & 0.0000 & 0.0000 & 0.0000 & 0.0000 & 0.0000 & 0.0000 \\
\midrule
Gpt-4.1-mini & 0.6365 & 0.0000 & 0.0000 & 0.0000 & 0.0000 & 0.0000 & 0.0000 \\
Gpt-4.1 & 0.3599 & 0.0000 & 0.0000 & 0.0000 & 0.0000 & 0.0000 & 0.0000 \\
\midrule
Claude-3.5-Haiku & 1.0000 & 0.0000 & 0.0000 & 0.0000 & 0.0000 & 0.0000 & 0.0000 \\
Claude-3.7-Sonnet & 0.2660 & 0.0000 & 0.0000 & 0.0000 & 0.0000 & 0.0000 & 0.0000 \\
\bottomrule
\end{tabular}
\end{table}

\begin{table}[H]
\centering
\caption{Detailed results for benchmark BIOMD0000000224 across all models.}
\label{tab:BIOMD0000000224}
\begin{tabular}{l|c|ccc|ccc}
\toprule
\multirow{2}{*}{\textbf{Model}} & \multirow{2}{*}{\textbf{STE $\downarrow$}} & \multicolumn{3}{c|}{\textbf{RMS (with modifiers)} $\uparrow$} & \multicolumn{3}{c}{\textbf{RMS (without modifiers)} $\uparrow$} \\ 
 & & \textbf{Precision} & \textbf{Recall} & \textbf{F1} & \textbf{Precision} & \textbf{Recall} & \textbf{F1} \\
\midrule
Gemini-2.5-Flash & 0.6407 & 0.2000 & 0.1667 & 0.1818 & 0.6000 & 0.5000 & 0.5455 \\
Gemini-2.5-Pro & 0.3212 & 0.2000 & 0.1667 & 0.1818 & 0.6000 & 0.5000 & 0.5455 \\
\midrule
Gpt-4.1-mini & 0.4353 & 0.0000 & 0.0000 & 0.0000 & 0.6667 & 0.3333 & 0.4444 \\
Gpt-4.1 & 0.3359 & 0.0000 & 0.0000 & 0.0000 & 0.3333 & 0.1667 & 0.2222 \\
\midrule
Claude-3.5-Haiku & 0.7506 & 0.0000 & 0.0000 & 0.0000 & 0.0000 & 0.0000 & 0.0000 \\
Claude-3.7-Sonnet & 0.4060 & 0.1667 & 0.1667 & 0.1667 & 0.5000 & 0.5000 & 0.5000 \\
\bottomrule
\end{tabular}
\end{table}

\begin{table}[H]
\centering
\caption{Detailed results for benchmark BIOMD0000000231 across all models.}
\label{tab:BIOMD0000000231}
\begin{tabular}{l|c|ccc|ccc}
\toprule
\multirow{2}{*}{\textbf{Model}} & \multirow{2}{*}{\textbf{STE $\downarrow$}} & \multicolumn{3}{c|}{\textbf{RMS (with modifiers)} $\uparrow$} & \multicolumn{3}{c}{\textbf{RMS (without modifiers)} $\uparrow$} \\ 
 & & \textbf{Precision} & \textbf{Recall} & \textbf{F1} & \textbf{Precision} & \textbf{Recall} & \textbf{F1} \\
\midrule
Gemini-2.5-Flash & 0.0231 & 0.0000 & 0.0000 & 0.0000 & 0.0000 & 0.0000 & 0.0000 \\
Gemini-2.5-Pro & 0.1208 & 0.0000 & 0.0000 & 0.0000 & 0.0000 & 0.0000 & 0.0000 \\
\midrule
Gpt-4.1-mini & 0.5512 & 0.0000 & 0.0000 & 0.0000 & 0.0000 & 0.0000 & 0.0000 \\
Gpt-4.1 & 0.3539 & 0.0000 & 0.0000 & 0.0000 & 0.0000 & 0.0000 & 0.0000 \\
\midrule
Claude-3.5-Haiku & 0.5060 & 0.0000 & 0.0000 & 0.0000 & 0.0000 & 0.0000 & 0.0000 \\
Claude-3.7-Sonnet & 0.0371 & 0.0000 & 0.0000 & 0.0000 & 0.0000 & 0.0000 & 0.0000 \\
\bottomrule
\end{tabular}
\end{table}

\begin{table}[H]
\centering
\caption{Detailed results for benchmark BIOMD0000000233 across all models.}
\label{tab:BIOMD0000000233}
\begin{tabular}{l|c|ccc|ccc}
\toprule
\multirow{2}{*}{\textbf{Model}} & \multirow{2}{*}{\textbf{STE $\downarrow$}} & \multicolumn{3}{c|}{\textbf{RMS (with modifiers)} $\uparrow$} & \multicolumn{3}{c}{\textbf{RMS (without modifiers)} $\uparrow$} \\ 
 & & \textbf{Precision} & \textbf{Recall} & \textbf{F1} & \textbf{Precision} & \textbf{Recall} & \textbf{F1} \\
\midrule
Gemini-2.5-Flash & 0.2611 & 0.0000 & 0.0000 & 0.0000 & 0.0000 & 0.0000 & 0.0000 \\
Gemini-2.5-Pro & 0.2732 & 0.3333 & 0.2500 & 0.2857 & 0.3333 & 0.2500 & 0.2857 \\
\midrule
Gpt-4.1-mini & 0.2065 & 0.0000 & 0.0000 & 0.0000 & 0.0000 & 0.0000 & 0.0000 \\
Gpt-4.1 & 0.2401 & 0.0000 & 0.0000 & 0.0000 & 0.0000 & 0.0000 & 0.0000 \\
\midrule
Claude-3.5-Haiku & 0.4110 & 0.0000 & 0.0000 & 0.0000 & 0.0000 & 0.0000 & 0.0000 \\
Claude-3.7-Sonnet & 0.3013 & 0.0000 & 0.0000 & 0.0000 & 0.0000 & 0.0000 & 0.0000 \\
\bottomrule
\end{tabular}
\end{table}

\begin{table}[H]
\centering
\caption{Detailed results for benchmark BIOMD0000000258 across all models.}
\label{tab:BIOMD0000000258}
\begin{tabular}{l|c|ccc|ccc}
\toprule
\multirow{2}{*}{\textbf{Model}} & \multirow{2}{*}{\textbf{STE $\downarrow$}} & \multicolumn{3}{c|}{\textbf{RMS (with modifiers)} $\uparrow$} & \multicolumn{3}{c}{\textbf{RMS (without modifiers)} $\uparrow$} \\ 
 & & \textbf{Precision} & \textbf{Recall} & \textbf{F1} & \textbf{Precision} & \textbf{Recall} & \textbf{F1} \\
\midrule
Gemini-2.5-Flash & 0.1344 & 0.3333 & 0.2500 & 0.2857 & 0.6667 & 0.5000 & 0.5714 \\
Gemini-2.5-Pro & 0.2931 & 0.0000 & 0.0000 & 0.0000 & 0.0000 & 0.0000 & 0.0000 \\
\midrule
Gpt-4.1-mini & 0.4833 & 0.3333 & 0.2500 & 0.2857 & 0.3333 & 0.2500 & 0.2857 \\
Gpt-4.1 & 0.3291 & 0.5000 & 0.2500 & 0.3333 & 1.0000 & 0.5000 & 0.6667 \\
\midrule
Claude-3.5-Haiku & 0.5966 & 0.0000 & 0.0000 & 0.0000 & 0.0000 & 0.0000 & 0.0000 \\
Claude-3.7-Sonnet & 0.0885 & 0.3333 & 0.2500 & 0.2857 & 1.0000 & 0.7500 & 0.8571 \\
\bottomrule
\end{tabular}
\end{table}

\begin{table}[H]
\centering
\caption{Detailed results for benchmark BIOMD0000000282 across all models.}
\label{tab:BIOMD0000000282}
\begin{tabular}{l|c|ccc|ccc}
\toprule
\multirow{2}{*}{\textbf{Model}} & \multirow{2}{*}{\textbf{STE $\downarrow$}} & \multicolumn{3}{c|}{\textbf{RMS (with modifiers)} $\uparrow$} & \multicolumn{3}{c}{\textbf{RMS (without modifiers)} $\uparrow$} \\ 
 & & \textbf{Precision} & \textbf{Recall} & \textbf{F1} & \textbf{Precision} & \textbf{Recall} & \textbf{F1} \\
\midrule
Gemini-2.5-Flash & 0.0918 & 0.0000 & 0.0000 & 0.0000 & 0.0000 & 0.0000 & 0.0000 \\
Gemini-2.5-Pro & 0.1365 & 0.0000 & 0.0000 & 0.0000 & 0.0000 & 0.0000 & 0.0000 \\
\midrule
Gpt-4.1-mini & 0.3644 & 0.0000 & 0.0000 & 0.0000 & 0.0000 & 0.0000 & 0.0000 \\
Gpt-4.1 & 0.3056 & 0.5000 & 0.3333 & 0.4000 & 0.5000 & 0.3333 & 0.4000 \\
\midrule
Claude-3.5-Haiku & 0.3980 & 0.0000 & 0.0000 & 0.0000 & 0.0000 & 0.0000 & 0.0000 \\
Claude-3.7-Sonnet & 0.2512 & 0.0000 & 0.0000 & 0.0000 & 0.0000 & 0.0000 & 0.0000 \\
\bottomrule
\end{tabular}
\end{table}

\begin{table}[H]
\centering
\caption{Detailed results for benchmark BIOMD0000000283 across all models.}
\label{tab:BIOMD0000000283}
\begin{tabular}{l|c|ccc|ccc}
\toprule
\multirow{2}{*}{\textbf{Model}} & \multirow{2}{*}{\textbf{STE $\downarrow$}} & \multicolumn{3}{c|}{\textbf{RMS (with modifiers)} $\uparrow$} & \multicolumn{3}{c}{\textbf{RMS (without modifiers)} $\uparrow$} \\ 
 & & \textbf{Precision} & \textbf{Recall} & \textbf{F1} & \textbf{Precision} & \textbf{Recall} & \textbf{F1} \\
\midrule
Gemini-2.5-Flash & 0.8293 & 0.5000 & 0.5000 & 0.5000 & 0.5000 & 0.5000 & 0.5000 \\
Gemini-2.5-Pro & 0.2757 & 0.0000 & 0.0000 & 0.0000 & 0.0000 & 0.0000 & 0.0000 \\
\midrule
Gpt-4.1-mini & 0.5541 & 0.0000 & 0.0000 & 0.0000 & 0.0000 & 0.0000 & 0.0000 \\
Gpt-4.1 & 0.5206 & 0.0000 & 0.0000 & 0.0000 & 0.0000 & 0.0000 & 0.0000 \\
\midrule
Claude-3.5-Haiku & 0.5216 & 0.0000 & 0.0000 & 0.0000 & 0.0000 & 0.0000 & 0.0000 \\
Claude-3.7-Sonnet & 0.2165 & 0.0000 & 0.0000 & 0.0000 & 0.0000 & 0.0000 & 0.0000 \\
\bottomrule
\end{tabular}
\end{table}

\begin{table}[H]
\centering
\caption{Detailed results for benchmark BIOMD0000000319 across all models.}
\label{tab:BIOMD0000000319}
\begin{tabular}{l|c|ccc|ccc}
\toprule
\multirow{2}{*}{\textbf{Model}} & \multirow{2}{*}{\textbf{STE $\downarrow$}} & \multicolumn{3}{c|}{\textbf{RMS (with modifiers)} $\uparrow$} & \multicolumn{3}{c}{\textbf{RMS (without modifiers)} $\uparrow$} \\ 
 & & \textbf{Precision} & \textbf{Recall} & \textbf{F1} & \textbf{Precision} & \textbf{Recall} & \textbf{F1} \\
\midrule
Gemini-2.5-Flash & 0.2432 & 0.0000 & 0.0000 & 0.0000 & 0.0000 & 0.0000 & 0.0000 \\
Gemini-2.5-Pro & 0.0467 & 0.5000 & 0.5000 & 0.5000 & 0.5000 & 0.5000 & 0.5000 \\
\midrule
Gpt-4.1-mini & 0.5681 & 0.3750 & 0.7500 & 0.5000 & 0.3750 & 0.7500 & 0.5000 \\
Gpt-4.1 & 0.3117 & 0.6000 & 0.7500 & 0.6667 & 0.6000 & 0.7500 & 0.6667 \\
\midrule
Claude-3.5-Haiku & 0.3284 & 0.5000 & 0.2500 & 0.3333 & 0.5000 & 0.2500 & 0.3333 \\
Claude-3.7-Sonnet & 0.0851 & 0.2500 & 0.2500 & 0.2500 & 0.2500 & 0.2500 & 0.2500 \\
\bottomrule
\end{tabular}
\end{table}

\begin{table}[H]
\centering
\caption{Detailed results for benchmark BIOMD0000000322 across all models.}
\label{tab:BIOMD0000000322}
\begin{tabular}{l|c|ccc|ccc}
\toprule
\multirow{2}{*}{\textbf{Model}} & \multirow{2}{*}{\textbf{STE $\downarrow$}} & \multicolumn{3}{c|}{\textbf{RMS (with modifiers)} $\uparrow$} & \multicolumn{3}{c}{\textbf{RMS (without modifiers)} $\uparrow$} \\ 
 & & \textbf{Precision} & \textbf{Recall} & \textbf{F1} & \textbf{Precision} & \textbf{Recall} & \textbf{F1} \\
\midrule
Gemini-2.5-Flash & 0.5064 & 0.0000 & 0.0000 & 0.0000 & 0.0000 & 0.0000 & 0.0000 \\
Gemini-2.5-Pro & 0.5171 & 0.3333 & 0.2222 & 0.2667 & 0.3333 & 0.2222 & 0.2667 \\
\midrule
Gpt-4.1-mini & 0.4708 & 0.1429 & 0.1111 & 0.1250 & 0.1429 & 0.1111 & 0.1250 \\
Gpt-4.1 & 0.4354 & 0.0000 & 0.0000 & 0.0000 & 0.0000 & 0.0000 & 0.0000 \\
\midrule
Claude-3.5-Haiku & 0.8012 & 0.3333 & 0.1111 & 0.1667 & 0.3333 & 0.1111 & 0.1667 \\
Claude-3.7-Sonnet & 0.5632 & 0.0000 & 0.0000 & 0.0000 & 0.0000 & 0.0000 & 0.0000 \\
\bottomrule
\end{tabular}
\end{table}

\begin{table}[H]
\centering
\caption{Detailed results for benchmark BIOMD0000000323 across all models.}
\label{tab:BIOMD0000000323}
\begin{tabular}{l|c|ccc|ccc}
\toprule
\multirow{2}{*}{\textbf{Model}} & \multirow{2}{*}{\textbf{STE $\downarrow$}} & \multicolumn{3}{c|}{\textbf{RMS (with modifiers)} $\uparrow$} & \multicolumn{3}{c}{\textbf{RMS (without modifiers)} $\uparrow$} \\ 
 & & \textbf{Precision} & \textbf{Recall} & \textbf{F1} & \textbf{Precision} & \textbf{Recall} & \textbf{F1} \\
\midrule
Gemini-2.5-Flash & 0.2559 & 0.0000 & 0.0000 & 0.0000 & 0.0000 & 0.0000 & 0.0000 \\
Gemini-2.5-Pro & 0.2966 & 0.5000 & 0.5000 & 0.5000 & 0.5000 & 0.5000 & 0.5000 \\
\midrule
Gpt-4.1-mini & 0.5472 & 0.0000 & 0.0000 & 0.0000 & 0.0000 & 0.0000 & 0.0000 \\
Gpt-4.1 & 0.5136 & 0.5000 & 0.5000 & 0.5000 & 0.5000 & 0.5000 & 0.5000 \\
\midrule
Claude-3.5-Haiku & 0.7152 & 0.0000 & 0.0000 & 0.0000 & 0.0000 & 0.0000 & 0.0000 \\
Claude-3.7-Sonnet & 0.1352 & 0.5000 & 0.5000 & 0.5000 & 0.5000 & 0.5000 & 0.5000 \\
\bottomrule
\end{tabular}
\end{table}

\begin{table}[H]
\centering
\caption{Detailed results for benchmark BIOMD0000000329 across all models.}
\label{tab:BIOMD0000000329}
\begin{tabular}{l|c|ccc|ccc}
\toprule
\multirow{2}{*}{\textbf{Model}} & \multirow{2}{*}{\textbf{STE $\downarrow$}} & \multicolumn{3}{c|}{\textbf{RMS (with modifiers)} $\uparrow$} & \multicolumn{3}{c}{\textbf{RMS (without modifiers)} $\uparrow$} \\ 
 & & \textbf{Precision} & \textbf{Recall} & \textbf{F1} & \textbf{Precision} & \textbf{Recall} & \textbf{F1} \\
\midrule
Gemini-2.5-Flash & 0.3470 & 0.5000 & 0.2500 & 0.3333 & 0.5000 & 0.3333 & 0.4000 \\
Gemini-2.5-Pro & 0.8958 & 0.1667 & 0.1250 & 0.1429 & 0.6667 & 0.6667 & 0.6667 \\
\midrule
Gpt-4.1-mini & 0.7202 & 0.2000 & 0.1250 & 0.1538 & 0.6000 & 0.5000 & 0.5455 \\
Gpt-4.1 & 0.7084 & 0.4286 & 0.3750 & 0.4000 & 0.5714 & 0.6667 & 0.6154 \\
\midrule
Claude-3.5-Haiku & 0.9347 & 0.0000 & 0.0000 & 0.0000 & 0.0000 & 0.0000 & 0.0000 \\
Claude-3.7-Sonnet & 0.5742 & 0.5714 & 0.5000 & 0.5333 & 1.0000 & 1.0000 & 1.0000 \\
\bottomrule
\end{tabular}
\end{table}

\begin{table}[H]
\centering
\caption{Detailed results for benchmark BIOMD0000000357 across all models.}
\label{tab:BIOMD0000000357}
\begin{tabular}{l|c|ccc|ccc}
\toprule
\multirow{2}{*}{\textbf{Model}} & \multirow{2}{*}{\textbf{STE $\downarrow$}} & \multicolumn{3}{c|}{\textbf{RMS (with modifiers)} $\uparrow$} & \multicolumn{3}{c}{\textbf{RMS (without modifiers)} $\uparrow$} \\ 
 & & \textbf{Precision} & \textbf{Recall} & \textbf{F1} & \textbf{Precision} & \textbf{Recall} & \textbf{F1} \\
\midrule
Gemini-2.5-Flash & 0.8159 & 0.0000 & 0.0000 & 0.0000 & 0.0000 & 0.0000 & 0.0000 \\
Gemini-2.5-Pro & 0.4504 & 0.1250 & 0.1250 & 0.1250 & 0.1250 & 0.1250 & 0.1250 \\
\midrule
Gpt-4.1-mini & 0.7683 & 0.0000 & 0.0000 & 0.0000 & 0.0000 & 0.0000 & 0.0000 \\
Gpt-4.1 & 0.7003 & 0.0000 & 0.0000 & 0.0000 & 0.0000 & 0.0000 & 0.0000 \\
\midrule
Claude-3.5-Haiku & 0.8223 & 0.0000 & 0.0000 & 0.0000 & 0.0000 & 0.0000 & 0.0000 \\
Claude-3.7-Sonnet & 0.6243 & 0.0000 & 0.0000 & 0.0000 & 0.0000 & 0.0000 & 0.0000 \\
\bottomrule
\end{tabular}
\end{table}

\begin{table}[H]
\centering
\caption{Detailed results for benchmark BIOMD0000000359 across all models.}
\label{tab:BIOMD0000000359}
\begin{tabular}{l|c|ccc|ccc}
\toprule
\multirow{2}{*}{\textbf{Model}} & \multirow{2}{*}{\textbf{STE $\downarrow$}} & \multicolumn{3}{c|}{\textbf{RMS (with modifiers)} $\uparrow$} & \multicolumn{3}{c}{\textbf{RMS (without modifiers)} $\uparrow$} \\ 
 & & \textbf{Precision} & \textbf{Recall} & \textbf{F1} & \textbf{Precision} & \textbf{Recall} & \textbf{F1} \\
\midrule
Gemini-2.5-Flash & 0.8930 & 0.1250 & 0.1250 & 0.1250 & 0.1250 & 0.1250 & 0.1250 \\
Gemini-2.5-Pro & 0.4009 & 0.1429 & 0.1250 & 0.1333 & 0.1429 & 0.1250 & 0.1333 \\
\midrule
Gpt-4.1-mini & 0.7740 & 0.0000 & 0.0000 & 0.0000 & 0.0000 & 0.0000 & 0.0000 \\
Gpt-4.1 & 0.8763 & 0.0000 & 0.0000 & 0.0000 & 0.0000 & 0.0000 & 0.0000 \\
\midrule
Claude-3.5-Haiku & 0.7591 & 0.0000 & 0.0000 & 0.0000 & 0.0000 & 0.0000 & 0.0000 \\
Claude-3.7-Sonnet & 0.7117 & 0.1250 & 0.1250 & 0.1250 & 0.1250 & 0.1250 & 0.1250 \\
\bottomrule
\end{tabular}
\end{table}

\begin{table}[H]
\centering
\caption{Detailed results for benchmark BIOMD0000000360 across all models.}
\label{tab:BIOMD0000000360}
\begin{tabular}{l|c|ccc|ccc}
\toprule
\multirow{2}{*}{\textbf{Model}} & \multirow{2}{*}{\textbf{STE $\downarrow$}} & \multicolumn{3}{c|}{\textbf{RMS (with modifiers)} $\uparrow$} & \multicolumn{3}{c}{\textbf{RMS (without modifiers)} $\uparrow$} \\ 
 & & \textbf{Precision} & \textbf{Recall} & \textbf{F1} & \textbf{Precision} & \textbf{Recall} & \textbf{F1} \\
\midrule
Gemini-2.5-Flash & 0.6096 & 0.0000 & 0.0000 & 0.0000 & 0.0000 & 0.0000 & 0.0000 \\
Gemini-2.5-Pro & 0.5411 & 0.0000 & 0.0000 & 0.0000 & 0.0000 & 0.0000 & 0.0000 \\
\midrule
Gpt-4.1-mini & 0.8952 & 0.0000 & 0.0000 & 0.0000 & 0.0000 & 0.0000 & 0.0000 \\
Gpt-4.1 & 0.7970 & 0.0000 & 0.0000 & 0.0000 & 0.0000 & 0.0000 & 0.0000 \\
\midrule
Claude-3.5-Haiku & 0.8390 & 0.0000 & 0.0000 & 0.0000 & 0.0000 & 0.0000 & 0.0000 \\
Claude-3.7-Sonnet & 0.5745 & 0.0000 & 0.0000 & 0.0000 & 0.0000 & 0.0000 & 0.0000 \\
\bottomrule
\end{tabular}
\end{table}

\begin{table}[H]
\centering
\caption{Detailed results for benchmark BIOMD0000000361 across all models.}
\label{tab:BIOMD0000000361}
\begin{tabular}{l|c|ccc|ccc}
\toprule
\multirow{2}{*}{\textbf{Model}} & \multirow{2}{*}{\textbf{STE $\downarrow$}} & \multicolumn{3}{c|}{\textbf{RMS (with modifiers)} $\uparrow$} & \multicolumn{3}{c}{\textbf{RMS (without modifiers)} $\uparrow$} \\ 
 & & \textbf{Precision} & \textbf{Recall} & \textbf{F1} & \textbf{Precision} & \textbf{Recall} & \textbf{F1} \\
\midrule
Gemini-2.5-Flash & 0.5061 & 0.1667 & 0.2000 & 0.1818 & 0.1667 & 0.2000 & 0.1818 \\
Gemini-2.5-Pro & 0.4931 & 0.1667 & 0.2000 & 0.1818 & 0.1667 & 0.2000 & 0.1818 \\
\midrule
Gpt-4.1-mini & 0.9735 & 0.0556 & 0.2000 & 0.0870 & 0.0556 & 0.2000 & 0.0870 \\
Gpt-4.1 & 0.5194 & 0.2500 & 0.2000 & 0.2222 & 0.2500 & 0.2000 & 0.2222 \\
\midrule
Claude-3.5-Haiku & 0.9661 & 0.0000 & 0.0000 & 0.0000 & 0.0000 & 0.0000 & 0.0000 \\
Claude-3.7-Sonnet & 0.5663 & 0.2500 & 0.4000 & 0.3077 & 0.2500 & 0.4000 & 0.3077 \\
\bottomrule
\end{tabular}
\end{table}

\begin{table}[H]
\centering
\caption{Detailed results for benchmark BIOMD0000000363 across all models.}
\label{tab:BIOMD0000000363}
\begin{tabular}{l|c|ccc|ccc}
\toprule
\multirow{2}{*}{\textbf{Model}} & \multirow{2}{*}{\textbf{STE $\downarrow$}} & \multicolumn{3}{c|}{\textbf{RMS (with modifiers)} $\uparrow$} & \multicolumn{3}{c}{\textbf{RMS (without modifiers)} $\uparrow$} \\ 
 & & \textbf{Precision} & \textbf{Recall} & \textbf{F1} & \textbf{Precision} & \textbf{Recall} & \textbf{F1} \\
\midrule
Gemini-2.5-Flash & 0.2280 & 0.5000 & 0.5000 & 0.5000 & 0.5000 & 0.5000 & 0.5000 \\
Gemini-2.5-Pro & 0.0018 & 1.0000 & 0.7500 & 0.8571 & 1.0000 & 0.7500 & 0.8571 \\
\midrule
Gpt-4.1-mini & 0.7069 & 0.0000 & 0.0000 & 0.0000 & 0.0000 & 0.0000 & 0.0000 \\
Gpt-4.1 & 0.1685 & 1.0000 & 0.7500 & 0.8571 & 1.0000 & 0.7500 & 0.8571 \\
\midrule
Claude-3.5-Haiku & 0.8193 & 0.3333 & 0.2500 & 0.2857 & 0.3333 & 0.2500 & 0.2857 \\
Claude-3.7-Sonnet & 0.0678 & 0.7500 & 0.7500 & 0.7500 & 0.7500 & 0.7500 & 0.7500 \\
\bottomrule
\end{tabular}
\end{table}

\begin{table}[H]
\centering
\caption{Detailed results for benchmark BIOMD0000000405 across all models.}
\label{tab:BIOMD0000000405}
\begin{tabular}{l|c|ccc|ccc}
\toprule
\multirow{2}{*}{\textbf{Model}} & \multirow{2}{*}{\textbf{STE $\downarrow$}} & \multicolumn{3}{c|}{\textbf{RMS (with modifiers)} $\uparrow$} & \multicolumn{3}{c}{\textbf{RMS (without modifiers)} $\uparrow$} \\ 
 & & \textbf{Precision} & \textbf{Recall} & \textbf{F1} & \textbf{Precision} & \textbf{Recall} & \textbf{F1} \\
\midrule
Gemini-2.5-Flash & 0.0563 & 0.0000 & 0.0000 & 0.0000 & 0.0000 & 0.0000 & 0.0000 \\
Gemini-2.5-Pro & 1.0000 & 0.0000 & 0.0000 & 0.0000 & 0.0000 & 0.0000 & 0.0000 \\
\midrule
Gpt-4.1-mini & 0.4653 & 0.0000 & 0.0000 & 0.0000 & 0.0000 & 0.0000 & 0.0000 \\
Gpt-4.1 & 0.3312 & 0.0000 & 0.0000 & 0.0000 & 0.0000 & 0.0000 & 0.0000 \\
\midrule
Claude-3.5-Haiku & 0.6334 & 0.0000 & 0.0000 & 0.0000 & 0.0000 & 0.0000 & 0.0000 \\
Claude-3.7-Sonnet & 0.4975 & 0.0000 & 0.0000 & 0.0000 & 0.0000 & 0.0000 & 0.0000 \\
\bottomrule
\end{tabular}
\end{table}

\begin{table}[H]
\centering
\caption{Detailed results for benchmark BIOMD0000000413 across all models.}
\label{tab:BIOMD0000000413}
\begin{tabular}{l|c|ccc|ccc}
\toprule
\multirow{2}{*}{\textbf{Model}} & \multirow{2}{*}{\textbf{STE $\downarrow$}} & \multicolumn{3}{c|}{\textbf{RMS (with modifiers)} $\uparrow$} & \multicolumn{3}{c}{\textbf{RMS (without modifiers)} $\uparrow$} \\ 
 & & \textbf{Precision} & \textbf{Recall} & \textbf{F1} & \textbf{Precision} & \textbf{Recall} & \textbf{F1} \\
\midrule
Gemini-2.5-Flash & 0.1108 & 0.5000 & 0.2222 & 0.3077 & 0.5000 & 0.2222 & 0.3077 \\
Gemini-2.5-Pro & 0.0726 & 0.2000 & 0.1111 & 0.1429 & 0.2000 & 0.1111 & 0.1429 \\
\midrule
Gpt-4.1-mini & 0.4716 & 0.0000 & 0.0000 & 0.0000 & 0.0000 & 0.0000 & 0.0000 \\
Gpt-4.1 & 0.0924 & 0.8000 & 0.4444 & 0.5714 & 0.8000 & 0.4444 & 0.5714 \\
\midrule
Claude-3.5-Haiku & 0.5709 & 0.0000 & 0.0000 & 0.0000 & 0.0000 & 0.0000 & 0.0000 \\
Claude-3.7-Sonnet & 0.0824 & 0.3333 & 0.2222 & 0.2667 & 0.4000 & 0.2222 & 0.2857 \\
\bottomrule
\end{tabular}
\end{table}

\begin{table}[H]
\centering
\caption{Detailed results for benchmark BIOMD0000000435 across all models.}
\label{tab:BIOMD0000000435}
\begin{tabular}{l|c|ccc|ccc}
\toprule
\multirow{2}{*}{\textbf{Model}} & \multirow{2}{*}{\textbf{STE $\downarrow$}} & \multicolumn{3}{c|}{\textbf{RMS (with modifiers)} $\uparrow$} & \multicolumn{3}{c}{\textbf{RMS (without modifiers)} $\uparrow$} \\ 
 & & \textbf{Precision} & \textbf{Recall} & \textbf{F1} & \textbf{Precision} & \textbf{Recall} & \textbf{F1} \\
\midrule
Gemini-2.5-Flash & 0.3934 & 0.0000 & 0.0000 & 0.0000 & 0.6667 & 0.2500 & 0.3636 \\
Gemini-2.5-Pro & 0.5931 & 0.0000 & 0.0000 & 0.0000 & 0.0000 & 0.0000 & 0.0000 \\
\midrule
Gpt-4.1-mini & 0.9993 & 0.0000 & 0.0000 & 0.0000 & 0.0000 & 0.0000 & 0.0000 \\
Gpt-4.1 & 0.7326 & 0.0000 & 0.0000 & 0.0000 & 0.0000 & 0.0000 & 0.0000 \\
\midrule
Claude-3.5-Haiku & 0.4821 & 0.0000 & 0.0000 & 0.0000 & 0.0000 & 0.0000 & 0.0000 \\
Claude-3.7-Sonnet & 0.6931 & 0.0000 & 0.0000 & 0.0000 & 1.0000 & 1.0000 & 1.0000 \\
\bottomrule
\end{tabular}
\end{table}

\begin{table}[H]
\centering
\caption{Detailed results for benchmark BIOMD0000000438 across all models.}
\label{tab:BIOMD0000000438}
\begin{tabular}{l|c|ccc|ccc}
\toprule
\multirow{2}{*}{\textbf{Model}} & \multirow{2}{*}{\textbf{STE $\downarrow$}} & \multicolumn{3}{c|}{\textbf{RMS (with modifiers)} $\uparrow$} & \multicolumn{3}{c}{\textbf{RMS (without modifiers)} $\uparrow$} \\ 
 & & \textbf{Precision} & \textbf{Recall} & \textbf{F1} & \textbf{Precision} & \textbf{Recall} & \textbf{F1} \\
\midrule
Gemini-2.5-Flash & 0.6830 & 0.0000 & 0.0000 & 0.0000 & 0.0000 & 0.0000 & 0.0000 \\
Gemini-2.5-Pro & 0.4543 & 0.0000 & 0.0000 & 0.0000 & 0.2000 & 0.1429 & 0.1667 \\
\midrule
Gpt-4.1-mini & 0.8708 & 0.0000 & 0.0000 & 0.0000 & 0.0000 & 0.0000 & 0.0000 \\
Gpt-4.1 & 0.6301 & 0.0000 & 0.0000 & 0.0000 & 0.0000 & 0.0000 & 0.0000 \\
\midrule
Claude-3.5-Haiku & 0.8708 & 0.0000 & 0.0000 & 0.0000 & 0.0000 & 0.0000 & 0.0000 \\
Claude-3.7-Sonnet & 0.7962 & 0.0000 & 0.0000 & 0.0000 & 0.0000 & 0.0000 & 0.0000 \\
\bottomrule
\end{tabular}
\end{table}

\begin{table}[H]
\centering
\caption{Detailed results for benchmark BIOMD0000000454 across all models.}
\label{tab:BIOMD0000000454}
\begin{tabular}{l|c|ccc|ccc}
\toprule
\multirow{2}{*}{\textbf{Model}} & \multirow{2}{*}{\textbf{STE $\downarrow$}} & \multicolumn{3}{c|}{\textbf{RMS (with modifiers)} $\uparrow$} & \multicolumn{3}{c}{\textbf{RMS (without modifiers)} $\uparrow$} \\ 
 & & \textbf{Precision} & \textbf{Recall} & \textbf{F1} & \textbf{Precision} & \textbf{Recall} & \textbf{F1} \\
\midrule
Gemini-2.5-Flash & 0.0161 & 0.0000 & 0.0000 & 0.0000 & 0.3333 & 0.2500 & 0.2857 \\
Gemini-2.5-Pro & 0.0020 & 0.0000 & 0.0000 & 0.0000 & 0.2500 & 0.2500 & 0.2500 \\
\midrule
Gpt-4.1-mini & 0.2175 & 0.0000 & 0.0000 & 0.0000 & 0.0000 & 0.0000 & 0.0000 \\
Gpt-4.1 & 0.3276 & 0.0000 & 0.0000 & 0.0000 & 0.0000 & 0.0000 & 0.0000 \\
\midrule
Claude-3.5-Haiku & 0.1641 & 0.0000 & 0.0000 & 0.0000 & 0.0000 & 0.0000 & 0.0000 \\
Claude-3.7-Sonnet & 0.0190 & 0.0000 & 0.0000 & 0.0000 & 0.0000 & 0.0000 & 0.0000 \\
\bottomrule
\end{tabular}
\end{table}

\begin{table}[H]
\centering
\caption{Detailed results for benchmark BIOMD0000000455 across all models.}
\label{tab:BIOMD0000000455}
\begin{tabular}{l|c|ccc|ccc}
\toprule
\multirow{2}{*}{\textbf{Model}} & \multirow{2}{*}{\textbf{STE $\downarrow$}} & \multicolumn{3}{c|}{\textbf{RMS (with modifiers)} $\uparrow$} & \multicolumn{3}{c}{\textbf{RMS (without modifiers)} $\uparrow$} \\ 
 & & \textbf{Precision} & \textbf{Recall} & \textbf{F1} & \textbf{Precision} & \textbf{Recall} & \textbf{F1} \\
\midrule
Gemini-2.5-Flash & 0.0056 & 0.0000 & 0.0000 & 0.0000 & 0.0000 & 0.0000 & 0.0000 \\
Gemini-2.5-Pro & 0.0005 & 0.0000 & 0.0000 & 0.0000 & 0.0000 & 0.0000 & 0.0000 \\
\midrule
Gpt-4.1-mini & 0.0000 & 0.0000 & 0.0000 & 0.0000 & 0.0000 & 0.0000 & 0.0000 \\
Gpt-4.1 & 0.1924 & 0.0000 & 0.0000 & 0.0000 & 0.0000 & 0.0000 & 0.0000 \\
\midrule
Claude-3.5-Haiku & 0.1399 & 0.0000 & 0.0000 & 0.0000 & 0.0000 & 0.0000 & 0.0000 \\
Claude-3.7-Sonnet & 0.0004 & 0.0000 & 0.0000 & 0.0000 & 0.0000 & 0.0000 & 0.0000 \\
\bottomrule
\end{tabular}
\end{table}

\begin{table}[H]
\centering
\caption{Detailed results for benchmark BIOMD0000000456 across all models.}
\label{tab:BIOMD0000000456}
\begin{tabular}{l|c|ccc|ccc}
\toprule
\multirow{2}{*}{\textbf{Model}} & \multirow{2}{*}{\textbf{STE $\downarrow$}} & \multicolumn{3}{c|}{\textbf{RMS (with modifiers)} $\uparrow$} & \multicolumn{3}{c}{\textbf{RMS (without modifiers)} $\uparrow$} \\ 
 & & \textbf{Precision} & \textbf{Recall} & \textbf{F1} & \textbf{Precision} & \textbf{Recall} & \textbf{F1} \\
\midrule
Gemini-2.5-Flash & 0.0001 & 0.0000 & 0.0000 & 0.0000 & 0.1429 & 0.1667 & 0.1538 \\
Gemini-2.5-Pro & 0.0033 & 0.0000 & 0.0000 & 0.0000 & 0.0000 & 0.0000 & 0.0000 \\
\midrule
Gpt-4.1-mini & 0.1217 & 0.0000 & 0.0000 & 0.0000 & 0.0000 & 0.0000 & 0.0000 \\
Gpt-4.1 & 0.1114 & 0.0000 & 0.0000 & 0.0000 & 0.0000 & 0.0000 & 0.0000 \\
\midrule
Claude-3.5-Haiku & 0.0934 & 0.0000 & 0.0000 & 0.0000 & 0.0000 & 0.0000 & 0.0000 \\
Claude-3.7-Sonnet & 0.0013 & 0.0000 & 0.0000 & 0.0000 & 0.0000 & 0.0000 & 0.0000 \\
\bottomrule
\end{tabular}
\end{table}

\begin{table}[H]
\centering
\caption{Detailed results for benchmark BIOMD0000000458 across all models.}
\label{tab:BIOMD0000000458}
\begin{tabular}{l|c|ccc|ccc}
\toprule
\multirow{2}{*}{\textbf{Model}} & \multirow{2}{*}{\textbf{STE $\downarrow$}} & \multicolumn{3}{c|}{\textbf{RMS (with modifiers)} $\uparrow$} & \multicolumn{3}{c}{\textbf{RMS (without modifiers)} $\uparrow$} \\ 
 & & \textbf{Precision} & \textbf{Recall} & \textbf{F1} & \textbf{Precision} & \textbf{Recall} & \textbf{F1} \\
\midrule
Gemini-2.5-Flash & 0.1334 & 0.0000 & 0.0000 & 0.0000 & 0.0000 & 0.0000 & 0.0000 \\
Gemini-2.5-Pro & 0.1455 & 0.0000 & 0.0000 & 0.0000 & 0.5000 & 0.3333 & 0.4000 \\
\midrule
Gpt-4.1-mini & 0.2124 & 0.0000 & 0.0000 & 0.0000 & 0.0000 & 0.0000 & 0.0000 \\
Gpt-4.1 & 0.2314 & 0.0000 & 0.0000 & 0.0000 & 0.0000 & 0.0000 & 0.0000 \\
\midrule
Claude-3.5-Haiku & 0.2269 & 0.0000 & 0.0000 & 0.0000 & 0.5000 & 0.3333 & 0.4000 \\
Claude-3.7-Sonnet & 0.2100 & 0.0000 & 0.0000 & 0.0000 & 0.0000 & 0.0000 & 0.0000 \\
\bottomrule
\end{tabular}
\end{table}

\begin{table}[H]
\centering
\caption{Detailed results for benchmark BIOMD0000000459 across all models.}
\label{tab:BIOMD0000000459}
\begin{tabular}{l|c|ccc|ccc}
\toprule
\multirow{2}{*}{\textbf{Model}} & \multirow{2}{*}{\textbf{STE $\downarrow$}} & \multicolumn{3}{c|}{\textbf{RMS (with modifiers)} $\uparrow$} & \multicolumn{3}{c}{\textbf{RMS (without modifiers)} $\uparrow$} \\ 
 & & \textbf{Precision} & \textbf{Recall} & \textbf{F1} & \textbf{Precision} & \textbf{Recall} & \textbf{F1} \\
\midrule
Gemini-2.5-Flash & 0.1024 & 0.0000 & 0.0000 & 0.0000 & 0.2500 & 0.3333 & 0.2857 \\
Gemini-2.5-Pro & 0.0935 & 0.0000 & 0.0000 & 0.0000 & 0.3333 & 0.3333 & 0.3333 \\
\midrule
Gpt-4.1-mini & 0.6115 & 0.0000 & 0.0000 & 0.0000 & 0.6667 & 0.6667 & 0.6667 \\
Gpt-4.1 & 0.7493 & 0.0000 & 0.0000 & 0.0000 & 0.3333 & 0.3333 & 0.3333 \\
\midrule
Claude-3.5-Haiku & 0.6208 & 0.0000 & 0.0000 & 0.0000 & 0.6667 & 0.6667 & 0.6667 \\
Claude-3.7-Sonnet & 0.1001 & 0.0000 & 0.0000 & 0.0000 & 0.3750 & 1.0000 & 0.5455 \\
\bottomrule
\end{tabular}
\end{table}

\begin{table}[H]
\centering
\caption{Detailed results for benchmark BIOMD0000000460 across all models.}
\label{tab:BIOMD0000000460}
\begin{tabular}{l|c|ccc|ccc}
\toprule
\multirow{2}{*}{\textbf{Model}} & \multirow{2}{*}{\textbf{STE $\downarrow$}} & \multicolumn{3}{c|}{\textbf{RMS (with modifiers)} $\uparrow$} & \multicolumn{3}{c}{\textbf{RMS (without modifiers)} $\uparrow$} \\ 
 & & \textbf{Precision} & \textbf{Recall} & \textbf{F1} & \textbf{Precision} & \textbf{Recall} & \textbf{F1} \\
\midrule
Gemini-2.5-Flash & 0.4210 & 0.0000 & 0.0000 & 0.0000 & 0.1667 & 0.3333 & 0.2222 \\
Gemini-2.5-Pro & 0.1022 & 0.0000 & 0.0000 & 0.0000 & 0.6000 & 1.0000 & 0.7500 \\
\midrule
Gpt-4.1-mini & 0.4094 & 0.0000 & 0.0000 & 0.0000 & 0.6000 & 1.0000 & 0.7500 \\
Gpt-4.1 & 0.5281 & 0.0000 & 0.0000 & 0.0000 & 0.1667 & 0.3333 & 0.2222 \\
\midrule
Claude-3.5-Haiku & 0.4153 & 0.0000 & 0.0000 & 0.0000 & 0.6667 & 0.6667 & 0.6667 \\
Claude-3.7-Sonnet & 0.2646 & 0.0000 & 0.0000 & 0.0000 & 0.6000 & 1.0000 & 0.7500 \\
\bottomrule
\end{tabular}%
\end{table}

\end{document}